\documentclass{article} 
\usepackage{style,times}

\usepackage{amsmath,amsfonts,bm}

\def\eqref#1{equation~\ref{#1}}

\def\1{\bm{1}}

\DeclareMathAlphabet{\mathsfit}{\encodingdefault}{\sfdefault}{m}{sl}
\SetMathAlphabet{\mathsfit}{bold}{\encodingdefault}{\sfdefault}{bx}{n}

\usepackage{graphicx,subcaption}
\usepackage{microtype, xspace}
\usepackage{hyperref}
\usepackage{url}
\usepackage{booktabs,multirow,longtable}
\usepackage{algorithm}
\usepackage[noend]{algpseudocode}
\usepackage{amsmath,amssymb,mathtools}
\usepackage{tabularx}
\usepackage{tikz}
\usetikzlibrary{arrows.meta,calc,shapes.geometric}

\title{DASH: Fast, Valid Counterfactuals for Deep Networks via Batched Directional Search}

\newcommand{\DASH}{\textsc{DASH}\xspace}

\author{
\begin{tabular}[t]{@{}l@{\qquad}l@{}}
\textbf{Shraman Pal}\textsuperscript{1,*} &
\textbf{Gabriel Medeiros}\textsuperscript{2,*}\\
{\normalfont\texttt{pal52@purdue.edu}} &
{\normalfont\texttt{medeiros.gabriel@ime.eb.br}}\\[0.6em]
\textbf{Clayton Escouper das Chagas}\textsuperscript{2} & \textbf{Can Li}\textsuperscript{1}
\\
{\normalfont\texttt{escouper@ime.eb.br}} & {\normalfont\texttt{canli@purdue.edu}}
\\[0.8em]
\multicolumn{2}{@{}l}{{\normalfont\textsuperscript{1} Davidson School of Chemical Engineering,
Purdue University, West Lafayette, IN 47907, USA}}\\
\multicolumn{2}{@{}l}{{\normalfont\textsuperscript{2} Computer Engineering Department,
Instituto Militar de Engenharia, Rio de Janeiro - RJ, Brazil}}
\end{tabular}
}

\newtheorem{prop}{Proposition}

\iclrfinalcopy
\begin{document}

\maketitle
\begin{center}
\small \textsuperscript{*}These authors contributed equally.
\end{center}

\begin{abstract}
Counterfactual explanations are most useful when they can be generated with low latency, remain close to the factual input, and satisfy input-domain, categorical, and actionability constraints. Achieving these objectives simultaneously is challenging for deep neural networks. Heuristic methods are often fast but may return invalid counterfactuals, whereas exact methods can certify global proximity but may not finish within practical time limits. We introduce \DASH, a batched heuristic search method for finding close, valid, and actionable counterfactuals for deep neural networks under $\ell_1$, $\ell_2$, and $\ell_\infty$ objectives. \DASH uses directional Lipschitz bounds and local affine models to generate anchors, then ranks and expands promising regions with batched network evaluations. We compare \DASH against nine prior heuristic methods and time-limited exact mixed-integer baselines on four tabular datasets, with network depths from 2 to 32, and evaluate scalability on PBMC3k. Across 9,000 tabular query-norm cases, \DASH returns a valid counterfactual within $5\%$ of the best heuristic-observed valid distance in $94.6\%$ of cases, with a median CPU search runtime of $0.061$\,s. PGD-bisect, the baseline with the highest pooled within-$5\%$ coverage, meets this criterion in $41.1\%$ of cases, with a median runtime of $0.298$\,s. These results show that the proposed search maintains high valid proximity across norms while keeping its search runtime practical.
\end{abstract}

\section{Introduction}
\label{sec:introduction}

Neural networks are increasingly used as predictive models that inform decisions, in settings such as mortgage default assessment from financial transactions~\citep{KVAMME2018207}, clinical-risk predictions from electronic health records~\citep{Rajkomar2018}, and process fault diagnosis from operating measurements~\citep{Wu2018fault}. In such applications, a prediction can prompt a further question: \emph{what sufficiently small change to the input would make the model predict a different, desired outcome?} For example, a borrower may seek changes to financial  attributes that would enable them to cross a credit approval threshold, while an engineer may seek nearby operating conditions associated with a desired process state. Counterfactual explanations (CFEs) formalize this query by finding an alternative input that produces a desired target prediction while minimizing its deviation from the original or \emph{factual} input~\citep{Wachter2017CounterfactualEW, karimi2020mace}. Let \(f\) denote a fixed predictor, \(x^0\) a factual input, \(\mathcal{Y}_{\mathrm{tar}}\) the target-output set, and \(\mathcal{A}(x^0)\) the set of \textit{actionable} modifications. CFEs can be mathematically formulated as an optimization problem
\begin{equation}
\label{eqn:cf-problem}
\min_{x}\;\; d_\nu(x,x^0)
\quad
\text{subject to}
\quad
f(x)\in\mathcal{Y}_{\mathrm{tar}},
\quad
x\in\mathcal{A}(x^0),
\end{equation}

where \(d_\nu(x,x^0)\) is a normalized \(\ell_\nu\) distance between \(x\) and \(x^0\). We call a point \emph{target-valid} when it satisfies the target constraint \(f(x)\in\mathcal{Y}_{\mathrm{tar}}\), \emph{actionable} when \(x\in\mathcal{A}(x^0)\), and a \emph{valid} counterfactual when both conditions hold. The actionable set may impose feature bounds (e.g., employment duration), immutable coordinates (e.g., location), permitted directions of change (e.g., age), and categorical assignments (e.g., employment type)~\citep{ustun2019actionable}. CFE methods can be measured on four evaluation criteria: likelihood of producing valid counterfactuals (\emph{validity}), time it takes to produce a counterfactual (\emph{runtime}), proximity to the factual (\emph{distance}), and whether it provides a certificate of proximity or global optimality (\emph{optimality certificate}). 

Existing CFE methods make different compromises among these criteria. Penalty-based heuristic methods like Wachter~\citep{Wachter2017CounterfactualEW}, S-CFE~\citep{sadiku2025scfe}, and DiCE~\citep{mothilal2020dice} replace the target constraint with a differentiable prediction-loss term and optimize it jointly with an input distance penalty. Consequently, their termination condition does not guarantee validity or distance optimality of the proposed counterfactual. Other heuristics use discrete or sample-based search \citep{brughmans2024nice}. However, they neither enforce actionability constraints nor give an optimality guarantee. Exact approaches instead encode the predictor neural network, target constraint, and the actionable set as a satisfiability or mixed-integer optimization problem ~\citep{karimi2020mace, mohammadi2021scaling}. ReLU networks permit exact mixed-integer linear encoding due to their piecewise affine nature, whereas other non-linear activations require either mixed-integer non-linear encoding or mixed-integer linear approximation. Finding a minimum-distance counterfactual even for ReLU networks is NP-hard ~\citep{salzer2021reachability, weng2018towards}. Consequently, solvers can spend substantial computation proving optimality even after a valid counterfactual has been found, limiting their applicability to large networks under practical runtime budgets.

These approaches leave a gap between heuristic methods, whose returned outputs may violate target or actionability conditions, and exact methods, which may not complete within a practical time limit. To address target-validity, actionability and runtime, we propose \DASH, a heuristic search method that balances all three axes for \(\ell_1\), \(\ell_2\), and \(\ell_\infty\) norms. \DASH uses local sensitivity information and linearization to estimate the decision boundary to achieve high validity, a strict enforcement of actionability constraints at each iteration, and batching the network evaluations to reduce computational overhead. We evaluate this approach against prior heuristic methods, and exact, time-limited, mixed-integer baselines on four tabular datasets and the PBMC3k single-cell RNA sequencing dataset. Pooled across all 9,000 tabular query-norm cases, \DASH finds a valid counterfactual within \(5\%\) of the best heuristic-observed valid distance \(\sim2.3\) times more often than PGD-bisect, the baseline with the highest pooled within-\(5\%\) coverage, with a \(\sim 5\) times faster median CPU search runtime. Whereas prior scalability experiments \citep{mohammadi2021scaling} considered architectures containing at most approximately \(18{,}000\) parameters, our PBMC3k evaluation includes a network with more than \(100{,}000\) parameters. As a heuristic, \DASH does not certify non-existence or global optimality. Nevertheless, on the \(3{,}448\) tabular query-norm cases for which a time-limited exact solver certified the optimum, \DASH returned a counterfactual within \(1\%\) of that optimum in \(82.7\%\) of cases. We measure distance and validity of the found counterfactual, and runtime of the method, under $\ell_1$, $\ell_2$, and $\ell_\infty$ objectives, and additionally evaluate a GPU implementation on PBMC3k.

\section{Related Work}
\label{sec:related}

\paragraph{Adversarial Methods}
Related to CFEs are minimum-norm adversarial boundary-crossing problems~\citep{pawelczyk2022exploring}. Adversarial attack methods solve this through margin penalties or barriers ~\citep{carlini2017towards, Finlay2019TheLA}, adaptive norm constraints ~\citep{pintor2021fast} or local affine boundary models ~\citep{MoosaviDezfooli2015DeepFoolAS,croce2020fab}.  Our PGD-bisect baseline combines projected gradient descent \citep{madry2018towards} with bisection. Applying these search ideas to mixed-feature counterfactuals requires explicit treatment of actionable constraints like immutable features, categorical assignments, and permitted directions of change~\citep{ustun2019actionable}.

\paragraph{Heuristic CFE methods.}
For differentiable predictors such as neural networks, several CFE methods have used first-order optimization. \citet{Wachter2017CounterfactualEW} convert the target constraint into a differentiable quadratic loss, and jointly optimize it with a normalized distance function, using gradient descent algorithms. DiCE ~\citep{mothilal2020dice} and S-CFE ~\citep{sadiku2025scfe} add a diversity term and a non-smooth \(\ell_p\) (\(0\leq p \leq1\)) sparsity penalty term respectively. These objectives guide the search, but minimizing a penalty does not by itself certify that the returned point satisfies every target and actionability constraint. Other heuristic CFE methods rely on sampling. NICE~\citep{brughmans2024nice} identifies the nearest training instance with the desired class and successively replaces coordinates of the factual input with the corresponding values from the chosen training instance. Growing Spheres~\citep{laugel2018sphere} samples increasingly distant hyperspherical shells to locate a point with the desired prediction, whereas the Greedy method of \citet{Schut2021GeneratingIC} uses Jacobian-guided feature updates. The limitation of these heuristic methods is that they often fail to find valid counterfactuals.

\paragraph{Exact counterfactual methods.}

Certifying a nearest counterfactual requires finding a valid counterfactual and ruling out all closer actionable points. One family of exact methods uses satisfiability modulo theories (SMT). ReLUplex~\citep{katz2017reluplex} develops an SMT-based verification method for ReLU networks, while MACE~\citep{karimi2020mace} applies SMT to counterfactual explanations using repeated queries to bound the optimal distance. An alternative is mixed-integer programming (MIP), where ReLU networks can be encoded exactly with mixed-integer linear constraints~\citep{fischetti2018deep,tjeng2018evaluating}. \citet{mohammadi2021scaling} use this approach to compute nearest counterfactuals with guarantees. Bound tightening, presolve, and stronger ReLU formulations can improve the tractability of the MIP formulations~\citep{fischetti2018deep,GRIMSTAD2019106580,tjeng2018evaluating,tsay2021partition,huchette2026deep}. Despite these advances, exact certification can remain computationally expensive for larger networks.

\section{Methodology}
\label{sec:method}

\paragraph{Notations and problem definition}
The counterfactual problem and distance \(d_\nu\) are defined in Eq.~\ref{eqn:cf-problem}. The target constraint is defined as \(f(x) \leq \tau\), where \(f:\mathbb{R}^{d}\rightarrow\mathbb{R}\). We assume that \(f\) is represented by a ReLU neural network and is therefore piecewise affine. Appendix~\ref{app:target-orientation} details the class of counterfactuals covered by this definition. \DASH searches using the scalar score \(\phi(x)=f(x)-\tau\), with the desired output satisfying \(\phi(x)\leq-\mu\), where \(\mu\geq0\) is a verification margin. We define the target gap and target-improving gradient as
\begin{equation}
\label{eq:target-gap-gradient}
\Delta(x)=[\phi(x)+\mu]_+,\qquad g(x)=-\nabla\phi(x).
\end{equation}
For a continuous feature \(j\), the notation \(x^{j\leftarrow u}\) replaces only \(x_j\in\mathbb{R}\) with \(u\in\mathbb{R}\). The sets \(\Lambda_{\mathrm{ax}}\), \(\Lambda_{\mathrm{gr}}\), and \(\Lambda_{\mathrm{fac}}\) denote the step-size multipliers used by the anchor-generation mechanisms detailed in the next subsection. Their values are provided in Appendix~\ref{app:dash-settings}. 

\paragraph{Overview of DASH}
We call every point generated during the search an \emph{anchor}. At iteration \(t\), \(\mathcal B_t\) is the beam of target-invalid anchors. \DASH first perturbs single features to produce one-directional anchors \(\mathcal Y_t^{\mathrm{dir}}\). One-feature movements test whether a small intervention can reach the target. \DASH then ranks them to select at most \(K_{\mathrm{dir}}\) target-invalid anchors from \(\mathcal Y_t^{\mathrm{dir}}\cup\mathcal B_t\) for gradient evaluation; these selected anchors form \(\mathcal G_t\). Local gradient-based anchors are generated from \(\mathcal G_t\). As the different anchors could lie in different activation regions of the network~\citep{berzins2023polyhedral}, \DASH also generates factual-origin anchors from \(x^0\) using local affine models at \(\mathcal B_t\cup\mathcal G_t\). Target-valid anchors update the incumbent. A final ranking retains at most \(K_{\mathrm{beam}}\) target-invalid anchors for \(\mathcal B_{t+1}\). The search runs for at most \(T\) iterations, with the map \(\Pi_{\mathrm{act}}\) enforcing actionability on generated anchors. The following subsections define these anchor sets and their ranking rule; Algorithm~\ref{alg:dash} assembles them.

\subsection{Anchor generation}
\label{subsec:anchor-generation}
\paragraph{One-directional anchors.}
One-directional anchors change one feature at a time. For each continuous feature \(j\), \DASH precomputes a global coordinate-wise Lipschitz bound \(L_j\) from the network weights (Appendix~\ref{app:screening}; see also \citealp{weng2018towards}) satisfying
\begin{equation}
\label{eq:directional-lipschitz}
|\phi(x+h e_j)-\phi(x)|\leq L_j|h|
\end{equation}
over the bounded input domain.
\begin{prop}[One-feature exclusion bound]
\label{prop:one-feature-exclusion}
Let \(\phi(x)>-\mu\) and \(L_j>0\). For any \(y=x+h e_j\) in the input domain satisfying \(\phi(y)\leq-\mu\),
\[
|h|\geq\frac{\Delta(x)}{L_j}.
\]
\end{prop}
The bound motivates the one-directional anchor scale \(\rho_j(x)=\Delta(x)/(L_j+\epsilon)\). Let \(J_{\mathrm{cont}}\) contain mutable continuous-feature indices with permitted intervals \([a_j,b_j]\). Let \(J_{\mathrm{cat}}\) contain the one-hot coordinate blocks representing mutable categorical features. For \(j\in J_{\mathrm{cat}}\), let \(V_j(x^0)\) be the set of permitted vertices and let \(x^{j\leftarrow v}\) replace the entire block by \(v\). The resulting one-feature anchor set is
\begin{equation}
\label{eq:directional-anchors}
\begin{aligned}
\mathcal Y_t^{\mathrm{dir}}=\Pi_{\mathrm{act}}\Biggl(
\bigcup_{x\in\mathcal B_t}\Bigl[
&\bigcup_{j\in J_{\mathrm{cont}}}
\bigl(\{x\pm m\rho_j(x)e_j:m\in\Lambda_{\mathrm{ax}}\}
\cup\{x^{j\leftarrow a_j},x^{j\leftarrow b_j}\}\bigr)\\
&\cup\bigcup_{j\in J_{\mathrm{cat}}}\{x^{j\leftarrow v}:v\in V_j(x^0),\ v\ne x_j\}
\Bigr]\Biggr),
\end{aligned}
\end{equation}
The first union searches the continuous variables while the second union searches the categorical variables. If \(\phi\) is monotone along a permitted feature interval, evaluating its endpoint guarantees that a target-valid anchor is found if one exists on that interval.

\paragraph{Gradient-based anchors}
While the one-directional anchors perturb a single feature, gradient-based anchors change multiple features simultaneously by using the gradient at a target-invalid anchor \(x^k\) in two complementary ways. The gradient at \(x^k\) defines a local affine model
\begin{equation}
\label{eq:local-affine-model}
\phi(x^k+h)\approx\phi(x^k)-g(x^k)^\top h.
\end{equation}
Under this model, the displacement \(h\) must satisfy
\(g(x^k)^\top h\geq\Delta(x^k)\) to reach the decision boundary. Let \(\nu^\ast\) denote the dual-norm of norm $\nu$. By norm duality, the estimated minimum distance and the associated steepest directions are
\begin{equation}
\label{eq:boundary-distance}
\widehat r_\nu(x^k)=
\frac{\Delta(x^k)}{\|g(x^k)\|_{\nu^\ast}+\epsilon},
\end{equation}
\begin{equation}
\label{eq:norm-directions}
\begin{aligned}
v_\infty(g)&=\operatorname{sign}(g),
&v_2(g)&=\frac{g}{\|g\|_2},
&v_1(g)&=\operatorname{sign}(g_{j^\star})e_{j^\star},
\quad j^\star\in\arg\max_j|g_j|.
\end{aligned}
\end{equation}

\emph{Local gradient-based} anchors continue from \(x^k\) along \(v_\nu(g(x^k))\). From the selected target-invalid anchors \(\mathcal G_t\), the local gradient-based anchor set is
\begin{equation}
\label{eq:gradient-anchors}
\mathcal Y_t^{\mathrm{loc}}=\Pi_{\mathrm{act}}\biggl(\bigcup_{x\in\mathcal G_t}
\{x+m\widehat r_\nu(x)v_\nu(g(x)):m\in\Lambda_{\mathrm{gr}}\}\biggr).
\end{equation}
A full-gradient step can change many features which decreases sparsity. As most counterfactual domains prefer sparser actions, \DASH also generates support-restricted versions using the largest-magnitude gradient features. The configured support sizes are given in Appendix~\ref{app:dash-settings}.

\begin{prop}[Exactness of the affine boundary estimate]
\label{prop:affine-region-exactness}
Let \(\nu\in\{1,2,\infty\}\), and let \(x^k\) satisfy \(\phi(x^k)>-\mu\) and \(g(x^k)\neq0\). Define the unregularized affine-boundary anchor
\[
x_\nu^\star=x^k+\frac{\Delta(x^k)}{\|g(x^k)\|_{\nu^\ast}}\,
v_\nu(g(x^k)).
\]
If the segment from \(x^k\) to \(x_\nu^\star\) remains in one affine region of \(\phi\), then \(\phi(x_\nu^\star)=-\mu\). Moreover, \(x_\nu^\star\) minimizes \(\|x-x^k\|_\nu\) over the target half-space of that local affine model.
\end{prop}

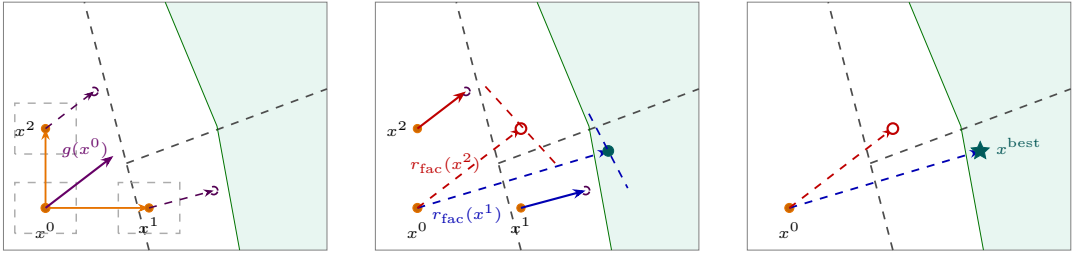
\begin{figure*}[t]
    \centering
\begin{tikzpicture}[
  x=1cm,y=1cm,
  every node/.style={font=\scriptsize},
  panel/.style={draw=black!35, rounded corners=2pt, line width=.35pt},
  cell/.style={draw=none},
  reluFacet/.style={draw=black!70, dashed, line width=.65pt},
  threshold/.style={draw=green!45!black,line width=.35pt},
  factual/.style={draw=orange!90!black,fill=orange!80!black},
  scout/.style={draw=orange!90!black,-{Stealth[length=1.5mm]},line width=.7pt},
  grad/.style={draw=violet!80!black,-{Stealth[length=1.7mm]},line width=.8pt},
  scoutPoint/.style={draw=orange!90!black,fill=orange!80!black},
  localProposal/.style={draw=violet!65!black,dashed,-{Stealth[length=1.35mm]},line width=.6pt},
  proposalPoint/.style={draw=violet!65!black,dashed,line width=.65pt},
  gradPone/.style={draw=blue!70!black,-{Stealth[length=1.7mm]},line width=.8pt},
  gradPtwo/.style={draw=red!75!black,-{Stealth[length=1.7mm]},line width=.8pt},
  tangentPone/.style={draw=blue!70!black,dashed,-{Stealth[length=1.4mm]},line width=.7pt},
  tangentPtwo/.style={draw=red!75!black,dashed,-{Stealth[length=1.4mm]},line width=.7pt},
  boundaryPone/.style={draw=blue!70!black,dashed,line width=.65pt},
  boundaryPtwo/.style={draw=red!75!black,dashed,line width=.65pt},
  infeasibleCandidate/.style={draw=red!75!black,fill=white,line width=.8pt}
]

\foreach \X in {0,4.92,9.84} {
  \begin{scope}[shift={(\X,0)}]

    \begin{scope}[shift={(.28cm,.24cm)},x=4.28cm,y=3.28cm]
      \path[cell,fill=green!20!teal!9]
        (.50,1)--(.66,.50)--(.73,0)--(1,0)--(1,1)--cycle;
      \draw[black!50,line width=.35pt] (0,0) rectangle (1,1);
      \draw[reluFacet] (.45,0)--(.25,1);
      \draw[reluFacet] (.38,.35)--(1,.65);
      \draw[threshold] (.50,1)--(.66,.50)--(.73,0);

      \coordinate (x0) at (.13,.17);
      \fill[factual] (x0) circle (1.7pt);
      \node[anchor=north,font=\tiny] at (.13,.15) {\(x^0\)};
    \end{scope}

    \ifdim\X pt=0pt
      \begin{scope}[shift={(.28cm,.24cm)},x=4.28cm,y=3.28cm]
        \coordinate (x0a) at (.13,.17);
        \coordinate (pone) at (.45,.17);
        \coordinate (ptwo) at (.13,.49);
        \draw[gray!65,dashed,line width=.55pt]
          (.035,.0675) rectangle (.225,.2725);
        \draw[gray!65,dashed,line width=.55pt]
          (.355,.0675) rectangle (.545,.2725);
        \draw[gray!65,dashed,line width=.55pt]
          (.035,.3875) rectangle (.225,.5925);
        \draw[scout] (x0a)--(pone);
        \draw[scout] (x0a)--(ptwo);
        \fill[scoutPoint] (pone) circle (1.55pt);
        \fill[scoutPoint] (ptwo) circle (1.55pt);
        \node[below,font=\tiny] at (pone) {\(x^1\)};
        \node[left,font=\tiny] at (ptwo) {\(x^2\)};
        \coordinate (qone) at (.65,.24);
        \coordinate (qtwo) at (.28,.64);
        \draw[localProposal] (pone)--(qone);
        \draw[localProposal] (ptwo)--(qtwo);
        \draw[proposalPoint] (qone) circle (1.5pt);
        \draw[proposalPoint] (qtwo) circle (1.5pt);
        \draw[grad] (x0a)--(.34,.38);
        \node[anchor=west,text=violet!80!black,font=\tiny]
          at (.15,.41) {\(g(x^0)\)};
      \end{scope}
    \fi

    \ifdim\X pt=4.92pt
      \begin{scope}[shift={(.28cm,.24cm)},x=4.28cm,y=3.28cm]
        \coordinate (x0t) at (.13,.17);
        \coordinate (poneg) at (.45,.17);
        \coordinate (ptwog) at (.13,.49);
        \fill[scoutPoint] (poneg) circle (1.55pt);
        \fill[scoutPoint] (ptwog) circle (1.55pt);
        \node[below,font=\tiny] at (poneg) {\(x^1\)};
        \node[left,font=\tiny] at (ptwog) {\(x^2\)};
        \draw[gradPone] (poneg)--(.65,.24);
        \draw[gradPtwo] (ptwog)--(.28,.64);
        \draw[proposalPoint] (.65,.24) circle (1.5pt);
        \draw[proposalPoint] (.28,.64) circle (1.5pt);
        \draw[tangentPone] (x0t)--(.72,.40)
          node[pos=.26,below=0.5pt,font=\tiny,text=blue!70!black]
          {\(r_{\mathrm{fac}}(x^1)\)};
        \draw[tangentPtwo] (x0t)--(.45,.49)
          node[pos=.27,above=1.5pt,font=\tiny,text=red!75!black]
          {\(r_{\mathrm{fac}}(x^2)\)};
        \draw[infeasibleCandidate] (.45,.49) circle (2.0pt);
        \filldraw[draw=teal!75!black,fill=teal!70!black]
          (.72,.40) circle (2.0pt);
        \draw[boundaryPone] (.66,.55)--(.78,.25);
        \draw[boundaryPtwo] (.34,.66)--(.57,.33);
      \end{scope}
    \fi

    \ifdim\X pt=9.84pt
      \begin{scope}[shift={(.28cm,.24cm)},x=4.28cm,y=3.28cm]
        \coordinate (x0e) at (.13,.17);
        \coordinate (ytanone) at (.72,.40);
        \coordinate (ytantwo) at (.45,.49);
        \draw[tangentPone] (x0e)--(ytanone);
        \draw[tangentPtwo] (x0e)--(ytantwo);
        \draw[infeasibleCandidate] (ytantwo) circle (2.0pt);
        \node[star,star points=5,star point ratio=2.1,
              draw=teal!75!black,fill=teal!70!black,
              minimum size=7pt,inner sep=0pt] (best) at (ytanone) {};
        \node[anchor=west,font=\tiny,text=teal!65!black] at (.74,.42)
          {\(x^{\mathrm{best}}\)};
      \end{scope}
    \fi

  \end{scope}
}
\end{tikzpicture}
\caption{\textbf{Anchor generation across ReLU regions.}
Black dashed lines separate activation regions, the green line is the target boundary, and the shaded region is target-valid. \textbf{left:} One-directional anchors \(x^1\) and \(x^2\) are in different regions. The gradient at each anchor is illustrated with colored dashed lines. \textbf{center:} Solid arrows show local gradient-based anchors from \(x^1\) and \(x^2\); dashed arrows show factual-origin anchors from the factual input \(x^0\). The latter use radii \(r_{\mathrm{fac}}(x^1)\) and \(r_{\mathrm{fac}}(x^2)\), as estimates to the target boundary (Eq.~\ref{eq:tangent-radius}). \textbf{right:} The red anchor will be used for further anchor generation. The blue anchor is the new incumbent 
 \(x^{\mathrm{best}}\).}
    \label{fig:search}
\end{figure*}

\emph{Factual-origin} anchors use the same direction and local affine model but originate from the factual input. This construction is motivated by seeking a boundary point near \(x^0\) under that local model, rather than taking another step from \(x^k\). \DASH estimates the decision boundary radius  \(r_{\mathrm{fac}}(x^k)\) from the factual, based on the local affine model at \(x^k\),
\begin{equation}
\label{eq:tangent-radius}
\phi(x^k)-g(x^k)^\top
\bigl(x^0+r_{\mathrm{fac}}(x^k)v_\nu(g(x^k))-x^k\bigr)=-\mu.
\end{equation}
The corresponding anchor set generated from the factual uses gradients evaluated at both the current beam \(\mathcal{B}_t\) and the anchors selected for gradient evaluation \(\mathcal{G}_t\):
\begin{equation}
\label{eq:tangent-anchors}
\mathcal Y_t^{\mathrm{fac}}=\Pi_{\mathrm{act}}\biggl(\bigcup_{x\in\mathcal B_t\cup\mathcal G_t}
\{x^0+m r_{\mathrm{fac}}(x)v_\nu(g(x)):m\in\Lambda_{\mathrm{fac}}\}\biggr).
\end{equation}
The two sets may overlap when a beam anchor is selected again in \(\mathcal G_t\). Geometrically, Eqn.~\ref{eq:tangent-radius} intersects the ray from \(x^0\) with the decision hyperplane predicted by the affine model at \(x^k\). Figure~\ref{fig:search} illustrates how the two anchor-generation mechanisms expose different affine regions. Appendix~\ref{app:factual-origin-ablation} reports a matched ablation testing the contribution of factual-origin anchors to proximity.

\paragraph{Actionability enforcement.}

For DASH, search-time actionability comprises feature bounds, immutable features, permitted directions of change, and valid categorical assignments. The enforcement map \(\Pi_{\mathrm{act}}\) applies these constraints to each generated anchor and acts elementwise on anchor sets. Its implementation is given in Appendix~\ref{app:actionability-implementation}.

\begin{prop}[Actionability preservation]
\label{prop:actionability-preservation}
Every anchor at every iteration evaluated by \DASH belongs to \(\mathcal A(x^0)\). In particular,
\( x \in\mathcal A(x^0), \;\; \forall x \in \mathcal B_t, \;\;\; \forall t\).
\end{prop}

Proofs of Propositions~\ref{prop:affine-region-exactness} and \ref{prop:actionability-preservation} are provided in
Appendix~\ref{app:method-properties}.

\begin{algorithm}[t]
\caption{\DASH: batched anchor search}
\label{alg:dash}
\begin{algorithmic}[1]
\Require Predictor \(f\), factual \(x^0\), threshold \(\tau\), actionable set
\(\mathcal A(x^0)\), Lipschitz constants \(\{L_j\}_{j=1}^d\), iteration limit \(T\), directional-selection capacity \(K_{\mathrm{dir}}\), and beam capacity \(K_{\mathrm{beam}}\)
\Ensure Verified target-valid, search-actionable counterfactual \(x^{\mathrm{best}}\) or \textsc{None}
\Statex Set \(d_\nu(\textsc{None},x^0)=+\infty\) for incumbent comparisons.

\State \textit{Initialize:} \(\mathcal B_0\gets\{x^0\}\), \(x^{\mathrm{best}}\gets\textsc{None}\)
\For{\(t=0,\ldots,T-1\)}
    \Statex \textit{One-directional anchors and selection for gradient evaluation}
    \State \textit{One-directional anchors:} \(\mathcal Y_t\gets\mathcal Y_t^{\mathrm{dir}}\) (Eq.~\ref{eq:directional-anchors}); \quad \(x^{\mathrm{best}}\gets\mathsf{Best}(x^{\mathrm{best}},\mathcal Y_t)\)
    \label{alg:dash-directional}
    
    \State \textit{Rank anchors for gradient evaluation:} \(\mathcal G_t\gets\mathsf R_{K_{\mathrm{dir}}}(\{y\in\mathcal Y_t\cup\mathcal B_t:\phi(y)>-\mu,\ d_\nu(y,x^0)<d_\nu(x^{\mathrm{best}},x^0)\})\) (Eqs.~\ref{eq:beam-score}, \ref{eq:diversity-rule})
    \label{alg:dash-select}
    \Statex \textit{Local gradient-based and factual-origin anchors}
    \If{\(\mathcal G_t\neq\varnothing\)}
        \State \textit{Local gradient-based anchors:} \(\mathcal Y_t\gets\mathcal Y_t\cup\mathcal Y_t^{\mathrm{loc}}\) (Eq.~\ref{eq:gradient-anchors})
        \State \textit{Factual-origin anchors:} \(\mathcal Y_t\gets\mathcal Y_t\cup\mathcal Y_t^{\mathrm{fac}}\) (Eq.~\ref{eq:tangent-anchors})
        \label{alg:dash-gradient-anchors}
    \EndIf
    \Statex \textit{Beam update with target-invalid anchors that are closer than incumbent}
    \State \(x^{\mathrm{best}}\gets\mathsf{Best}(x^{\mathrm{best}},\mathcal Y_t)\)
    \If{\(\mathcal G_t=\varnothing\)}
        \State \textbf{break}
    \EndIf
    \State \textit{Rank next beam:} \(\mathcal B_{t+1}\gets\mathsf R_{K_{\mathrm{beam}}}(\{y\in\mathcal Y_t:\phi(y)>-\mu,\ d_\nu(y,x^0)<d_\nu(x^{\mathrm{best}},x^0)\})\) (Eqs.~\ref{eq:beam-score}, \ref{eq:diversity-rule})
    \label{alg:dash-beam}
    \If{\(\mathcal B_{t+1}=\varnothing\)}
        \State \textbf{break}
    \EndIf
\EndFor
\Statex \textit{Final polishing and bisection}
\If{\(x^{\mathrm{best}}\neq\textsc{None}\)}
    \State \(\widetilde x\gets\mathsf{Polish}(x^0,x^{\mathrm{best}})\) (Subsection~\ref{subsec:prac-eff})
    \If{\(\widetilde x\in\mathcal A(x^0)\), \(\phi(\widetilde x)\leq-\mu\), and \(d_\nu(\widetilde x,x^0)<d_\nu(x^{\mathrm{best}},x^0)\)}
        \State \textit{Accept verified improvement:} \(x^{\mathrm{best}}\gets\widetilde x\)
    \EndIf
\EndIf
\State \Return \(x^{\mathrm{best}}\)
\end{algorithmic}
\end{algorithm}

\subsection{Anchor ranking}
\label{subsec:cand-rank}

After evaluating a target-invalid anchor \(x^\ell\), \DASH estimates the total distance of a continuation through it using the beam ranking
\begin{equation}
\label{eq:beam-score}
q_{\mathrm{beam}}(x^\ell)
=d_\nu(x^\ell,x^0)+\frac{\Delta(x^\ell)}{\kappa(x^\ell)+\epsilon},
\qquad
\kappa(x^\ell)=
\frac{[\phi(x^0)-\phi(x^\ell)]_+}{d_\nu(x^\ell,x^0)+\epsilon}.
\end{equation}
Here, \(\kappa(x^\ell)\) is the observed target improvement per unit distance from \(x^0\). The first term is the distance already incurred, while the second estimates the additional distance required if the observed rate persists.

At iteration \(t\), the evaluated target-invalid anchors are considered in increasing order of \(q_{\mathrm{beam}}\) to choose at most \(K_{\mathrm{beam}}\) anchors. A diversity term is also applied to choose a diverse set of regions.
\begin{equation}
\label{eq:diversity-rule}
\min_{z\in\mathcal S}d_\nu(x^\ell,z)\geq\delta_{\mathrm{div}},
\end{equation}
where \(\mathcal S\) is the set already selected. Applied with capacity \(K_{\mathrm{dir}}\), this rule selects the anchors expanded by the gradient-based search in
the current iteration. Applied with capacity \(K_{\mathrm{beam}}\), it constructs the beam carried to
the next iteration. The score concentrates computation on anchors with favorable
estimated distance. In Algorithm~\ref{alg:dash}, \(\mathsf R_K\) applies this rule with capacity \(K\).

\subsection{Efficiency Enhancements}
\label{subsec:prac-eff}
\DASH has CPU and GPU implementations. Both evaluate anchors and compute gradients in batches. The CPU implementation screens one-directional anchors before network evaluation using the directional score \(q_{\mathrm{dir}}\) (Appendix~\ref{app:screening}). The GPU implementation omits this screening and performs anchor generation, ranking, network evaluation, gradient computation, and beam selection on the GPU device. Finite step-size multiplier sets, support-size sets, and beam capacity control time and memory. The exact numerical settings and tie-breaking rules are reported in Appendix~\ref{app:dash-settings} and Appendix~\ref{app:screening}. Additionally, if a target-valid anchor has been found when the search terminates, \DASH can reduce its distance by bisection over the continuous features while holding its categorical assignments fixed. We denote this operation by \(\mathsf{Polish}(x^0,x^{\mathrm{best}})\). The polished anchor is accepted only after validity verification. 

\section{Experiments and Results}
\label{sec:experiments}

\subsection{Experimental Setup}

\paragraph{Datasets and predictors.}
Our primary testbed is actionable tabular counterfactual search, where continuous and categorical features coexist. We evaluate HELOC, Give Me Some Credit (GMC), Adult Income, and COMPAS \citep{pawelczyk2021carla} using fully connected ReLU networks of width \(64\) and depths \(2,4,8,16,32\). We also evaluate PBMC3k \citep{tenx2017pbmc3k} to test the same search on a wider network. MNIST is used to test a no-actionable dataset, whose results are reported in Appendix~\ref{app:mnist-results}. For each of three training seeds, we select \(50\) correctly classified held-out factual inputs and seek the opposite class with verification margin \(\mu=10^{-4}\). Each method receives the same factual inputs under each objective norm. This yields \(9{,}000\) tabular and \(450\) PBMC3k queries per method. Data preparation and query selection are described in Appendices~\ref{app:datasets} and~\ref{app:predictors-and-queries}.

\paragraph{Baselines and budgets.}
We compare \DASH with Wachter~\citep{Wachter2017CounterfactualEW}, DiCE~\citep{mothilal2020dice}, Greedy~\citep{Schut2021GeneratingIC}, NICE~\citep{brughmans2024nice}, FMN~\citep{pintor2021fast}, FAB~\citep{croce2020fab}, DeepFool~\citep{MoosaviDezfooli2015DeepFoolAS}, norm-extended versions of Growing Spheres (GS)~\citep{laugel2018sphere} and PGD-bisect~\citep{madry2018towards}, and with the MIP-OBJ and MIP-EXP exact formulations~\citep{mohammadi2021scaling} on tabular networks of depths \(2,4,8\) due to the heavy computational overhead. Per-query limits are \(60\) seconds for tabular and PBMC3k. The MNIST budget is reported in Appendix~\ref{app:mnist-results}. CPU experiments used AMD EPYC 7643; GPU experiments used one NVIDIA A100-40gb. Further hardware details, baseline settings, and common verification are given in Appendices~\ref{app:predictors-and-queries} and~\ref{app:baseline-details}.

\paragraph{Constraints and output verification.}
Tabular queries use dataset-specific feature bounds, immutability, permitted change directions, and categorical constraints. PBMC3k additionally requires increases of at least \(0.05\) units in two markers from a prespecified eight-gene pool. Outputs are then verified. The principal DASH search uses \(T=10\), beam capacity \(K_{\mathrm{beam}}=16\), and directional-selection capacity \(K_{\mathrm{dir}}=16\); other settings and detailed constraints are given in Appendices~\ref{app:dash-settings}, \ref{app:actionability}, and~\ref{app:output-completion}.

\begin{figure*}[t]
  \centering
  \includegraphics[width=\textwidth]{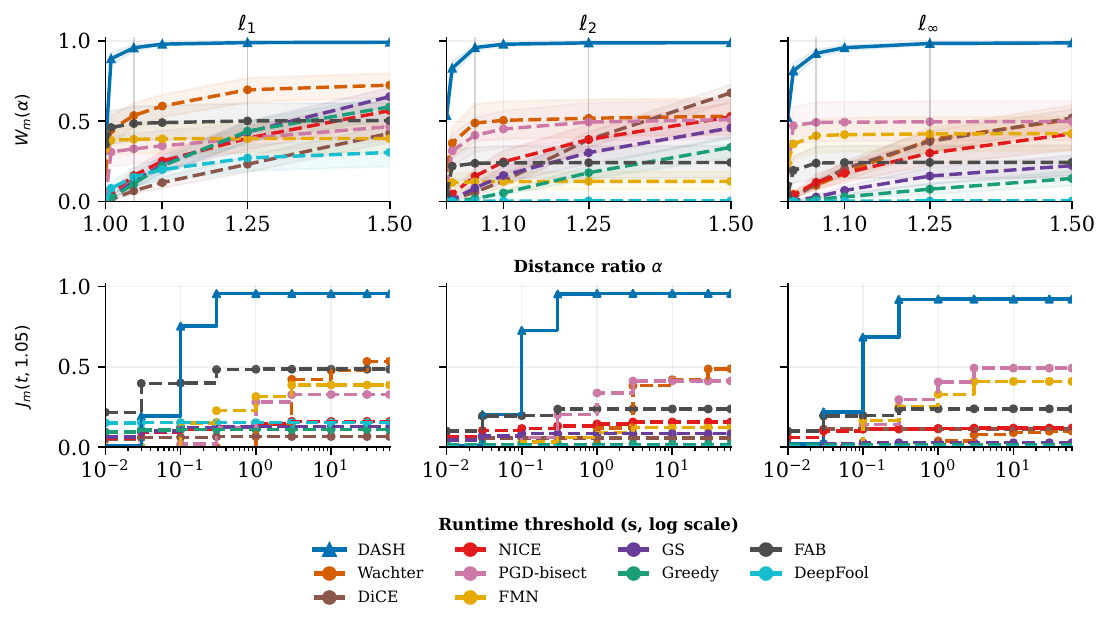}
  \caption{\textbf{Tabular performance profiles.} Columns show the \(\ell_1\), \(\ell_2\), and \(\ell_\infty\) objectives; rows show \(W_m(\alpha)\) and \(J_m(t,1.05)\). Results pool four datasets, five network depths, and three training seeds. Top-row shading gives \(95\%\) bootstrap intervals clustered by trained model (dataset $\times$ depth $\times$ seed).}
  \label{fig:tabular-primary}
\end{figure*}

\paragraph{Metrics}
Let \(m\in\mathcal M\) index \DASH and the heuristic baselines available for the corresponding queries, and let \(q\in\mathcal Q\) index the factual-predictor-norm queries. Let \(x_{q,m}\) and \(t_{q,m}\) denote method \(m\)'s verified output and recorded search time on query \(q\), respectively. Define
\begin{equation}
\label{eq:distance-ratio}
d_q^{\mathrm{best}}=
\min_{\substack{m\in\mathcal M\\x_{q,m}\ \mathrm{valid}}}
d_\nu(x_{q,m},x_q^0),
\qquad
R_{q,m}=
\begin{cases}
d_\nu(x_{q,m},x_q^0)/d_q^{\mathrm{best}}, & x_{q,m}\ \mathrm{valid},\\
+\infty, & \text{otherwise}.
\end{cases}
\end{equation}
The reference \(d_q^{\mathrm{best}}\) is the best heuristic-observed valid distance, not a certified optimum; the MIP formulations are compared with certified optima in Section~\ref{subsec:results-depth}. Thus, \(R_{q,m}\) measures the distance achieved by method \(m\) relative to this reference for query \(q\), with \(R_{q,m}=1\) indicating the best heuristic-observed performance. Our distance and joint quality-runtime profiles are
\begin{equation}
\label{eq:performance-profiles}
W_m(\alpha)=\frac{1}{|\mathcal Q|}\sum_{q\in\mathcal Q}
\mathbf 1[R_{q,m}\leq\alpha],
\qquad
J_m(t,\alpha)=\frac{1}{|\mathcal Q|}\sum_{q\in\mathcal Q}
\mathbf 1[R_{q,m}\leq\alpha\land t_{q,m}\leq t].
\end{equation}
Queries for which a method returns an invalid output or no output remain in the denominators and therefore count as failures. Thus, \(W_m(1.05)\) is the fraction of queries for which method \(m\) returns a valid counterfactual within \(5\%\) of the best heuristic-observed valid distance, while \(J_m(t,1.05)\) additionally requires the search to finish within time \(t\). These profiles follow the relative-performance comparison of \citet{dolan2002performance}.

\subsection{Results}
\label{sec:results}

\paragraph{Tabular counterfactuals}
\label{subsec:results-tabular-quality}
Figure~\ref{fig:tabular-primary} separates distance and joint quality-runtime performance by objective norm. DASH's distance profile lies above the comparator profiles for every \(\alpha\geq1.01\) under each norm. At \(\alpha=1.05\), it covers \(95.7\%\), \(95.9\%\), and \(92.3\%\) of tabular queries under \(\ell_1\), \(\ell_2\), and \(\ell_\infty\), respectively. Wachter has the second-highest coverage under \(\ell_1\) and \(\ell_2\), whereas PGD-bisect does under \(\ell_\infty\). The runtime row shows that the separation persists when unsuccessful queries and search time are counted jointly: DASH leads every comparator from \(0.1\)\,s on and reaches a higher plateau under each norm. The other baselines ordering changes with the norm, especially as PGD-bisect and FMN improve under \(\ell_\infty\).

\begin{figure*}[t]
  \centering
  \includegraphics[width=\textwidth]{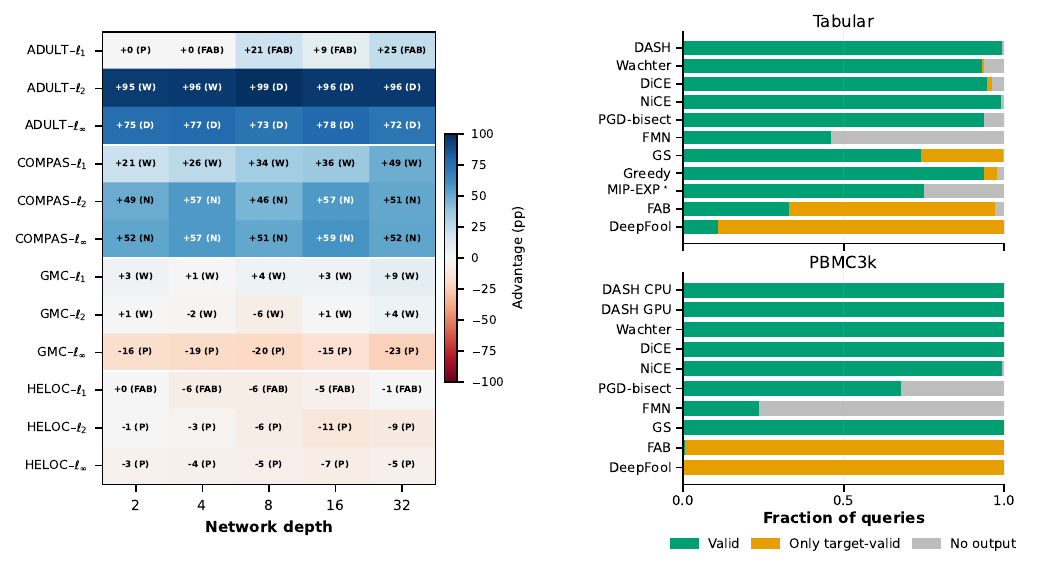}
  \caption{\textbf{Tabular robustness and output outcomes.} Left: Percentage-point difference in \(W_m(1.05)\) between \DASH and the cell-wise best prior heuristic. Parentheses identify that heuristic (W: Wachter, D: DiCE, N: NICE, P: PGD-bisect; FAB is written in full). Right: Final verification outcomes for the methods in the outcome analysis on tabular and PBMC3k queries, pooled across norms. ``Only target-valid'' denotes a target-valid but constraint-invalid output. MIP-EXP$^\star$ covers tabular depths \(2\), \(4\), and \(8\) only.}
  \label{fig:tabular-robustness}
\end{figure*}

\paragraph{Depth and certified distance}
\label{subsec:results-depth}
The left side of Figure~\ref{fig:tabular-robustness} resolves the pooled tabular comparison by dataset, norm, and depth. DASH has a positive valid-within-\(5\%\) advantage over the strongest comparator in \(36\) of \(60\) configurations, including every COMPAS cell and every Adult cell under \(\ell_2\) and \(\ell_\infty\), and ties in three. The reversals are GMC under \(\ell_\infty\) and HELOC, where the norm-adapted PGD-bisect (\(\ell_2\), \(\ell_\infty\)) and FAB (\(\ell_1\), depths 4-32) have higher coverage, and two GMC \(\ell_2\) cells, where Wachter does. The heatmap shows no shared monotonic decline in the DASH advantage as depth increases.

\begin{figure*}[t]
  \centering
  \includegraphics[width=\textwidth]{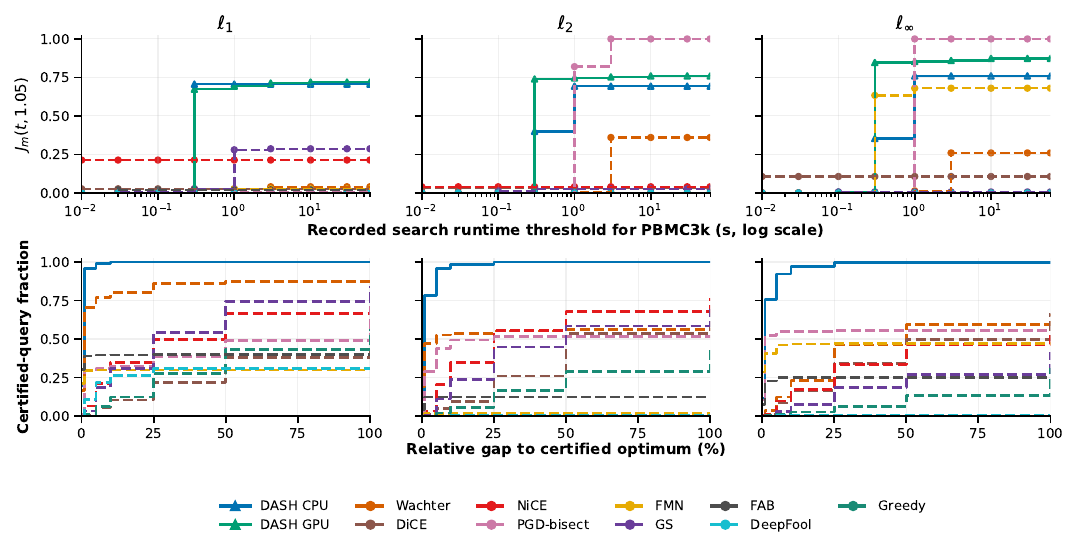}
  \caption{\textbf{PBMC3k runtime and tabular certified distance.} Columns show the \(\ell_1\), \(\ell_2\), and \(\ell_\infty\) objectives. The upper row reports PBMC3k joint quality-search-runtime profiles after the same output-only marker completion for every method; recorded times exclude completion. The lower row reports the fraction of tabular queries with a valid output within each relative gap to a certified optimum. Certification covers depths \(2\), \(4\), and \(8\); unsuccessful returns remain in the denominator.}
  \label{fig:high-dimensional-primary}
\end{figure*}

The bottom row of Figure~\ref{fig:high-dimensional-primary} compares outputs with certified optima on the subset of tabular queries for which an exact method finishes within the time limit with a certified optimum. DASH returns a valid output on all \(1{,}071\), \(1{,}022\), and \(1{,}355\) certified queries under \(\ell_1\), \(\ell_2\), and \(\ell_\infty\), respectively; \(99.0\%\), \(95.9\%\), and \(92.3\%\) are within \(5\%\) of the corresponding certified optimum. These fractions include all certified queries in their denominators rather than conditioning on a method's successful returns. Appendix~\ref{app:certified-gap} defines the plotted gap.

\paragraph{Wider network}
\label{subsec:results-high-dimensional}
The upper row of Figure~\ref{fig:high-dimensional-primary} tests whether the tabular search behavior carries to the wider PBMC3k network. The corresponding MNIST comparison appears in Appendix~\ref{app:mnist-results}. On PBMC3k, all methods receive the same returned-output marker completion before validity and distance are scored; Greedy did not run at this scale. DASH CPU and GPU end above the comparators under \(\ell_1\), with final valid-within-\(5\%\) coverages of \(70.7\%\) and \(72.0\%\) (Growing Spheres, the next best, reaches \(28.7\%\)). Under \(\ell_2\) and \(\ell_\infty\), PGD-bisect reaches \(100\%\), above DASH GPU (\(76.0\%\) and \(87.3\%\)) and CPU (\(69.3\%\) and \(76.0\%\)). The three norms therefore do not exhibit the same method ordering in this high-dimensional setting.

\paragraph{Failure modes.} The right side of Figure~\ref{fig:tabular-robustness} separates invalid returns from queries with no output. On tabular queries, PGD-bisect returns valid counterfactuals in \(93.6\%\) of cases despite meeting the within-\(5\%\) criterion in \(41.1\%\); FMN more often returns no output, while FAB and DeepFool predominantly return target-valid but constraint-invalid points, including on PBMC3k after shared completion. This figure illustrates a major difference between CFE methods, built mostly for tabular recourse, and adversarial attack methods, built mostly for high dimensional datasets with simple bound constraints.

\subsection{Conclusion}

\paragraph{Discussion.} The results distinguish finding a target-valid point from finding a close counterfactual that satisfies the specified actionability constraints. Across the tabular tasks that contain both continuous and categorical variables, \DASH combines high valid-within-\(5\%\) coverage with short search times under all three norms. Its proximity is also supported on the subset where an exact solver certifies the optimum. These findings support boundary-guided, actionable search as a practical approach to tabular counterfactual generation, especially when categorical variables coexist with continuous variables. For continuous only datasets, like PBMC3k, \DASH outperforms all methods on \(\ell_1\) and remains competitive otherwise.

\paragraph{Limitations and future work.}
\DASH does not certify global optimality or non-existence: a finite beam and runtime budget can cause it to miss an existing counterfactual. Its search also relies on local gradient and boundary estimates, and the current implementation is limited to ReLU networks and the actionability constraints considered here. Future work could combine DASH's incumbent with certified lower bounds and extend the search to causal, user-specific, or more general linear constraints. On some continuous-only settings and norms, adversarial attack methods outperform the current configuration; adapting the balance between one-directional and gradient-based anchors may improve performance in those settings.


\clearpage
\section*{AI Use Statement}
In this work, generative AI tools assisted with code implementation, figure preparation, and manuscript polishing. The code was reviewed and run through multiple tests to ensure correctness. The figure and plots were manually verified with final results. It was not used for other tasks with required disclosures. The authors take responsibility for the final content, including all AI-assisted claims and artifacts.

\section*{Reproducibility Statement}
Algorithm~\ref{alg:dash} specifies the search. The appendix provides proofs, implementation settings, dataset preprocessing and actionability rules, predictor training and query selection, baseline configurations, and evaluation procedures (Appendices~\ref{app:method-properties}-\ref{app:baseline-details}).

\bibliography{reference,baseline_norm_adaptation_references}
\bibliographystyle{copyright}

\pagebreak

\appendix

\section{DASH implementation details}

\subsection{Target orientation and output scores}
\label{app:target-orientation}

The binary target may lie on either side of a scalar decision threshold. Let
\(f_{\mathrm{raw}}\) and \(\tau_{\mathrm{raw}}\) denote the original predictor and
threshold. Set \(s=1\) when the target is
\(f_{\mathrm{raw}}(x)\leq\tau_{\mathrm{raw}}-\mu\) and \(s=-1\) when it is
\(f_{\mathrm{raw}}(x)\geq\tau_{\mathrm{raw}}+\mu\). The main text uses the
oriented quantities \(f=s f_{\mathrm{raw}}\) and
\(\tau=s\tau_{\mathrm{raw}}\). Hence
\(\phi(x)=f(x)-\tau\leq-\mu\) in either case, and
\(g(x)=-\nabla\phi(x)\) points toward decreasing oriented output. This
sign change is without loss of generality for a scalar threshold target; it
does not by itself turn an arbitrary vector target into a scalar one.

For a \(K\)-class predictor with pre-softmax logits
\(z(x)\in\mathbb R^K\), a specified target class \(c\) has the scalar
margin
\begin{equation}
\label{eq:multiclass-target-score}
\phi_c(x)=\max_{j\ne c}\{z_j(x)-z_c(x)\}.
\end{equation}
The condition \(\phi_c(x)\leq-\mu\) says that class \(c\) exceeds every
other logit by at least \(\mu\). If the target is any class other than the
factual class \(y\), use
\(\phi_{\ne y}(x)=z_y(x)-\max_{j\ne y}z_j(x)\); its value is at most
\(-\mu\) exactly when some competing class exceeds \(y\) by that margin.
At \(\mu=0\), ties must also be resolved according to the classifier's
label rule during final verification. Logit-margin objectives of this
form are used for targeted multiclass attacks~\citep{carlini2017towards}.

For a vector regressor \(F(x)\in\mathbb R^p\) and a desired output box
\(\underline{y}_i\leq F_i(x)\leq\overline{y}_i\) for all \(i\), define
\begin{equation}
\label{eq:vector-regression-target-score}
\phi_{\mathrm{box}}(x)=
\max_i\{\underline{y}_i-F_i(x),\ F_i(x)-\overline{y}_i\}.
\end{equation}
Then \(\phi_{\mathrm{box}}(x)\leq-\mu\) is equivalent to
\(\underline{y}_i+\mu\leq F_i(x)\leq\overline{y}_i-\mu\) for every output. With \(\mu=0\),
this recovers the original box. For a nonempty tightened box, each interval
must have width at least \(2\mu\). Output coordinates should be scaled
before applying a common margin.
Desired output intervals across a forecast horizon have been studied in
counterfactual forecasting~\citep{wang2023forecastcf}. More generally, a
polyhedral target \(\{y:Ay\leq b\}\) has score
\(\phi(x)=\max_r\{(AF(x))_r-b_r\}\), after scaling its inequalities when
a shared verification margin is used.

These reductions preserve the quantities used by the search. Because
\(|\max_r h_r(x+h e_j)-\max_r h_r(x)|\leq
\max_r|h_r(x+h e_j)-h_r(x)|\), if every constituent score \(h_r\)
has coordinate bound \(L_{rj}\), then a finite maximum or minimum has bound
\(L_j=\max_r L_{rj}\). For a vector-output ReLU network, the
\((i,j)\) entry of \(|W_D|\cdots|W_1|\) is a valid bound
\(L^F_{ij}\) for output \(i\) along input feature \(j\), by the same
chain-rule argument used for the scalar network in
Appendix~\ref{app:screening}. A logit difference
\(z_k-z_c\) has a coordinate bound no larger than the sum of the bounds
for \(z_k\) and \(z_c\). For the polyhedral score above, a valid bound
for constituent \(r\) is \(L_{rj}\leq\sum_i|A_{ri}|L^F_{ij}\), where
\(L^F_{ij}\) bounds the change in output \(F_i\) along input feature \(j\).
ReLU logits and ReLU regression outputs make these scores piecewise
affine~\citep{berzins2023polyhedral}.
At a tie, an active branch can supply a search direction, but
Proposition~\ref{prop:affine-region-exactness} assumes that \(\phi\) is
differentiable at the anchor and applies only while the network activation
pattern and active score term remain fixed along the segment. The current implementation and
experiments use scalar-output binary predictors. Applying these reductions
to vector-output networks requires supplying the scalar score, its
gradient, and its directional bounds; arbitrary nonlinear output targets
do not inherit the piecewise-affine exactness claim.

\subsection{Directional scales and anchor screening}
\label{app:screening}

For a scalar-output Dense/ReLU network with weight matrices
\(W_1,\ldots,W_D\), the row vector
\(L=|W_D|\cdots|W_1|\) gives directional bounds \(L_j\).
Each ReLU slope lies in \([0,1]\), so the chain rule shows that this
activation-agnostic product bounds the output change under a displacement
of feature \(j\) as in Eq.~\ref{eq:directional-lipschitz}. Related
coordinate-wise bound propagation is studied by \citet{weng2018towards}. For
\(L_j>0\), \DASH uses
\(\rho_j(x)=\Delta(x)/(L_j+\epsilon)\) and the multiplier and endpoint
anchors of Eq.~\ref{eq:directional-anchors}. A direction with
\(L_j=0\) is skipped. The bounds supply anchor scales; they do not certify
that a proposed anchor reaches the target.

\paragraph{Proof of Proposition~\ref{prop:one-feature-exclusion}.} Let \(x+h e_j\) be target-valid while \(x\) is target-invalid. Then \(|\phi(x+h e_j)-\phi(x)|\geq\phi(x)+\mu=\Delta(x)\), so Eq.~\ref{eq:directional-lipschitz} gives \(L_j|h|\geq\Delta(x)\).

For a one-directional anchor \(x^\ell\) generated from \(x^k\), the CPU screening score is
\begin{equation}
\label{eq:directional-screening}
q_{\mathrm{dir}}(x^k,x^\ell)=
\frac{d_\nu(x^\ell,x^k)}
{\max\{g(x^k)^\top(x^\ell-x^k),\epsilon\}}.
\end{equation}
The numerator is the movement from \(x^k\), and the denominator is the target
improvement predicted by its local affine model. A smaller score indicates
more predicted progress per unit movement. The CPU implementation associates
each new anchor with the anchor that generated
it when computing \(q_{\mathrm{dir}}\). If multiple parents generate the same
point, score ordering and duplicate detection determine which anchor is
admitted first. Screening preserves the lowest-scoring one-directional
anchor for each feature-and-sign tag. An explicitly added feature-bound
endpoint has a separate tag. Among the remaining continuous anchors, it
retains the lowest scores until the total reaches
\[
\max\!\left\{n_{\mathrm{protected}},
1,\left\lceil \gamma N_{\mathrm{cont}}\right\rceil\right\},
\]
where \(N_{\mathrm{cont}}\) counts continuous one-directional anchors and
\(n_{\mathrm{protected}}\) counts protected tags. Permitted categorical
transitions are retained separately. This coverage floor applies to
one-directional anchors; gradient-based anchors are ranked by the same score
without feature-and-sign protection. Factual-origin anchors are evaluated without
this CPU screening step. The final tabular CPU suite sets \(\gamma=1\).
The batched GPU implementation does not use the CPU directional-screening filter.

For beam selection, anchors are ordered by \(q_{\mathrm{beam}}\) from Eqn.~\ref{eq:beam-score}. An anchor is accepted when it is at least \(\delta_{\mathrm{div}}\) from every previously accepted anchor. If fewer than the requested number of anchors meet this separation criterion, the remaining positions are filled by the lowest-scoring rejected anchors. Previously evaluated anchors are identified using a coordinate hash with tolerance \(10^{-10}\).

\subsection{Actionability enforcement}
\label{app:actionability-implementation}

For each continuous feature \(j\), let
\(I_j(x^0)=[a_j(x^0),b_j(x^0)]\) combine its input bounds, immutability,
and permitted direction of change. An immutable feature has
\(a_j(x^0)=b_j(x^0)=x_j^0\). For a generated anchor \(z\), the enforcement map is
\[
[\Pi_{\mathrm{act}}(z)]_j=\min\{b_j(x^0),\max\{a_j(x^0),z_j\}\}.
\]
Categorical anchors are generated as permitted one-hot vertices;
assignments outside that set are discarded. The map therefore clips
feature-wise constraints and preserves valid categorical assignments. It is
not a metric projection onto a general mixed-integer actionable set. The
PBMC3k marker-group criterion is applied to returned outputs as described in
Appendix~\ref{app:output-completion}.

\subsection{Proofs of method properties}
\label{app:method-properties}

\paragraph{Proof of Proposition~\ref{prop:affine-region-exactness}.}
Write \(\delta=\Delta(x^k)=\phi(x^k)+\mu>0\) and \(g=g(x^k)\).
The directions in Eq.~\ref{eq:norm-directions} satisfy
\(\|v_\nu(g)\|_\nu=1\) and \(g^\top v_\nu(g)=\|g\|_{\nu^\ast}\).
On the assumed segment the affine model is exact, so
\(\phi(x_\nu^\star)=\phi(x^k)-g^\top(x_\nu^\star-x^k)
=\phi(x^k)-\delta=-\mu\).
Any \(z\) in the affine target half-space satisfies
\(g^\top(z-x^k)\geq\delta\). H\"older's inequality gives
\(\|z-x^k\|_\nu\geq\delta/\|g\|_{\nu^\ast}\), and
\(x_\nu^\star\) attains this bound. The minimizer need not be unique.

\paragraph{Proof of Proposition~\ref{prop:actionability-preservation}.}
The factual anchor \(x^0\) satisfies the search-time feature constraints.
Continuous clipping maps every generated feature value into its permitted interval,
while categorical transitions replace a valid one-hot vertex with another
permitted vertex. Thus every admitted anchor belongs to
\(\mathcal A(x^0)\). Each later beam is drawn from evaluated anchors, so
induction over beam iterations proves the claim.

\subsection{Search configuration and final verification}
\label{app:dash-settings}

Table~\ref{tab:app-dash-settings} reports the CPU search settings for the
final suite. The plotted high-dimensional batched GPU runs likewise use
\(K_{\mathrm{beam}}=K_{\mathrm{dir}}=16\), \(T=10\), and the wide gradient-step multipliers, but
do not apply the CPU directional-screening filter. Step multipliers probe shorter and longer movements than
the local boundary estimate. For each configured support size, the
gradient construction retains the actionable features with largest absolute
gradient components. A zero usable gradient produces no local gradient-based or factual-origin
anchor; a factual-origin ray is also skipped when its estimated intersection
radius is nonpositive or undefined. All returned anchors undergo the
target-validity and search-time actionability checks of
Eq.~\ref{eqn:cf-problem}. If a valid incumbent exists, \DASH performs at
most 20 bisection steps on its continuous features toward \(x^0\), holding
the incumbent's categorical assignments fixed. The polished result is kept
only if it remains target-valid and actionable. A returned anchor is valid
under these checks, while \textsc{None} does not establish infeasibility.

\begin{table}[t]
\centering
\small
\caption{DASH search settings.}
\label{tab:app-dash-settings}
\begin{tabular}{lr}
\toprule
Setting & Value \\
\midrule
Beam capacity \(K_{\mathrm{beam}}\) & \(16\) \\
Directional-selection capacity \(K_{\mathrm{dir}}\) & \(16\) \\
Iteration limit \(T\) & \(10\) \\
Verification margin \(\mu\) & \(10^{-4}\) \\
CPU tabular directional-screening fraction \(\gamma\) & \(1\) \\
Axis multipliers \(\Lambda_{\mathrm{ax}}\) & \(\{1,2,4,8,16,32\}\) \\
Gradient-step multipliers \(\Lambda_{\mathrm{gr}}\) &
\(\{0.125,0.25,0.5,1,2,4,8,16,32\}\) \\
Gradient support sizes & \(\{1,2,4,8,16\}\) \\
Factual-origin multipliers \(\Lambda_{\mathrm{fac}}\) &
\(\{0.75,1,1.25,1.5,2\}\) \\
Numerical gradient and Lipschitz tolerance & \(10^{-12}\) \\
Beam-diversity tolerance \(\delta_{\mathrm{div}}\) & \(10^{-5}\) \\
Anchor-hash tolerance & \(10^{-10}\) \\
Continuous polishing steps & \(20\) \\
\bottomrule
\end{tabular}
\end{table}

\section{Experimental protocol}

\subsection{Datasets and preprocessing}
\label{app:datasets}

Table~\ref{tab:app-data} reports the processed input dimensions, one-hot groups, class directions, and train-test counts. Continuous tabular features are scaled to \([0,1]\) using training-set extrema. Categorical features are one-hot encoded using levels observed in the training data. GMC and HELOC missing values are imputed using training-set medians. COMPAS length of stay is derived from jail admission and release dates. MNIST pixels are scaled to \([0,1]\). PBMC3k uses 50 selected genes and training-set min-max scaling. The reported \(d_\nu\) values are \(\ell_\nu\) distances in these processed coordinates.

\begin{table}[t]
\centering
\small
\caption{Processed datasets and query directions.}
\label{tab:app-data}
\begin{tabular}{lrrll}
\toprule
Dataset & Dim. & Groups & Factual \(\rightarrow\) target & Train / test \\
\midrule
Adult & 108 & 8 & \(\leq\$50\mathrm{K}\rightarrow>\$50\mathrm{K}\) & 30,162 / 15,060 \\
COMPAS & 13 & 3 & recidivated \(\rightarrow\) not recidivated & 4,938 / 1,234 \\
GMC & 10 & 0 & delinquent \(\rightarrow\) not delinquent & 120,000 / 30,000 \\
HELOC & 23 & 0 & Bad \(\rightarrow\) Good & 7,897 / 1,974 \\
MNIST \(3,8\) & 784 & 0 & \(3\rightarrow8\) & 11,982 / 1,984 \\
PBMC3k & 50 & 0 & other \(\rightarrow\) CD14+ monocyte & 2,110 / 528 \\
\bottomrule
\end{tabular}
\end{table}

\subsection{Dataset-specific actionability specifications}
\label{app:actionability}

All returned outputs are first required to satisfy the feature bounds induced by preprocessing. Table~\ref{tab:app-actionability} specifies the additional dataset-specific restrictions. Immutable features remain equal to their factual values, nondecreasing and nonincreasing features may change only in the indicated direction, and categorical features must remain valid one-hot assignments.

\begin{table*}[t]
\centering
\small
\caption{Actionability specifications used in the experiments.}
\label{tab:app-actionability}
\begin{tabular}{p{0.14\textwidth} p{0.23\textwidth} p{0.25\textwidth} p{0.26\textwidth}}
\toprule
Dataset & Immutable features & Directional constraints & Additional constraints \\
\midrule
Adult &
Age, fnlwgt, race, sex, and native-country features &
Education-num is nondecreasing &
Categorical groups retain valid one-hot assignments \\

COMPAS &
None beyond feature bounds &
Age and priors-count are nondecreasing &
Categorical groups retain valid one-hot assignments \\

GMC &
Age, NumberOf\-Time30-59\-Days\-Past\-Due\-Not\-Worse, NumberOf\-Time60-89\-Days\-Past\-Due\-Not\-Worse, NumberOf\-Times90\-Days\-Late, and NumberOf\-Dependents &
Revolving\-Utilization\-Of\-Unsecured\-Lines and DebtRatio are nonincreasing. MonthlyIncome is nondecreasing &
None \\

HELOC &
None beyond feature bounds &
ExternalRiskEstimate is nondecreasing. NetFractionRevolvingBurden and NetFractionInstallBurden are nonincreasing &
None \\

MNIST &
None beyond pixel bounds &
None &
Every pixel remains in \([0,1]\) \\

PBMC3k &
All genes outside the marker panels are immutable &
CD14, LYZ, FCN1, LST1, CTSS, TYROBP, S100A8, and S100A9 are nondecreasing. CD3D, CD3E, LCK, CD2, CD7, IL7R, LTB, and PTPRCAP are nonincreasing &
At least two monocyte-marker genes must increase by at least \(0.05\) \\
\bottomrule
\end{tabular}
\end{table*}

The PBMC3k constraints define a marker-based computational intervention model. They restrict the search to expression changes consistent with an increased monocyte-marker signal and a reduced T-cell-marker signal. They do not establish that the corresponding expression changes are biologically feasible interventions.

\subsection{Common PBMC3k output completion}
\label{app:output-completion}

The marker-group requirement in Table~\ref{tab:app-actionability} is an
additional PBMC3k output constraint applied equally to every method. DASH's
search-time actionability set consists of feature bounds, immutable
features, permitted directions, and valid categorical assignments. After a
method returns an output \(x\), the common final check verifies target
validity and all these base constraints. An output that passes is retained.
If \(x\) reaches the target and satisfies the base constraints but fails the
marker-group condition, we enumerate the \(\binom{8}{2}=28\) unordered
pairs \((i,j)\) of monocyte markers. For each pair, we set
\(x_i\leftarrow\max\{x_i,x_i^0+0.05\}\) and
\(x_j\leftarrow\max\{x_j,x_j^0+0.05\}\), clip the result to the processed
feature bounds, and reevaluate the predictor and every actionability
condition. Among passing completed outputs, we keep the one with smallest
\(d_\nu(\cdot,x^0)\). If none passes, that method has no valid
output for the query. This completion is applied once to the returned
output of each method, with the same selection rule and verification. The
reported search time excludes completion time. The saved benchmark records
do not isolate completion wall-clock time, so we do not report a numerical
completion-time summary.

\subsection{Predictors, queries, and compute budgets}
\label{app:predictors-and-queries}

Tabular predictors are fully connected ReLU networks of width \(64\) and depths \(2,4,8,16,\) and \(32\). MNIST uses a width-\(256\), eight-layer ReLU network, and PBMC3k uses a width-\(128\), eight-layer ReLU network. Models are trained with Adam~\citep{Kingma2014AdamAM}, learning rate \(10^{-3}\), weight decay \(10^{-4}\), and at most 1,000 epochs. Early stopping uses a 20\% validation partition of the training data and patience 100. Tabular models use full-batch training, while MNIST and PBMC3k use minibatches of 512 and 256, respectively.

We train each architecture using seeds \(17\), \(29\), and \(43\). For each trained predictor, we select 50 correctly classified held-out factual inputs and seek the opposite prediction under each of the three norms. All methods receive the same factual inputs for a given predictor and norm. The per-query limit is 60 seconds for tabular and PBMC3k experiments and 360 seconds for MNIST experiments. Every CPU heuristic runs on one thread per query; the exact MIP baselines use four Gurobi threads.

The CPU host has one AMD EPYC 7643 processor, two hardware threads per core, and 994\,GiB of memory. CPU-run metadata records Julia 1.12.1. The GPU host provides an NVIDIA A100 with 40\,GiB of memory, driver 590.48.01, and Julia 1.11.4. 

\subsection{Baseline configurations and common evaluation}
\label{app:baseline-details}

Table~\ref{tab:app-baselines} reports the implemented baseline settings. Methods receive the same coordinate bounds, except FAB, which is clipped only to the feature range, and DeepFool, which is unconstrained; neither receives the actionability constraints. Methods differ in whether they enforce one-hot assignments during search. Consequently, every returned output is evaluated using the common target-validity and actionability checks before inclusion in validity or proximity metrics.

\begin{table}[t]
\centering
\small
\caption{Baseline implementation settings.}
\label{tab:app-baselines}
\begin{tabular}{@{}l p{0.70\textwidth}@{}}
\toprule
Method & Configuration \\
\midrule
Wachter & 10 restarts, learning rate \(10^{-3}\), 100 inner steps, \(\lambda\)-growth 5 \\
DiCE & proximity weight \(0.5\), Adam~\citep{Kingma2014AdamAM} learning rate \(0.05\), at most 5,000 iterations \\
NICE & target-side training pool, norm-specific nearest neighbor, grouped one-hot copying \\
Growing Spheres & 1,000 samples per shell, shell width \(0.2\), 1,000 shells \\
PGD-bisect & 10 restarts, 100 gradient steps, bisection tolerance \(10^{-4}\) \\
FMN & 1,000 steps, \(\alpha:1\rightarrow0.01\), \(\gamma:0.05\rightarrow0.001\) \\
Greedy & coordinate step \(0.1\), at most 10,000 steps \\
FAB & Norm-specific box-constrained projections; 100 iterations, \(\alpha_{\max}=0.1\), \(\eta=1.05\), \(\beta=0.9\), 30 polish steps \\
DeepFool & norm-specific linearized steps; at most 50 iterations, overshoot \(0.02\), no box clipping \\
MIP-OBJ and MIP-EXP & tolerance \(10^{-4}\), 4 Gurobi threads \\
\bottomrule
\end{tabular}
\end{table}

Wachter and DiCE project categorical groups before evaluating output validity. NICE copies one-hot groups jointly from its target-side training instance. PGD-bisect and FMN apply categorical projection before final evaluation. Under \(\ell_2\), FMN returns no counterfactual on any of the \(750\) Adult queries; these queries count as failures. Growing Spheres and Greedy use coordinate bounds but do not consistently repair categorical assignments during their search. The mixed-integer formulations encode the applicable bounds and categorical constraints directly. MIP methods are evaluated only on tabular predictors of depths \(2\), \(4\), and \(8\).

\paragraph{FAB reimplementation.}
We use a repository reimplementation of FAB~\citep{croce2020fab}, not the
authors' released code. At each iteration, it forms a local affine decision
hyperplane and computes box-constrained projections under the requested
\(\ell_1\), \(\ell_2\), or \(\ell_\infty\) norm. The \(\ell_1\) projection
uses a continuous-knapsack solution, while the scalar parameters for the
\(\ell_2\) and \(\ell_\infty\) projections are found by bisection. It then
combines projections from the current point and factual, applies an
overshoot and backward correction, and performs a final boundary search.
We use at most 100 iterations, \(\alpha_{\max}=0.1\), overshoot
\(\eta=1.05\), backward-step factor \(\beta=0.9\), and 30 final
boundary-search steps. The
implementation enforces box bounds but does not repair categorical or other
tabular actionability constraints during its search; all outputs undergo the
common final validity checks.

\paragraph{Norm-adapted PGD-bisect.}
Our PGD-bisect baseline combines a projected-gradient search at a fixed radius with bisection over that radius, following the projected-gradient attack of \citet{madry2018towards}. For each requested \(p\in\{1,2,\infty\}\), an inner restart searches the intersection \(S_p(\epsilon)=\{x:\|x-x^0\|_p\leq\epsilon,\ \ell\leq x\leq u\}\), where \(\ell\) and \(u\) are feature-bound vectors. Its ascent direction is the signed gradient for \(p=\infty\), the Euclidean-normalized gradient for \(p=2\), and a signed coordinate of largest gradient magnitude for \(p=1\). Each iteration proposes a step of length \(2.5\epsilon/100\) in the corresponding norm, then takes the Euclidean projection onto \(S_p(\epsilon)\). For \(p=\infty\), this projection is coordinate-wise clipping. For \(p=1\), we use the clipped soft-threshold form of the exact ball-box projection given by \citet{croce2021mindbox}; a scalar multiplier is found by bisection so that the projected perturbation has \(\ell_1\) norm at most \(\epsilon\). The \(p=2\) projection similarly clips a radially shrunk perturbation and solves for its scalar multiplier. This is a norm adaptation of our PGD-bisect baseline, not a reproduction of the adaptive \(\ell_1\)-APGD optimizer of \citet{croce2021mindbox}. We use ten restarts and an outer-radius tolerance of \(10^{-4}\). Random starts are sampled from the requested norm ball and clipped to the feature bounds; categorical projection is applied only to the returned point before common verification.

\paragraph{Norm-adapted Growing Spheres.}
The original method searches Euclidean shells and then reduces unnecessary feature changes \citep{laugel2018sphere}. We retain those two phases while replacing the shell by \(\{x:a\leq\|x-x^0\|_p\leq b\}\) for the requested \(p\in\{1,2,\infty\}\). Gaussian vectors normalized in \(\ell_2\) generate Euclidean directions, independently signed exponential variables normalized in \(\ell_1\) generate cross-polytope directions, and uniform-cube vectors normalized in \(\ell_\infty\) generate cube directions. These constructions follow the probabilistic geometry of \(\ell_p\) balls \citep{barthe2005lpball}. Conditional on a shell, the radius is sampled with \(R^d\) uniform on \([a^d,b^d]\), where \(d\) is the input dimension. We clip sampled points to the feature bounds, choose the target-valid point with smallest distance in the requested norm, and apply the same feature-reversion pass as in the Euclidean implementation. Clipping can move points inward and therefore changes the ideal uniform shell distribution near the domain boundary. The \(\ell_1\) and \(\ell_\infty\) versions are our norm adaptations, not variants supplied by the original authors. All three use 1,000 samples per shell, shell width \(0.2\), and at most 1,000 shells; outputs receive the same target and actionability checks as every other method.

\section{Supplementary analyses}

\subsection{MNIST scalability evaluation}
\label{app:mnist-results}

\begin{figure*}[t]
  \centering
  \includegraphics[width=\textwidth]{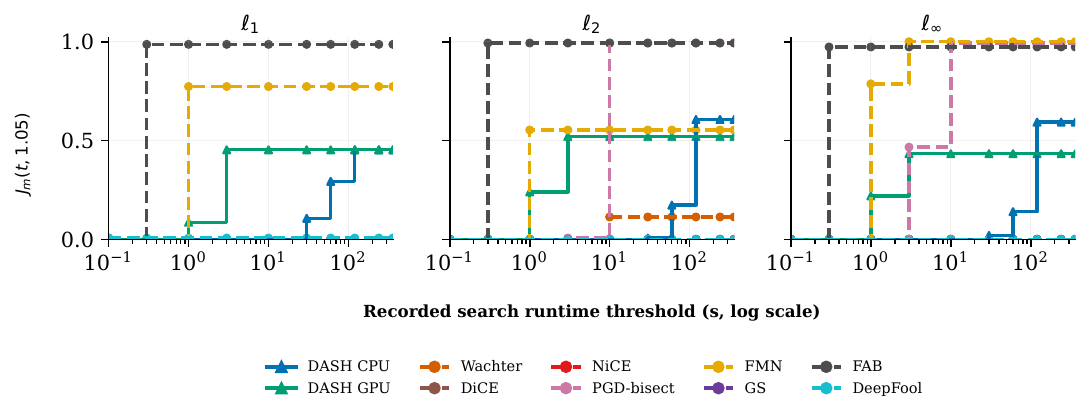}
  \caption{\textbf{MNIST joint quality-search-runtime profiles.} Columns show the \(\ell_1\), \(\ell_2\), and \(\ell_\infty\) objectives. Each curve gives the fraction of all matched queries returning a valid counterfactual within \(5\%\) of the best heuristic-observed valid distance by the indicated search-time threshold.}
  \label{fig:mnist-appendix}
\end{figure*}

We additionally evaluate the same search on MNIST~\citep{lecun2010mnist}, using a width-\(256\), eight-layer ReLU network. Pixels are scaled to \([0,1]\), and bounded pixel changes are permitted. For each of three model seeds, we select \(50\) correctly classified held-out images, yielding \(150\) factual inputs and \(450\) query-norm cases per method. The per-query limit is \(360\) seconds. Figure~\ref{fig:mnist-appendix} reports joint valid-within-\(5\%\) quality-search-runtime profiles for the CPU and GPU implementations and available baselines. At the final threshold, FAB reaches \(98.7\%\), \(99.3\%\), and \(97.3\%\) under \(\ell_1\), \(\ell_2\), and \(\ell_\infty\), respectively, while DASH CPU reaches \(45.3\%\), \(60.7\%\), and \(59.3\%\). Native \(\ell_1\) PGD-bisect and Growing Spheres runs do not cover the full matched query set and are omitted from that panel. The MNIST setting tests scaling on a larger input space without the tabular actionability constraints of the primary evaluation.

\subsection{Certified-distance gap}
\label{app:certified-gap}

For queries on which a mixed-integer formulation certifies an optimum \(d_{\nu}^{\star}\), we report the relative gap of method \(m\) as
\[
g_m =
100\left(
\frac{d_{\nu}(x_m,x^0)}{d_{\nu}^{\star}}-1
\right).
\]
A method that returns no valid counterfactual is assigned \(g_m=\infty\). The certified-gap curves therefore retain unsuccessful returns in their denominators and should be interpreted only over queries for which the exact formulation completed with an optimum certificate.


\newsavebox{\appresultsbox}
\providecommand{\adjustbox}[2]{\sbox{\appresultsbox}{#2}%
  \ifdim\wd\appresultsbox>\linewidth\resizebox{\linewidth}{!}{\usebox{\appresultsbox}}\else\usebox{\appresultsbox}\fi}

\section{Full benchmark results}
\label{app:full_results}

\paragraph{Protocol.}
Four tabular datasets (Adult, COMPAS, GMC, HELOC), fully connected ReLU networks of width 64 and depth
$\{2,4,8,16,32\}$, three model seeds (17, 29, 43), 50 held-out factuals per (dataset, depth, seed) cell,
the same factuals for every arm, and a 60\,s budget per factual on one CPU thread. Every arm except FAB and DeepFool
receives the same basic actionability box (immutable, non-decreasing and non-increasing coordinates); validity is checked
identically for every arm (see Metrics). DASH (beam 16, 10 iterations) and every baseline except Greedy* search the
objective norm of the table; Greedy* is norm-agnostic and is reported under every norm. \emph{DASH + data seed} is an
ablation that also generates anchors by copying single features, or one-hot groups, from the nearest training point already on the
target side, as NICE does. It leaves the rate of Table~\ref{tab:carla_headline} unchanged (94.6\% for both) at 11\% more
search time (median 0.068 vs 0.061\,s), so DASH is reported without it, and the ablation enters no best heuristic-observed reference.
MNIST (3 vs 8, $n=784$, depth 8, width 256, no actionability constraints, 360\,s) uses the same seeds and 50 factuals;
Greedy* did not run there, and under $\ell_1$ GS* and PGD-bisect* cover 70 of the 150 factuals. PBMC3k is reported in the main text only: its numbers there include the marker
completion of Appendix~\ref{app:output-completion}, which these tables do not apply.
The exact references MIP-EXP (bounded-distance feasibility queries with radius refinement) and MIP-OBJ
(direct distance minimization) use Gurobi~\citep{gurobi} with 4 threads, a 60\,s budget per factual and an absolute
optimality gap of $10^{-4}$; they were run at depths 2, 4 and 8 on the same factuals. In the per-cell tables their
medians are over the instances they solve, which at depth 8 is a small and easy subset, so they are left out of the
depth figures and compared through Tables~\ref{tab:carla_optimality} (certified optima) and \ref{tab:carla_mip_depth} (all queries) instead.
MIP-EXP proved four factual-model pairs infeasible (no point in the actionability box flips the label: one
COMPAS and one GMC factual at depth 2, two GMC factuals at depth 8; the same four under each norm, 12 of the
$3\times3{,}000$ queries). No arm returned a valid counterfactual on them; they are kept in every count below and count as a
failure for every arm (0.13\% of the queries), so that every denominator is the full query set.

\paragraph{Baselines marked with an asterisk.}
All baselines are our own Julia reimplementations. DiCE~\citep{mothilal2020dice} and NICE~\citep{brughmans2024nice} handle
categorical features in their original formulation and are used as published; for Wachter~\citep{Wachter2017CounterfactualEW},
whose original formulation enumerates a single binary attribute, we project categorical groups as described in
Appendix~\ref{app:baseline-details}. GS*,
PGD-bisect* and FMN* are continuous methods (Growing Spheres, \citealp{laugel2018sphere}; the PGD attack of
\citealp{madry2018towards}; the fast minimum-norm attack of \citealp{pintor2021fast}), run in the variant that searches the
objective norm of the table (Appendix~\ref{app:baseline-details}); PGD-bisect* wraps the attack in a bisection on the
perturbation radius, since PGD alone has no notion of a minimal perturbation. PGD-bisect* and FMN* receive a one-hot repair that
projects the returned point onto the nearest legal category assignment (or one alternative category per group) and re-checks the
label; GS* and Greedy* receive none. Greedy* is the single-feature saliency-map search of \citet{Schut2021GeneratingIC}, an extension of JSMA
\citep{papernot2016limitations}, which originally perturbs feature pairs, run on a single predictor rather than their ensemble. Scoring the un-repaired outputs instead changes the within-5\% rates of Table~\ref{tab:carla_headline}
by at most 0.4 points (PGD-bisect* 40.7 vs 41.1, FMN* 30.5 vs 30.7); the raw outputs are stored alongside the
repaired ones in every results file. FAB~\citep{croce2020fab} and DeepFool~\citep{MoosaviDezfooli2015DeepFoolAS} run unmodified, in
their $\ell_1$, $\ell_2$ and $\ell_\infty$ variants, and carry no asterisk: FAB is clipped to the range of each feature and DeepFool is
unconstrained; neither receives the actionability box or a one-hot repair, and the common validity check counts the resulting
violations as failures. The main text omits the asterisks.

\paragraph{Metrics.}
Success is the fraction of factuals for which the arm returns a point that crosses the threshold with
margin $10^{-4}$ (CARLA's success rate); validity additionally requires the box and one-hot constraints
to hold, and is the column every other metric conditions on. The gap between the two is the arm's
constraint handling: GS* flips the label on every Adult and COMPAS query but is valid on only 45.6\% and 52.0\% of them
(Greedy*: 91.3\% and 87.1\%), and FAB and DeepFool flip it on at least 90\% of the queries of every tabular dataset but are
valid on only 1.5-29.6\% of the Adult, COMPAS and GMC queries. Every other column is a
median over the valid rows of the cell (NA = no valid counterfactual). $\ell_0$ counts changed features
(a one-hot switch counts as one); redundancy is the number of changed features that can be reverted
without losing validity~\citep{pawelczyk2021carla}; yNN is the fraction of the 5 nearest training points (Euclidean) that the
model assigns to the counterfactual's class; yNN-dist is their mean distance; time is wall-clock per
factual. Bold marks
the best value per column among arms with validity $\geq 80\%$ (a threshold we adopt to exclude arms with low validity);
distances within $10^{-3}$ of the best share the bold. Wins/ties/losses are paired per factual over
instances valid for both arms with the same $10^{-3}$ tie band; the Wilcoxon signed-rank test excludes
ties.

\paragraph{Summary across objectives.}
Table~\ref{tab:carla_headline} is the statistic the abstract and the introduction quote (Section~\ref{sec:results} quotes its per-norm
counterparts, Tables~\ref{tab:carla_bestknown}, \ref{tab:carla_l2_bestknown} and \ref{tab:carla_linf_bestknown}); Table~\ref{tab:carla_ratio_all} pools
the same 9{,}000 tabular queries into one distribution per arm: the ratio of each returned distance to the best that DASH or any of the
nine prior methods found on that query, with failures counted as infinite. Where FAB or FMN* is valid it is often as close as DASH or
closer: on the instances valid for both, FAB is closer than DASH more often than the reverse under every norm, and FMN* under
$\ell_1$ and $\ell_2$ (Tables~\ref{tab:carla_wtl}, \ref{tab:carla_l2_wtl} and \ref{tab:carla_linf_wtl}); their shortfall in
Table~\ref{tab:carla_headline} is validity (33.1\% and 45.9\%, against 99.4\% for DASH).
\begin{table}[htbp]
\centering\scriptsize\setlength{\tabcolsep}{3pt}
\caption{Headline statistic, all three objectives pooled, tabular datasets (Adult, COMPAS, GMC, HELOC, depths 2--32, 9{,}000 queries), no query excluded: share of queries on which the arm returns a valid counterfactual within 5\% of the smallest valid distance found by DASH or any of the nine prior methods, and the arm's median search time over all its queries. These are the numbers the abstract and the introduction quote; Section~\ref{sec:results} quotes the per-norm rates of Tables~\ref{tab:carla_bestknown}, \ref{tab:carla_l2_bestknown} and \ref{tab:carla_linf_bestknown}. 95\% CI = percentile cluster bootstrap over trained models (dataset $\times$ depth $\times$ seed), 2{,}000 resamples.}
\label{tab:carla_headline}
\setbox0=\hbox{%
\begin{tabular}{lrrrrr}
\toprule
method & queries & valid \% & within 5 \% & 95 \% CI & median time (s) \\
\midrule
DASH & 9000 & 99.4 & \textbf{94.6} & 93.0--96.1 & 0.061 \\
PGD-bisect* & 9000 & 93.6 & 41.1 & 33.5--48.6 & 0.298 \\
Wachter & 9000 & 93.0 & 37.5 & 31.8--43.5 & 1.583 \\
FAB & 9000 & 33.1 & 32.1 & 23.2--41.6 & 0.014 \\
FMN* & 9000 & 45.9 & 30.7 & 24.9--36.9 & 0.156 \\
NICE & 9000 & 98.9 & 14.6 & 10.4--19.0 & 0.023 \\
DiCE & 9000 & 94.5 & 8.0 & 6.0--10.0 & 0.001 \\
GS* & 9000 & 74.1 & 8.0 & 5.6--10.8 & 0.014 \\
DeepFool & 9000 & 10.8 & 5.2 & 3.4--7.2 & 0.000 \\
Greedy* & 9000 & 93.6 & 4.8 & 3.8--5.7 & 0.001 \\
\bottomrule
\end{tabular}}
\ifdim\wd0>\linewidth\resizebox{\linewidth}{!}{\box0}\else\box0\fi
\end{table}

\begin{table}[htbp]
\centering\scriptsize\setlength{\tabcolsep}{3pt}
\caption{Distance relative to the best-known counterfactual over every query of the benchmark (9000 queries: adult, compas, gmc, heloc, all three objectives, all depths): per query, the arm's distance divided by the smallest valid distance among DASH and the nine prior methods; an invalid answer, or an arm that did not run on that query, counts as infinite (rows = the arm's actual rows). Quantiles are over all queries; within $x$ = share of queries with ratio $\leq 1+x$ (2x, 10x = ratio $\leq 2$, $\leq 10$).}
\label{tab:carla_ratio_all}
\setbox0=\hbox{%
\begin{tabular}{lrrrrrrrrrrr}
\toprule
method & rows & queries & valid \% & median ratio & 75th pct & 90th pct & within 1 \% & within 5 \% & within 25 \% & within 2x & within 10x \\
\midrule
DASH & 9000 & 9000 & 99.4 & 1.00 & 1.00 & 1.02 & 84.5 & 94.6 & 98.8 & 99.1 & 99.4 \\
PGD-bisect* & 9000 & 9000 & 93.6 & 1.55 & 5.91 & 43.45 & 36.5 & 41.1 & 46.2 & 56.1 & 80.7 \\
Wachter & 9000 & 9000 & 93.0 & 1.19 & 4.84 & 18.58 & 28.6 & 37.5 & 53.3 & 62.8 & 86.1 \\
FAB & 9000 & 9000 & 33.1 & fail & fail & fail & 29.2 & 32.1 & 32.9 & 33.1 & 33.1 \\
FMN* & 9000 & 9000 & 45.9 & fail & fail & fail & 28.6 & 30.7 & 31.3 & 32.1 & 40.9 \\
NICE & 9000 & 9000 & 98.9 & 1.49 & 3.06 & 6.33 & 4.6 & 14.6 & 36.1 & 63.6 & 95.1 \\
GS* & 9000 & 9000 & 74.1 & 1.63 & fail & fail & 1.3 & 8.0 & 29.9 & 57.6 & 72.4 \\
DiCE & 9000 & 9000 & 94.5 & 1.44 & 2.15 & 8.86 & 1.6 & 8.0 & 33.1 & 72.8 & 90.4 \\
DeepFool & 9000 & 9000 & 10.8 & fail & fail & fail & 2.8 & 5.2 & 9.3 & 10.7 & 10.8 \\
Greedy* & 9000 & 9000 & 93.6 & 1.94 & 5.26 & 22.20 & 0.9 & 4.8 & 23.2 & 51.6 & 84.8 \\
\bottomrule
\end{tabular}}
\ifdim\wd0>\linewidth\resizebox{\linewidth}{!}{\box0}\else\box0\fi
\end{table}

\subsection{Factual-origin anchor ablation}
\label{app:factual-origin-ablation}
Table~\ref{tab:tangent_ablation} removes the factual-origin anchors from DASH and keeps every other component (one-directional anchors, including categorical transitions; local gradient-based anchors; ranking; beam selection; and polishing) unchanged. Validity is unaffected (100\% for both arms),
but the within-5\% rate drops from 95.6\% to 90.1\%: removing factual-origin anchors gives a worse distance on 180 queries and a better
one on 55 (Wilcoxon signed-rank $p<10^{-19}$). The effect is concentrated where the search must move several continuous
features jointly: under $\ell_2$ and $\ell_\infty$ and on HELOC and Adult. Under $\ell_1$ a single-feature move is typically
optimal, and the one-directional anchors already find it. Factual-origin anchors cost about 2\% more median runtime. The ablation ran on an Apple
M-series laptop (one thread per query), so runtimes are compared only within the table; on the same 900 queries its DASH arm
matches the cluster run's validity on every query and its distance on 823, and floating-point differences between the machines
change the search path on the other 77, moving the distance by at most 6\% in either direction.
\begin{table}[htbp]
\centering\scriptsize\setlength{\tabcolsep}{4pt}
\caption{Factual-origin anchor ablation: DASH against the same search without factual-origin anchors (all other components unchanged),
tabular datasets, depths 2-32, seed 17, the first 15 factuals of each model (900 queries per arm), paper settings; both arms return a valid counterfactual on every query.
Within 5\% = valid and within 5\% of the best-known distance of Table~\ref{tab:carla_headline} (DASH and the nine prior methods,
full paper run). B/T/L = queries where removing the factual-origin anchors gives a better/tied/worse distance (relative tie band $10^{-3}$);
90th = 90th percentile of the distance ratio (no factual-origin anchors / DASH). Both arms ran on the same machine, each in its own process,
so their runtimes are comparable with each other but not with the other tables.}
\label{tab:tangent_ablation}
\begin{tabular}{lrrrrrrr}
\toprule
 & queries & DASH & no factual-origin & B/T/L & 90th & median s: DASH & median s: no factual-origin anchors \\
\midrule
$\ell_1$ & 300 & 97.3 & 97.7 & 6/290/4 & 1.000 & 0.095 & 0.092 \\
$\ell_2$ & 300 & 96.3 & 88.7 & 25/181/94 & 1.034 & 0.110 & 0.105 \\
$\ell_\infty$ & 300 & 93.0 & 84.0 & 24/194/82 & 1.051 & 0.117 & 0.113 \\
\midrule
Adult & 225 & 100.0 & 90.2 & 15/157/53 & 1.035 & 0.159 & 0.152 \\
COMPAS & 225 & 99.6 & 100.0 & 5/209/11 & 1.000 & 0.052 & 0.048 \\
GMC & 225 & 90.2 & 92.0 & 17/194/14 & 1.000 & 0.057 & 0.054 \\
HELOC & 225 & 92.4 & 78.2 & 18/105/102 & 1.079 & 0.149 & 0.147 \\
\midrule
all & 900 & 95.6 & 90.1 & 55/665/180 & 1.022 & 0.109 & 0.107 \\
\bottomrule
\end{tabular}
\end{table}

\subsection{$\ell_1$ objective (headline)}
\begin{table}[htbp]
\centering\scriptsize\setlength{\tabcolsep}{3pt}
\caption{CARLA axes at depth 8 ($\ell_1$ objective, basic actionability; seeds 17/29/43).}
\label{tab:carla_main_d8}
\setbox0=\hbox{%
\begin{tabular}{lrrrrrrrrr}
\toprule
dataset & method & success \% & valid \% & $\ell_1$ & l0 & redund. & yNN & yNN-dist & time (s) \\
\midrule
adult & DASH & \textbf{100} & \textbf{100} & 0.694 & \textbf{1.0} & \textbf{0.0} & 0.000 & 1.004 & 0.0873 \\
 & DASH + data seed & \textbf{100} & \textbf{100} & 0.694 & \textbf{1.0} & \textbf{0.0} & 0.000 & 1.004 & 0.0940 \\
 & Wachter & 83 & 78 & 2.432 & 5.0 & 2.0 & 0.000 & 1.438 & 2.0056 \\
 & DiCE & 97 & 95 & 1.074 & 4.0 & \textbf{0.0} & 0.000 & \textbf{0.872} & 0.0021 \\
 & Greedy* & \textbf{100} & 82 & \textbf{0.600} & \textbf{1.0} & \textbf{0.0} & 0.000 & 0.933 & \textbf{0.0012} \\
 & GS* & 100 & 37 & 0.584 & 1.0 & 0.0 & 0.000 & 0.634 & 0.1199 \\
 & PGD-bisect* & 70 & 70 & 0.478 & 1.0 & 0.0 & 0.000 & 0.773 & 0.4647 \\
 & FMN* & 67 & 67 & 0.478 & 1.0 & 0.0 & 0.000 & 0.769 & 0.1431 \\
 & NICE & \textbf{100} & \textbf{100} & 2.565 & 3.0 & \textbf{0.0} & \textbf{0.200} & 0.972 & 0.0248 \\
 & FAB & 100 & 78 & 0.514 & 1.0 & 0.0 & 0.000 & 0.913 & 0.0151 \\
 & DeepFool & 100 & 60 & 0.460 & 1.0 & 0.0 & 0.000 & 0.715 & 0.0003 \\
 & MIP-EXP & 2 & 2 & 0.017 & 1.0 & 0.0 & 0.200 & 0.067 & 7.6279 \\
 & MIP-OBJ & 0 & 0 & NA & NA & NA & NA & NA & NA \\
compas & DASH & \textbf{98} & \textbf{98} & \textbf{0.144} & \textbf{1.0} & \textbf{0.0} & 0.400 & 0.027 & 0.0388 \\
 & DASH + data seed & \textbf{98} & \textbf{98} & \textbf{0.144} & \textbf{1.0} & \textbf{0.0} & 0.400 & 0.027 & 0.0407 \\
 & Wachter & 91 & 91 & \textbf{0.144} & 2.0 & 1.0 & 0.600 & \textbf{0.027} & 1.7761 \\
 & DiCE & 94 & 94 & 0.149 & \textbf{1.0} & \textbf{0.0} & 0.800 & \textbf{0.026} & 0.0009 \\
 & Greedy* & 97 & 85 & 0.200 & \textbf{1.0} & \textbf{0.0} & \textbf{1.000} & 0.029 & \textbf{0.0004} \\
 & GS* & 100 & 59 & 0.101 & 1.0 & 0.0 & 0.600 & 0.025 & 0.0101 \\
 & PGD-bisect* & 95 & 95 & 0.432 & 2.0 & 1.0 & 0.800 & 0.032 & 0.2912 \\
 & FMN* & 7 & 7 & 0.389 & 1.0 & 0.0 & 0.600 & 0.070 & 0.1478 \\
 & NICE & \textbf{98} & \textbf{98} & 0.154 & \textbf{1.0} & \textbf{0.0} & 0.600 & \textbf{0.026} & 0.0075 \\
 & FAB & 99 & 7 & 0.006 & 1.0 & 0.0 & 0.000 & 0.005 & 0.0138 \\
 & DeepFool & 100 & 10 & 0.322 & 1.0 & 0.0 & 0.800 & 0.040 & 0.0002 \\
 & MIP-EXP & 45 & 45 & 0.049 & 1.0 & 0.0 & 0.600 & 0.015 & 5.8384 \\
 & MIP-OBJ & 0 & 0 & NA & NA & NA & NA & NA & NA \\
gmc & DASH & \textbf{99} & \textbf{99} & 0.039 & \textbf{1.0} & \textbf{0.0} & 0.800 & 0.049 & 0.0478 \\
 & DASH + data seed & \textbf{99} & \textbf{99} & 0.039 & \textbf{1.0} & \textbf{0.0} & 0.800 & 0.049 & 0.0543 \\
 & Wachter & 97 & 97 & 0.038 & 3.0 & 1.0 & 0.800 & 0.049 & 1.3670 \\
 & DiCE & 91 & 91 & 0.050 & \textbf{1.0} & \textbf{0.0} & 0.800 & 0.056 & 0.0004 \\
 & Greedy* & 91 & 91 & 0.100 & \textbf{1.0} & \textbf{0.0} & \textbf{1.000} & 0.095 & \textbf{0.0003} \\
 & GS* & \textbf{99} & \textbf{99} & 0.040 & \textbf{1.0} & \textbf{0.0} & 0.800 & 0.050 & 0.0097 \\
 & PGD-bisect* & 97 & 97 & 0.087 & 4.0 & 2.0 & 0.800 & 0.055 & 0.1687 \\
 & FMN* & 1 & 1 & 0.072 & 1.0 & 0.0 & 1.000 & 0.077 & 0.1489 \\
 & NICE & 98 & 98 & 0.048 & \textbf{1.0} & \textbf{0.0} & 0.800 & 0.050 & 0.7981 \\
 & FAB & 100 & 1 & 0.005 & 1.0 & 0.0 & 0.900 & 0.052 & 0.0139 \\
 & DeepFool & 100 & 0 & NA & NA & NA & NA & NA & NA \\
 & MIP-EXP & 87 & 87 & \textbf{0.031} & \textbf{1.0} & \textbf{0.0} & 0.800 & \textbf{0.044} & 4.4221 \\
 & MIP-OBJ & 2 & 2 & 0.389 & 1.0 & 0.0 & 0.000 & 0.391 & 1.6557 \\
heloc & DASH & \textbf{100} & \textbf{100} & 0.436 & \textbf{2.0} & \textbf{0.0} & 0.200 & 0.416 & 0.2130 \\
 & DASH + data seed & \textbf{100} & \textbf{100} & 0.430 & \textbf{2.0} & \textbf{0.0} & 0.200 & 0.411 & 0.2203 \\
 & Wachter & \textbf{100} & \textbf{100} & 0.485 & 20.0 & 9.5 & 0.200 & 0.318 & 1.8254 \\
 & DiCE & 99 & 99 & 0.699 & 9.0 & \textbf{0.0} & 0.200 & 0.317 & \textbf{0.0005} \\
 & Greedy* & 99 & 99 & 0.500 & 3.0 & \textbf{0.0} & 0.200 & 0.384 & 0.0010 \\
 & GS* & \textbf{100} & \textbf{100} & 0.582 & 6.0 & \textbf{0.0} & 0.200 & 0.360 & 0.0159 \\
 & PGD-bisect* & \textbf{100} & \textbf{100} & 0.538 & 6.0 & 1.5 & 0.200 & 0.401 & 0.3132 \\
 & FMN* & 67 & 67 & 0.372 & 3.0 & 0.0 & 0.200 & 0.367 & 0.1556 \\
 & NICE & \textbf{100} & \textbf{100} & 0.634 & 6.0 & 1.0 & \textbf{0.400} & \textbf{0.287} & 0.0115 \\
 & FAB & 98 & 98 & \textbf{0.418} & 3.0 & \textbf{0.0} & 0.200 & 0.371 & 0.0137 \\
 & DeepFool & 99 & 35 & 0.305 & 1.0 & 0.0 & 0.400 & 0.359 & 0.0003 \\
 & MIP-EXP & 7 & 7 & 0.032 & 1.0 & 0.0 & 0.400 & 0.222 & 11.0365 \\
 & MIP-OBJ & 0 & 0 & NA & NA & NA & NA & NA & NA \\
\bottomrule
\end{tabular}}
\ifdim\wd0>\linewidth\resizebox{\linewidth}{!}{\box0}\else\box0\fi
\par\smallskip\footnotesize success = the returned point crosses the threshold (CARLA's success rate); valid = it also satisfies the actionability box and one-hot constraints. Bold = best per column within a dataset among arms with validity $\geq$ 80\%. Medians are over each arm's own valid rows; an arm with low validity is summarised on an easier subset, so compare arms with different validity through the paired table (Table~\ref{tab:carla_wtl}) and the best-known table (Table~\ref{tab:carla_bestknown}).
\end{table}

\begin{table}[htbp]
\centering\scriptsize\setlength{\tabcolsep}{3pt}
\caption{$\ell_1$ relative to the certified optimum (MIP-EXP or MIP-OBJ; Gurobi, 4 threads, 60\,s budget per factual, absolute optimality gap $10^{-4}$, depths 2/4/8, same 50 factuals per cell as the main run). certified n = instances on which either exact formulation proved an optimum in that dataset; arm valid = how many of those the arm returns a valid point on; at optimum = arm valid and distance $\leq$ ref $+10^{-3}$ (one-sided: below the solver's incumbent is inside its gap); within 5\% = arm valid and distance $\leq 1.05\times$ref (purely relative: where the optimum is below 0.02 the absolute tie band is the wider of the two, so the columns are not nested); both percentages over certified n; ratio columns over the arm-valid subset. A valid point can lie up to $6\times10^{-4}$ below the certified optimum (the solver's absolute gap and numerical tolerances), so ratios below 1 occur where the optimum itself is small: every ratio below 0.95 has an optimum under $2.5\times10^{-3}$.}
\label{tab:carla_optimality}
\setbox0=\hbox{%
\begin{tabular}{lrrrrrrr}
\toprule
dataset & arm & certified n & arm valid & at optimum \% & within 5 \% & median ratio & max ratio \\
\midrule
adult & DASH & 187 & 187 & 100.0 & 100.0 & 1.000 & 1.001 \\
adult & DASH + data seed & 187 & 187 & 100.0 & 100.0 & 1.000 & 1.001 \\
adult & Wachter & 187 & 182 & 41.7 & 42.8 & 2.567 & 12.574 \\
adult & DiCE & 187 & 187 & 0.5 & 1.1 & 2.028 & 62.500 \\
adult & Greedy* & 187 & 187 & 1.1 & 10.2 & 1.260 & 125.000 \\
adult & GS* & 187 & 141 & 0.5 & 6.4 & 1.261 & 2.142 \\
adult & PGD-bisect* & 187 & 187 & 100.0 & 100.0 & 1.000 & 1.007 \\
adult & FMN* & 187 & 173 & 92.5 & 92.5 & 1.000 & 1.000 \\
adult & NICE & 187 & 187 & 0.0 & 1.6 & 5.690 & 548.469 \\
adult & FAB & 187 & 187 & 100.0 & 100.0 & 1.000 & 1.000 \\
adult & DeepFool & 187 & 186 & 55.6 & 82.4 & 1.011 & 1.531 \\
compas & DASH & 313 & 313 & 97.4 & 99.7 & 1.000 & 1.065 \\
compas & DASH + data seed & 313 & 313 & 97.1 & 99.7 & 1.000 & 1.065 \\
compas & Wachter & 313 & 302 & 84.3 & 85.9 & 1.000 & 86.976 \\
compas & DiCE & 313 & 311 & 1.9 & 15.0 & 1.269 & 166.667 \\
compas & Greedy* & 313 & 305 & 0.6 & 5.1 & 1.514 & 333.333 \\
compas & GS* & 313 & 237 & 3.2 & 9.6 & 1.211 & 23.564 \\
compas & PGD-bisect* & 313 & 308 & 2.2 & 2.6 & 1.510 & 2000.933 \\
compas & FMN* & 313 & 1 & 0.3 & 0.3 & 0.979 & 0.979 \\
compas & NICE & 313 & 311 & 6.7 & 35.8 & 1.079 & 9.149 \\
compas & FAB & 313 & 36 & 11.5 & 11.5 & 1.000 & 1.000 \\
compas & DeepFool & 313 & 2 & 0.3 & 0.3 & 1.049 & 1.080 \\
gmc & DASH & 413 & 413 & 99.3 & 99.0 & 1.000 & 1.148 \\
gmc & DASH + data seed & 413 & 413 & 99.3 & 99.0 & 1.000 & 1.148 \\
gmc & Wachter & 413 & 413 & 97.6 & 97.1 & 1.000 & 1.973 \\
gmc & DiCE & 413 & 406 & 0.5 & 2.7 & 2.240 & 521.374 \\
gmc & Greedy* & 413 & 407 & 0.2 & 1.9 & 4.367 & 1042.749 \\
gmc & GS* & 413 & 413 & 32.7 & 37.5 & 1.078 & 27.974 \\
gmc & PGD-bisect* & 413 & 406 & 12.1 & 9.2 & 1.913 & 15.053 \\
gmc & FMN* & 413 & 45 & 10.9 & 10.9 & 1.000 & 1.002 \\
gmc & NICE & 413 & 406 & 29.8 & 27.4 & 1.129 & 24.263 \\
gmc & FAB & 413 & 45 & 10.9 & 10.9 & 1.000 & 1.000 \\
gmc & DeepFool & 413 & 46 & 4.4 & 2.9 & 1.092 & 2.278 \\
heloc & DASH & 158 & 158 & 70.9 & 96.2 & 1.001 & 1.086 \\
heloc & DASH + data seed & 158 & 158 & 71.5 & 96.2 & 1.001 & 1.086 \\
heloc & Wachter & 158 & 158 & 24.7 & 46.2 & 1.062 & 1.413 \\
heloc & DiCE & 158 & 158 & 0.0 & 0.6 & 1.518 & 33.333 \\
heloc & Greedy* & 158 & 158 & 0.0 & 15.2 & 1.182 & 66.667 \\
heloc & GS* & 158 & 158 & 0.0 & 1.3 & 1.449 & 9.312 \\
heloc & PGD-bisect* & 158 & 158 & 55.7 & 60.8 & 1.001 & 6.897 \\
heloc & FMN* & 158 & 100 & 62.0 & 63.3 & 1.000 & 1.034 \\
heloc & NICE & 158 & 158 & 0.0 & 0.0 & 1.604 & 14.781 \\
heloc & FAB & 158 & 158 & 79.7 & 98.7 & 1.000 & 1.106 \\
heloc & DeepFool & 158 & 100 & 19.0 & 42.4 & 1.028 & 1.341 \\
all & DASH & 1071 & 1071 & 94.7 & 99.0 & 1.000 & 1.148 \\
all & DASH + data seed & 1071 & 1071 & 94.7 & 99.0 & 1.000 & 1.148 \\
all & Wachter & 1071 & 1055 & 73.2 & 76.8 & 1.000 & 86.976 \\
all & DiCE & 1071 & 1062 & 0.8 & 5.7 & 1.699 & 521.374 \\
all & Greedy* & 1071 & 1057 & 0.5 & 6.3 & 1.705 & 1042.749 \\
all & GS* & 1071 & 949 & 13.6 & 18.6 & 1.178 & 27.974 \\
all & PGD-bisect* & 1071 & 1059 & 31.0 & 30.7 & 1.512 & 2000.933 \\
all & FMN* & 1071 & 319 & 29.6 & 29.8 & 1.000 & 1.034 \\
all & NICE & 1071 & 1062 & 13.4 & 21.3 & 1.249 & 548.469 \\
all & FAB & 1071 & 426 & 36.8 & 39.6 & 1.000 & 1.106 \\
all & DeepFool & 1071 & 334 & 14.3 & 21.8 & 1.022 & 2.278 \\
\bottomrule
\end{tabular}}
\ifdim\wd0>\linewidth\resizebox{\linewidth}{!}{\box0}\else\box0\fi
\end{table}

\begin{table}[htbp]
\centering\scriptsize\setlength{\tabcolsep}{3pt}
\caption{MIP-EXP against DASH by depth, $\ell_1$ objective, same queries, pooled over datasets and seeds. Median times are over all of these queries, solved or not (MIP-EXP stops at its 60\,s budget).}
\label{tab:carla_mip_depth}
\setbox0=\hbox{%
\begin{tabular}{lrrrrrr}
\toprule
depth & queries & MIP-EXP valid \% & MIP-EXP certified \% & MIP-EXP median time (s) & DASH valid \% & DASH median time (s) \\
\midrule
2 & 600 & 99.7 & 94.5 & 1.977 & 99.7 & 0.031 \\
4 & 600 & 81.0 & 46.8 & 60.000 & 99.5 & 0.051 \\
8 & 600 & 35.5 & 31.5 & 60.000 & 99.2 & 0.073 \\
\bottomrule
\end{tabular}}
\ifdim\wd0>\linewidth\resizebox{\linewidth}{!}{\box0}\else\box0\fi
\end{table}

\begin{table}[htbp]
\centering\scriptsize\setlength{\tabcolsep}{3pt}
\caption{Best-known counterfactual, $\ell_1$ objective: population is every tabular query of this norm (the queries column); best = smallest valid distance among DASH and the nine prior methods on that query (the exact solvers and the ablation are not in the reference); a query counts for an arm at $\tau$ iff the arm is valid and its distance $\leq(1+\tau)\times$best; column best uses the absolute tie band best $+10^{-3}$ instead, which is wider than 1\% where best is below 0.1; invalid rows never count, no penalty constant; 95\% CI = percentile cluster bootstrap over trained models (dataset $\times$ depth $\times$ seed), 2{,}000 resamples. Sorted by within 5\%; bold = best per column.}
\label{tab:carla_bestknown}
\setbox0=\hbox{%
\begin{tabular}{lrrrrrrr}
\toprule
method & queries & valid \% & best & within 1 \% & within 5 \% & 95 \% CI (within 5 \%) & within 10 \% \\
\midrule
DASH & 3000 & \textbf{99.4} & \textbf{85.6} & \textbf{88.9} & \textbf{95.7} & 93.7--97.5 & \textbf{98.0} \\
Wachter & 3000 & 93.6 & 46.3 & 44.6 & 53.6 & 45.8--61.0 & 59.5 \\
FAB & 3000 & 50.6 & 40.9 & 46.1 & 48.6 & 37.5--59.9 & 49.2 \\
FMN* & 3000 & 40.1 & 37.8 & 38.2 & 38.8 & 29.2--48.0 & 39.1 \\
PGD-bisect* & 3000 & 92.3 & 30.9 & 30.9 & 32.8 & 24.1--42.4 & 34.7 \\
NICE & 3000 & 98.9 & 7.9 & 5.0 & 16.1 & 12.1--20.2 & 25.2 \\
DeepFool & 3000 & 31.6 & 8.9 & 8.2 & 15.3 & 9.9--21.1 & 19.9 \\
GS* & 3000 & 79.8 & 7.9 & 2.4 & 12.8 & 9.0--16.9 & 22.7 \\
Greedy* & 3000 & 93.6 & 0.9 & 2.2 & 11.1 & 8.6--13.7 & 22.1 \\
DiCE & 3000 & 94.9 & 0.9 & 1.2 & 6.7 & 4.6--9.0 & 11.8 \\
\bottomrule
\end{tabular}}
\ifdim\wd0>\linewidth\resizebox{\linewidth}{!}{\box0}\else\box0\fi
\end{table}

\begin{figure}[htbp]
\centering
\includegraphics[width=\linewidth]{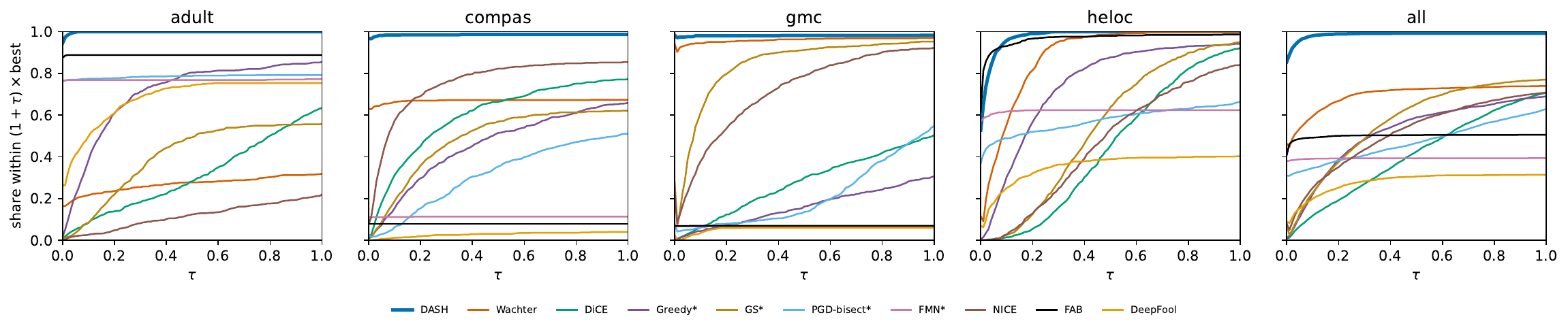}
\caption{Performance profile, $\ell_1$ objective: share of the tabular queries (750 per dataset panel, 3,000 in the pooled panel) where the arm is valid and within $(1+\tau)\times$ the best heuristic-observed valid distance of Table~\ref{tab:carla_bestknown} ($\tau=\alpha-1$ in the notation of Section~\ref{sec:results}; at $\tau=0$ the $10^{-3}$ tie band applies), as $\tau$ ranges from 0 (the heuristic reference) to 1 (twice that reference). The right end of a curve cannot exceed the arm's validity. The reference excludes the exact solvers; Table~\ref{tab:carla_optimality} compares with the certified optimum where it exists.}
\label{fig:carla_profile}
\end{figure}
\begin{table}[htbp]
\centering\scriptsize\setlength{\tabcolsep}{3pt}
\caption{DASH against each arm on $\ell_1$, paired per factual over the instances where both are valid: DASH wins = DASH is closer by more than $10^{-3}$, losses = the other arm is closer; cells won = (depth, seed) cells ("all" row: dataset, depth, seed cells) with $\geq 1$ pair, where the median paired difference favours DASH by more than $10^{-3}$, printed as $k/m$; median ratio = median of DASH distance / other distance (below 1 favours DASH); two-sided Wilcoxon signed-rank on the non-tied pairs. Pairs within a cell share one trained model, so the pooled p-value overstates the evidence; cells won counts model-level agreement.}
\label{tab:carla_wtl}
\setbox0=\hbox{%
\begin{tabular}{lrrrrrrrr}
\toprule
dataset & vs & n paired & DASH wins & ties & DASH losses & cells won & median ratio & Wilcoxon p \\
\midrule
adult & DASH + data seed & 750 & 9 & 732 & 9 & 0/15 & 1.000 & 0.52 \\
adult & Wachter & 654 & 531 & 123 & 0 & 15/15 & 0.296 & 1.1e-88 \\
adult & DiCE & 717 & 708 & 2 & 7 & 15/15 & 0.563 & 6.3e-118 \\
adult & Greedy* & 685 & 665 & 9 & 11 & 15/15 & 0.892 & 2.8e-109 \\
adult & GS* & 426 & 420 & 6 & 0 & 15/15 & 0.796 & 1.5e-70 \\
adult & PGD-bisect* & 604 & 31 & 572 & 1 & 0/15 & 1.000 & 1.4e-09 \\
adult & FMN* & 588 & 13 & 563 & 12 & 0/15 & 1.000 & 0.026 \\
adult & NICE & 750 & 748 & 2 & 0 & 15/15 & 0.249 & 4.2e-124 \\
adult & FAB & 666 & 7 & 629 & 30 & 0/15 & 1.000 & 1.4e-05 \\
adult & DeepFool & 565 & 369 & 196 & 0 & 11/15 & 0.966 & 3.2e-62 \\
adult & MIP-EXP & 303 & 109 & 194 & 0 & 3/8 & 1.000 & 1.3e-19 \\
adult & MIP-OBJ & 299 & 127 & 172 & 0 & 3/6 & 1.000 & 1.4e-22 \\
compas & DASH + data seed & 740 & 1 & 738 & 1 & 0/15 & 1.000 & NA \\
compas & Wachter & 669 & 194 & 467 & 8 & 0/15 & 1.000 & 4.3e-34 \\
compas & DiCE & 668 & 653 & 15 & 0 & 15/15 & 0.840 & 1.3e-108 \\
compas & Greedy* & 653 & 646 & 7 & 0 & 15/15 & 0.727 & 1.9e-107 \\
compas & GS* & 477 & 456 & 20 & 1 & 15/15 & 0.850 & 2.5e-75 \\
compas & PGD-bisect* & 685 & 664 & 20 & 1 & 15/15 & 0.570 & 1.7e-110 \\
compas & FMN* & 99 & 16 & 79 & 4 & 0/9 & 1.000 & 8.2e-05 \\
compas & NICE & 728 & 671 & 52 & 5 & 15/15 & 0.934 & 6.4e-109 \\
compas & FAB & 59 & 0 & 58 & 1 & 0/15 & 1.001 & NA \\
compas & DeepFool & 33 & 32 & 1 & 0 & 7/7 & 0.807 & 4.7e-10 \\
compas & MIP-EXP & 359 & 44 & 306 & 9 & 0/9 & 1.000 & 1.4e-07 \\
compas & MIP-OBJ & 292 & 62 & 223 & 7 & 1/6 & 1.000 & 1.8e-12 \\
gmc & DASH + data seed & 742 & 0 & 742 & 0 & 0/15 & 1.000 & NA \\
gmc & Wachter & 736 & 30 & 703 & 3 & 0/15 & 1.000 & 1.1e-05 \\
gmc & DiCE & 717 & 712 & 5 & 0 & 15/15 & 0.515 & 3.2e-118 \\
gmc & Greedy* & 720 & 714 & 3 & 3 & 15/15 & 0.294 & 6.5e-116 \\
gmc & GS* & 742 & 530 & 209 & 3 & 15/15 & 0.933 & 2.8e-85 \\
gmc & PGD-bisect* & 730 & 672 & 58 & 0 & 14/15 & 0.518 & 1.1e-111 \\
gmc & FMN* & 49 & 0 & 49 & 0 & 0/6 & 1.000 & NA \\
gmc & NICE & 735 & 558 & 177 & 0 & 15/15 & 0.863 & 4.3e-93 \\
gmc & FAB & 52 & 1 & 51 & 0 & 0/9 & 1.000 & NA \\
gmc & DeepFool & 46 & 28 & 18 & 0 & 2/3 & 0.916 & 7.5e-09 \\
gmc & MIP-EXP & 430 & 14 & 413 & 3 & 0/9 & 0.999 & 0.19 \\
gmc & MIP-OBJ & 297 & 1 & 293 & 3 & 0/8 & 1.000 & NA \\
heloc & DASH + data seed & 750 & 65 & 577 & 108 & 0/15 & 1.000 & 0.0016 \\
heloc & Wachter & 750 & 576 & 84 & 90 & 15/15 & 0.937 & 6.1e-79 \\
heloc & DiCE & 746 & 744 & 2 & 0 & 15/15 & 0.665 & 1.9e-123 \\
heloc & Greedy* & 749 & 706 & 4 & 39 & 15/15 & 0.872 & 3.4e-114 \\
heloc & GS* & 750 & 746 & 0 & 4 & 15/15 & 0.715 & 3.4e-124 \\
heloc & PGD-bisect* & 750 & 425 & 197 & 128 & 8/15 & 0.887 & 7.2e-67 \\
heloc & FMN* & 468 & 29 & 224 & 215 & 0/15 & 1.002 & 1.5e-21 \\
heloc & NICE & 750 & 748 & 0 & 2 & 15/15 & 0.689 & 2e-124 \\
heloc & FAB & 741 & 141 & 316 & 284 & 0/15 & 1.001 & 8.3e-08 \\
heloc & DeepFool & 303 & 224 & 56 & 23 & 14/15 & 0.956 & 3.6e-33 \\
heloc & MIP-EXP & 205 & 43 & 116 & 46 & 3/9 & 1.000 & 0.037 \\
heloc & MIP-OBJ & 299 & 108 & 118 & 73 & 3/6 & 1.000 & 1.6e-13 \\
all & DASH + data seed & 2982 & 75 & 2789 & 118 & 0/60 & 1.000 & 0.0043 \\
all & Wachter & 2809 & 1331 & 1377 & 101 & 30/60 & 0.992 & 6.9e-212 \\
all & DiCE & 2848 & 2817 & 24 & 7 & 60/60 & 0.641 & $<10^{-300}$ \\
all & Greedy* & 2807 & 2731 & 23 & 53 & 60/60 & 0.788 & $<10^{-300}$ \\
all & GS* & 2395 & 2152 & 235 & 8 & 60/60 & 0.827 & $<10^{-300}$ \\
all & PGD-bisect* & 2769 & 1792 & 847 & 130 & 37/60 & 0.674 & 6e-282 \\
all & FMN* & 1204 & 58 & 915 & 231 & 0/45 & 1.000 & 5.9e-09 \\
all & NICE & 2963 & 2725 & 231 & 7 & 60/60 & 0.732 & $<10^{-300}$ \\
all & FAB & 1518 & 149 & 1054 & 315 & 0/54 & 1.000 & 3.7e-10 \\
all & DeepFool & 947 & 653 & 271 & 23 & 34/40 & 0.953 & 1.1e-103 \\
all & MIP-EXP & 1297 & 210 & 1029 & 58 & 6/35 & 1.000 & 6.9e-27 \\
all & MIP-OBJ & 1187 & 298 & 806 & 83 & 7/26 & 1.000 & 3.2e-51 \\
\bottomrule
\end{tabular}}
\ifdim\wd0>\linewidth\resizebox{\linewidth}{!}{\box0}\else\box0\fi
\end{table}

\begin{table}[htbp]
\centering\scriptsize\setlength{\tabcolsep}{3pt}
\caption{Validity (\%) per arm and cell, $\ell_1$ objective.}
\label{tab:carla_validity}
\setbox0=\hbox{%
\begin{tabular}{lrrrrrrrrrrrrrr}
\toprule
dataset & depth & DASH & DASH + data seed & Wachter & DiCE & Greedy* & GS* & PGD-bisect* & FMN* & NICE & FAB & DeepFool & MIP-EXP & MIP-OBJ \\
\midrule
adult & 2 & 100 & 100 & 97 & 100 & 100 & 73 & 100 & 91 & 100 & 100 & 99 & 100 & 100 \\
 & 4 & 100 & 100 & 90 & 100 & 100 & 79 & 100 & 92 & 100 & 100 & 95 & 100 & 99 \\
 & 8 & 100 & 100 & 78 & 95 & 82 & 37 & 70 & 67 & 100 & 78 & 60 & 2 & 0 \\
 & 16 & 100 & 100 & 87 & 95 & 91 & 66 & 84 & 89 & 100 & 91 & 84 & -- & -- \\
 & 32 & 100 & 100 & 83 & 88 & 83 & 29 & 49 & 53 & 100 & 75 & 39 & -- & -- \\
compas & 2 & 99 & 99 & 93 & 98 & 97 & 77 & 96 & 0 & 98 & 9 & 1 & 99 & 99 \\
 & 4 & 98 & 98 & 90 & 97 & 97 & 67 & 91 & 0 & 95 & 7 & 0 & 95 & 95 \\
 & 8 & 98 & 98 & 91 & 94 & 85 & 59 & 95 & 7 & 98 & 7 & 10 & 45 & 0 \\
 & 16 & 98 & 98 & 92 & 86 & 79 & 59 & 89 & 23 & 97 & 8 & 10 & -- & -- \\
 & 32 & 100 & 100 & 80 & 70 & 77 & 56 & 87 & 35 & 99 & 7 & 1 & -- & -- \\
gmc & 2 & 99 & 99 & 99 & 99 & 99 & 99 & 98 & 28 & 97 & 27 & 30 & 99 & 98 \\
 & 4 & 100 & 100 & 99 & 97 & 99 & 100 & 97 & 1 & 98 & 2 & 1 & 100 & 98 \\
 & 8 & 99 & 99 & 97 & 91 & 91 & 99 & 97 & 1 & 98 & 1 & 0 & 87 & 2 \\
 & 16 & 99 & 99 & 99 & 98 & 99 & 99 & 98 & 3 & 99 & 2 & 0 & -- & -- \\
 & 32 & 98 & 98 & 96 & 92 & 93 & 98 & 96 & 0 & 98 & 3 & 0 & -- & -- \\
heloc & 2 & 100 & 100 & 100 & 100 & 100 & 100 & 100 & 64 & 100 & 100 & 59 & 100 & 99 \\
 & 4 & 100 & 100 & 100 & 100 & 100 & 100 & 100 & 36 & 100 & 100 & 27 & 29 & 100 \\
 & 8 & 100 & 100 & 100 & 99 & 99 & 100 & 100 & 67 & 100 & 98 & 35 & 7 & 0 \\
 & 16 & 100 & 100 & 100 & 98 & 100 & 100 & 100 & 77 & 100 & 96 & 35 & -- & -- \\
 & 32 & 100 & 100 & 100 & 100 & 100 & 100 & 100 & 68 & 100 & 100 & 47 & -- & -- \\
\bottomrule
\end{tabular}}
\ifdim\wd0>\linewidth\resizebox{\linewidth}{!}{\box0}\else\box0\fi
\end{table}

\begin{table}[htbp]
\centering\scriptsize\setlength{\tabcolsep}{3pt}
\caption{MNIST (3 vs 8, $n=784$, depth 8, width 256, 150 factuals), $\ell_1$ objective. Greedy* did not run at this scale; under $\ell_1$ the norm-adapted GS* and PGD-bisect* runs cover 70 of the 150 factuals (validity over the arm's own rows). PBMC3k is reported in the main text, after the shared marker completion. DASH = the CPU implementation; NA = no valid counterfactual, or a metric the run did not record.}
\label{tab:carla_large}
\setbox0=\hbox{%
\begin{tabular}{lrrrrrrrrr}
\toprule
dataset & method & success \% & valid \% & $\ell_1$ & l0 & redund. & yNN & yNN-dist & time (s) \\
\midrule
mnist & DASH & \textbf{100} & \textbf{100} & 8.890 & \textbf{11.0} & \textbf{0.0} & 0.000 & 5.583 & 61.8500 \\
mnist & DASH + data seed & \textbf{100} & \textbf{100} & 8.913 & \textbf{11.0} & \textbf{0.0} & 0.000 & 5.583 & 63.4336 \\
mnist & Wachter & \textbf{100} & \textbf{100} & 22.274 & 607.0 & 274.0 & 0.000 & \textbf{5.136} & 3.4466 \\
mnist & DiCE & 85 & 85 & 17.454 & 177.0 & \textbf{0.0} & 0.000 & 5.159 & \textbf{0.0436} \\
mnist & GS* & 57 & 57 & 15.816 & 34.5 & 17.5 & 0.000 & 5.695 & 27.8042 \\
mnist & PGD-bisect* & 23 & 23 & 173.9 & 462.0 & 459.0 & 0.000 & 11.548 & 3.0613 \\
mnist & FMN* & 77 & 77 & 8.135 & 12.0 & 0.0 & 0.000 & 5.742 & 0.8633 \\
mnist & NICE & \textbf{100} & \textbf{100} & 11.790 & 22.0 & 6.0 & \textbf{0.200} & 5.647 & 0.7475 \\
mnist & FAB & \textbf{100} & \textbf{100} & \textbf{7.882} & 16.0 & \textbf{0.0} & 0.000 & 5.688 & 0.1730 \\
mnist & DeepFool & 100 & 1 & 0.034 & 1.0 & 0.0 & 0.000 & 5.614 & 0.0011 \\
\bottomrule
\end{tabular}}
\ifdim\wd0>\linewidth\resizebox{\linewidth}{!}{\box0}\else\box0\fi
\end{table}

\begin{figure}[htbp]
\centering
\includegraphics[width=\linewidth]{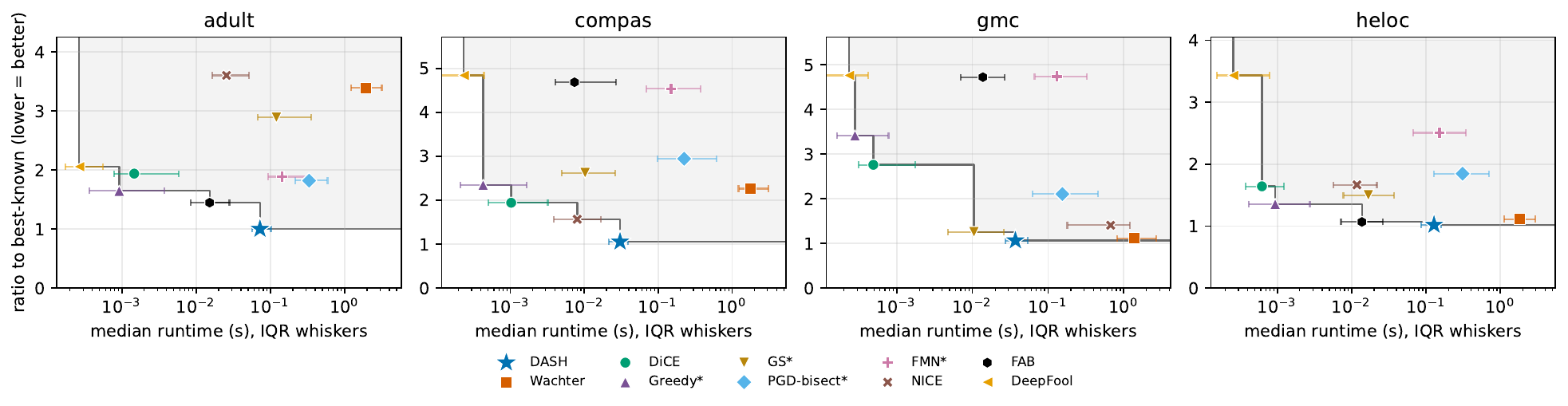}
\caption{Runtime-quality trade-off, $\ell_1$ objective, tabular datasets, all depths and seeds pooled.
Horizontal axis: median runtime per query (25th-75th percentile whiskers). Vertical axis: mean failure-penalized
ratio to the best heuristic-observed valid distance - a valid counterfactual scores its distance divided by the smallest valid distance DASH or any
of the nine prior methods found on that query (the reference of Table~\ref{tab:carla_bestknown}); a failure, or a
ratio above 5, scores 5. The staircase is the Pareto front, drawn to the axis edges; the shaded region is dominated.}
\label{fig:carla_pareto}
\end{figure}
\begin{figure}[htbp]
\centering
\includegraphics[width=\linewidth]{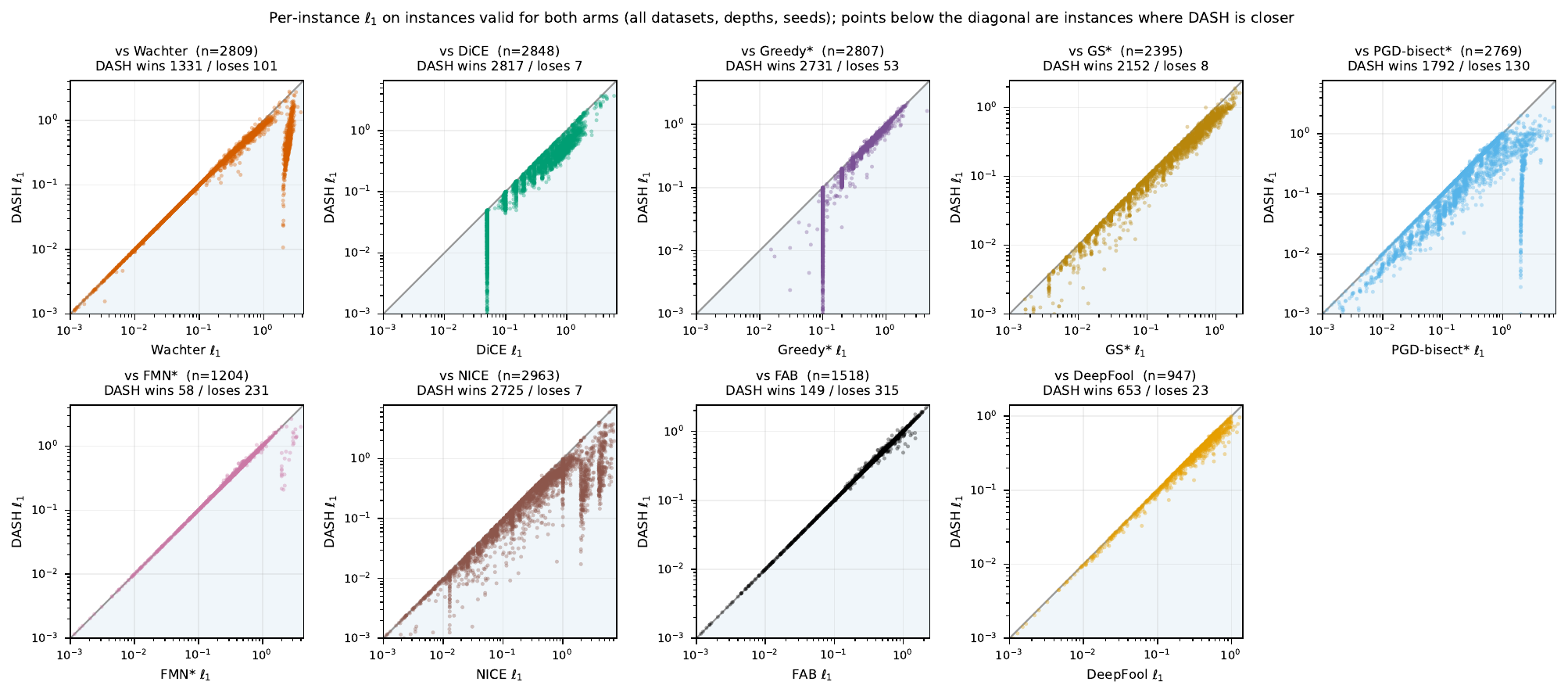}
\caption{The paired comparison behind Table~\ref{tab:carla_wtl}: for each baseline, one point per instance
valid for both arms, DASH's $\ell_1$ against the baseline's (log-log). Points below the diagonal are instances
where DASH is closer; the vertical streaks are instances where the baseline returns the same distance for many
factuals (a fixed step size or a one-hot flip) while DASH finds a closer point.}
\label{fig:carla_paired}
\end{figure}
\begin{figure}[htbp]
\centering
\includegraphics[width=\linewidth]{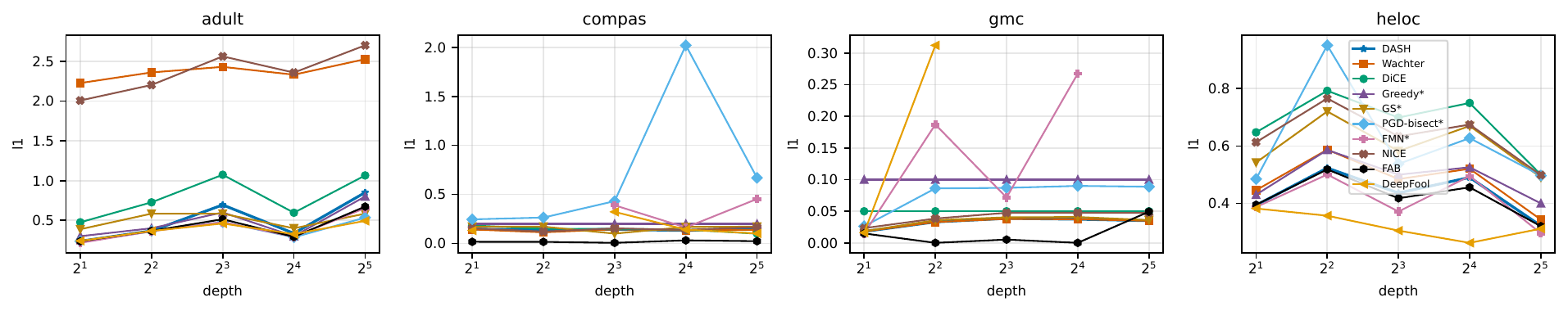}\\
\includegraphics[width=\linewidth]{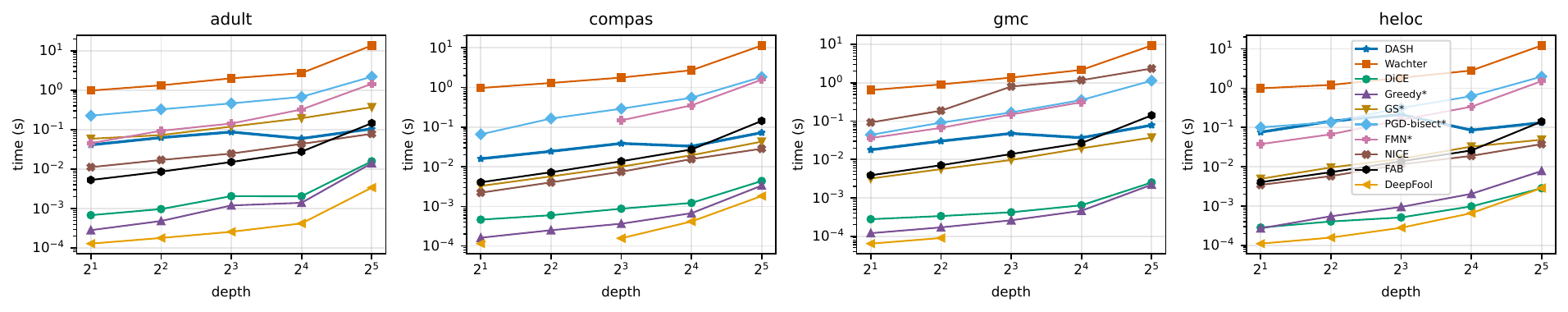}
\caption{Median $\ell_1$ (top) and median runtime (bottom, log scale) of the valid counterfactuals per
arm as network depth grows, $\ell_1$ objective. DASH + data seed is omitted (it overlaps DASH).}
\label{fig:carla_l1_depth}
\end{figure}
\clearpage
{\scriptsize\setlength{\tabcolsep}{3pt}
\begin{longtable}{lrrrrrrrrrrrr}
\caption{All tabular cells, $\ell_1$ objective (medians over valid rows; bold = best per column within a cell among arms with validity $\geq$ 80\%). Medians are over each arm's own valid rows; an arm with low validity is summarised on an easier subset, so compare arms with different validity through the paired table (Table~\ref{tab:carla_wtl}) and the best-known table (Table~\ref{tab:carla_bestknown}).} \label{tab:carla_per_cell} \\
\toprule
dataset & depth & method & success \% & valid \% & $\ell_1$ & l0 & redund. & yNN & yNN-dist & time (s) & $\ell_2$ & $\ell_\infty$ \\
\midrule
\endfirsthead
\caption[]{All tabular cells, $\ell_1$ objective (medians over valid rows; bold = best per column within a cell among arms with validity $\geq$ 80\%). Medians are over each arm's own valid rows; an arm with low validity is summarised on an easier subset, so compare arms with different validity through the paired table (Table~\ref{tab:carla_wtl}) and the best-known table (Table~\ref{tab:carla_bestknown}).} \\
\toprule
dataset & depth & method & success \% & valid \% & $\ell_1$ & l0 & redund. & yNN & yNN-dist & time (s) & $\ell_2$ & $\ell_\infty$ \\
\midrule
\endhead
\midrule
\multicolumn{13}{r}{Continued on next page} \\
\midrule
\endfoot
\bottomrule
\endlastfoot
adult & 2 & DASH & \textbf{100} & \textbf{100} & 0.237 & \textbf{1.0} & \textbf{0.0} & \textbf{0.000} & 0.344 & 0.0411 & 0.237 & 0.237 \\
adult & 2 & DASH + data seed & \textbf{100} & \textbf{100} & 0.237 & \textbf{1.0} & \textbf{0.0} & \textbf{0.000} & 0.344 & 0.0462 & 0.237 & 0.237 \\
adult & 2 & Wachter & \textbf{100} & 97 & 2.229 & 4.0 & 1.0 & \textbf{0.000} & 1.426 & 0.9827 & 1.424 & 1.000 \\
adult & 2 & DiCE & \textbf{100} & \textbf{100} & 0.475 & 3.0 & \textbf{0.0} & \textbf{0.000} & 0.401 & 0.0007 & 0.293 & \textbf{0.190} \\
adult & 2 & Greedy* & \textbf{100} & \textbf{100} & 0.300 & \textbf{1.0} & \textbf{0.0} & \textbf{0.000} & 0.400 & 0.0003 & 0.300 & 0.300 \\
adult & 2 & GS* & 100 & 73 & 0.388 & 1.0 & 0.0 & 0.000 & 0.417 & 0.0589 & 0.388 & 0.388 \\
adult & 2 & PGD-bisect* & \textbf{100} & \textbf{100} & 0.237 & \textbf{1.0} & \textbf{0.0} & \textbf{0.000} & 0.344 & 0.2265 & 0.237 & 0.237 \\
adult & 2 & FMN* & 91 & 91 & \textbf{0.216} & \textbf{1.0} & \textbf{0.0} & \textbf{0.000} & \textbf{0.341} & 0.0454 & \textbf{0.216} & 0.216 \\
\end{longtable}
}

\clearpage

\subsection{$\ell_2$ objective}
\begin{table}[htbp]
\centering\scriptsize\setlength{\tabcolsep}{3pt}
\caption{CARLA axes at depth 8 ($\ell_2$ objective, basic actionability; seeds 17/29/43).}
\label{tab:carla_l2_main_d8}
\setbox0=\hbox{%
\begin{tabular}{lrrrrrrrrr}
\toprule
dataset & method & success \% & valid \% & $\ell_2$ & l0 & redund. & yNN & yNN-dist & time (s) \\
\midrule
adult & DASH & \textbf{100} & \textbf{100} & \textbf{0.526} & 4.0 & \textbf{0.0} & 0.000 & 0.904 & 0.1010 \\
 & DASH + data seed & \textbf{100} & \textbf{100} & \textbf{0.526} & 4.0 & \textbf{0.0} & 0.000 & 0.903 & 0.1046 \\
 & Wachter & 79 & 77 & 1.428 & 5.0 & 3.0 & 0.000 & 1.430 & 1.8042 \\
 & DiCE & 99 & 92 & 0.550 & 4.0 & \textbf{0.0} & 0.000 & \textbf{0.853} & 0.0017 \\
 & Greedy* & \textbf{100} & 82 & 0.600 & \textbf{1.0} & \textbf{0.0} & 0.000 & 0.933 & \textbf{0.0012} \\
 & GS* & 100 & 17 & 0.544 & 2.0 & 0.0 & 0.000 & 0.941 & 0.0319 \\
 & PGD-bisect* & 89 & 89 & 1.447 & 5.0 & 3.0 & 0.000 & 1.448 & 0.5200 \\
\bottomrule
\end{tabular}}
\ifdim\wd0>\linewidth\resizebox{\linewidth}{!}{\box0}\else\box0\fi
\par\smallskip\footnotesize success = the returned point crosses the threshold (CARLA's success rate); valid = it also satisfies the actionability box and one-hot constraints. Bold = best per column within a dataset among arms with validity $\geq$ 80\%. Medians are over each arm's own valid rows; an arm with low validity is summarised on an easier subset, so compare arms with different validity through the paired table (Table~\ref{tab:carla_l2_wtl}) and the best-known table (Table~\ref{tab:carla_l2_bestknown}).
\end{table}

\begin{table}[htbp]
\centering\scriptsize\setlength{\tabcolsep}{3pt}
\caption{$\ell_2$ relative to the certified optimum (MIP-EXP or MIP-OBJ; Gurobi, 4 threads, 60\,s budget per factual, absolute optimality gap $10^{-4}$, depths 2/4/8, same 50 factuals per cell as the main run). certified n = instances on which either exact formulation proved an optimum in that dataset; arm valid = how many of those the arm returns a valid point on; at optimum = arm valid and distance $\leq$ ref $+10^{-3}$ (one-sided: below the solver's incumbent is inside its gap); within 5\% = arm valid and distance $\leq 1.05\times$ref (purely relative: where the optimum is below 0.02 the absolute tie band is the wider of the two, so the columns are not nested); both percentages over certified n; ratio columns over the arm-valid subset. A valid point can lie up to $6\times10^{-4}$ below the certified optimum (the solver's absolute gap and numerical tolerances), so ratios below 1 occur where the optimum itself is small: every ratio below 0.95 has an optimum under $2.5\times10^{-3}$.}
\label{tab:carla_l2_optimality}
\setbox0=\hbox{%
\begin{tabular}{lrrrrrrr}
\toprule
dataset & arm & certified n & arm valid & at optimum \% & within 5 \% & median ratio & max ratio \\
\midrule
adult & DASH & 184 & 184 & 76.6 & 94.6 & 1.001 & 1.162 \\
adult & DASH + data seed & 184 & 184 & 76.1 & 94.6 & 1.001 & 1.162 \\
adult & Wachter & 184 & 178 & 4.3 & 6.5 & 5.346 & 47.124 \\
adult & DiCE & 184 & 183 & 0.0 & 0.5 & 1.351 & 71.445 \\
adult & Greedy* & 184 & 184 & 0.0 & 1.1 & 1.327 & 142.891 \\
adult & GS* & 184 & 86 & 0.0 & 1.1 & 1.278 & 6.897 \\
adult & PGD-bisect* & 184 & 184 & 0.0 & 0.0 & 7.444 & 2020.784 \\
adult & FMN* & 184 & 0 & 0.0 & 0.0 & NA & NA \\
adult & NICE & 184 & 184 & 0.0 & 0.0 & 4.175 & 626.971 \\
adult & FAB & 184 & 0 & 0.0 & 0.0 & NA & NA \\
adult & DeepFool & 184 & 0 & 0.0 & 0.0 & NA & NA \\
compas & DASH & 310 & 310 & 83.2 & 96.1 & 1.002 & 1.158 \\
compas & DASH + data seed & 310 & 310 & 82.6 & 95.8 & 1.002 & 1.158 \\
compas & Wachter & 310 & 291 & 2.6 & 1.9 & 15.074 & 361.026 \\
compas & DiCE & 310 & 308 & 1.3 & 12.3 & 1.237 & 166.667 \\
compas & Greedy* & 310 & 302 & 0.0 & 3.5 & 1.598 & 333.334 \\
compas & GS* & 310 & 199 & 0.3 & 8.4 & 1.168 & 5.715 \\
compas & PGD-bisect* & 310 & 291 & 0.0 & 0.3 & 18.918 & 4714.050 \\
compas & FMN* & 310 & 0 & 0.0 & 0.0 & NA & NA \\
compas & NICE & 310 & 308 & 5.5 & 37.4 & 1.078 & 9.149 \\
compas & FAB & 310 & 1 & 0.3 & 0.3 & 0.858 & 0.858 \\
compas & DeepFool & 310 & 0 & 0.0 & 0.0 & NA & NA \\
gmc & DASH & 409 & 409 & 90.5 & 95.8 & 1.001 & 1.153 \\
gmc & DASH + data seed & 409 & 409 & 91.2 & 96.1 & 1.001 & 1.153 \\
gmc & Wachter & 409 & 409 & 97.1 & 98.3 & 1.000 & 2.220 \\
gmc & DiCE & 409 & 405 & 0.0 & 2.9 & 2.266 & 524.464 \\
gmc & Greedy* & 409 & 404 & 0.0 & 1.7 & 4.484 & 1048.928 \\
gmc & GS* & 409 & 409 & 14.4 & 20.5 & 1.124 & 34.253 \\
gmc & PGD-bisect* & 409 & 409 & 78.7 & 79.7 & 1.013 & 1.650 \\
gmc & FMN* & 409 & 1 & 0.2 & 0.2 & 1.000 & 1.000 \\
gmc & NICE & 409 & 402 & 26.7 & 22.5 & 1.119 & 24.268 \\
gmc & FAB & 409 & 7 & 1.7 & 1.7 & 0.291 & 0.443 \\
gmc & DeepFool & 409 & 2 & 0.5 & 0.0 & 1.251 & 1.383 \\
heloc & DASH & 119 & 119 & 79.0 & 97.5 & 1.002 & 1.080 \\
heloc & DASH + data seed & 119 & 119 & 79.0 & 96.6 & 1.002 & 1.080 \\
heloc & Wachter & 119 & 119 & 78.2 & 98.3 & 1.001 & 1.108 \\
heloc & DiCE & 119 & 119 & 0.0 & 0.0 & 1.454 & 76.923 \\
heloc & Greedy* & 119 & 119 & 0.0 & 0.0 & 2.523 & 153.847 \\
heloc & GS* & 119 & 119 & 0.0 & 0.0 & 1.783 & 46.314 \\
heloc & PGD-bisect* & 119 & 119 & 100.0 & 100.0 & 1.000 & 1.008 \\
heloc & FMN* & 119 & 17 & 14.3 & 14.3 & 1.000 & 1.001 \\
heloc & NICE & 119 & 119 & 0.0 & 0.0 & 1.990 & 97.371 \\
heloc & FAB & 119 & 119 & 89.9 & 100.0 & 1.001 & 1.013 \\
heloc & DeepFool & 119 & 2 & 0.8 & 0.8 & 1.061 & 1.092 \\
all & DASH & 1022 & 1022 & 84.4 & 95.9 & 1.001 & 1.162 \\
all & DASH + data seed & 1022 & 1022 & 84.4 & 95.8 & 1.001 & 1.162 \\
all & Wachter & 1022 & 997 & 49.5 & 52.5 & 1.013 & 361.026 \\
all & DiCE & 1022 & 1015 & 0.4 & 5.0 & 1.451 & 524.464 \\
all & Greedy* & 1022 & 1009 & 0.0 & 2.0 & 2.207 & 1048.928 \\
all & GS* & 1022 & 813 & 5.9 & 11.0 & 1.199 & 46.314 \\
all & PGD-bisect* & 1022 & 1003 & 43.2 & 43.6 & 1.096 & 4714.050 \\
all & FMN* & 1022 & 18 & 1.8 & 1.8 & 1.000 & 1.001 \\
all & NICE & 1022 & 1013 & 12.3 & 20.4 & 1.188 & 626.971 \\
all & FAB & 1022 & 127 & 11.3 & 12.4 & 1.000 & 1.013 \\
all & DeepFool & 1022 & 4 & 0.3 & 0.1 & 1.105 & 1.383 \\
\bottomrule
\end{tabular}}
\ifdim\wd0>\linewidth\resizebox{\linewidth}{!}{\box0}\else\box0\fi
\end{table}

\begin{table}[htbp]
\centering\scriptsize\setlength{\tabcolsep}{3pt}
\caption{MIP-EXP against DASH by depth, $\ell_2$ objective, same queries, pooled over datasets and seeds. Median times are over all of these queries, solved or not (MIP-EXP stops at its 60\,s budget).}
\label{tab:carla_l2_mip_depth}
\setbox0=\hbox{%
\begin{tabular}{lrrrrrr}
\toprule
depth & queries & MIP-EXP valid \% & MIP-EXP certified \% & MIP-EXP median time (s) & DASH valid \% & DASH median time (s) \\
\midrule
2 & 600 & 96.5 & 82.0 & 1.924 & 99.7 & 0.035 \\
4 & 600 & 76.5 & 48.7 & 57.127 & 99.5 & 0.056 \\
8 & 600 & 38.0 & 31.5 & 60.000 & 99.2 & 0.076 \\
\bottomrule
\end{tabular}}
\ifdim\wd0>\linewidth\resizebox{\linewidth}{!}{\box0}\else\box0\fi
\end{table}

\begin{table}[htbp]
\centering\scriptsize\setlength{\tabcolsep}{3pt}
\caption{Best-known counterfactual, $\ell_2$ objective: population is every tabular query of this norm (the queries column); best = smallest valid distance among DASH and the nine prior methods on that query (the exact solvers and the ablation are not in the reference); a query counts for an arm at $\tau$ iff the arm is valid and its distance $\leq(1+\tau)\times$best; column best uses the absolute tie band best $+10^{-3}$ instead, which is wider than 1\% where best is below 0.1; invalid rows never count, no penalty constant; 95\% CI = percentile cluster bootstrap over trained models (dataset $\times$ depth $\times$ seed), 2{,}000 resamples. Sorted by within 5\%; bold = best per column.}
\label{tab:carla_l2_bestknown}
\setbox0=\hbox{%
\begin{tabular}{lrrrrrrr}
\toprule
method & queries & valid \% & best & within 1 \% & within 5 \% & 95 \% CI (within 5 \%) & within 10 \% \\
\midrule
DASH & 3000 & \textbf{99.4} & \textbf{82.9} & \textbf{83.0} & \textbf{95.9} & 94.1--97.3 & \textbf{98.0} \\
Wachter & 3000 & 92.6 & 36.6 & 36.4 & 48.9 & 37.2--60.4 & 50.6 \\
PGD-bisect* & 3000 & 96.2 & 37.7 & 31.5 & 41.3 & 30.3--52.9 & 45.2 \\
FAB & 3000 & 24.3 & 18.8 & 22.0 & 23.8 & 13.7--33.9 & 24.1 \\
NICE & 3000 & 98.9 & 8.2 & 4.7 & 15.7 & 11.1--20.6 & 24.7 \\
FMN* & 3000 & 15.8 & 11.4 & 11.8 & 12.3 & 7.0--18.2 & 12.4 \\
GS* & 3000 & 70.4 & 3.8 & 1.1 & 8.4 & 5.8--11.4 & 16.2 \\
DiCE & 3000 & 95.0 & 1.0 & 1.4 & 5.8 & 3.6--8.3 & 12.8 \\
Greedy* & 3000 & 93.6 & 0.1 & 0.3 & 1.8 & 1.1--2.6 & 5.4 \\
DeepFool & 3000 & 0.5 & 0.2 & 0.0 & 0.2 & 0.0--0.3 & 0.3 \\
\bottomrule
\end{tabular}}
\ifdim\wd0>\linewidth\resizebox{\linewidth}{!}{\box0}\else\box0\fi
\end{table}

\begin{figure}[htbp]
\centering
\includegraphics[width=\linewidth]{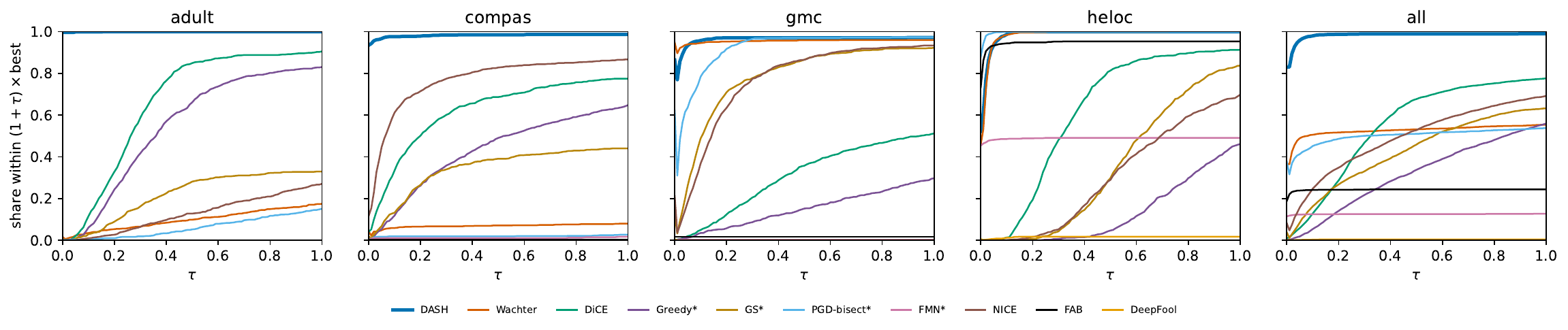}
\caption{Performance profile, $\ell_2$ objective (see Figure~\ref{fig:carla_profile} for the definition).}
\label{fig:carla_l2_profile}
\end{figure}
\begin{table}[htbp]
\centering\scriptsize\setlength{\tabcolsep}{3pt}
\caption{DASH against each arm on $\ell_2$, paired per factual over the instances where both are valid: DASH wins = DASH is closer by more than $10^{-3}$, losses = the other arm is closer; cells won = (depth, seed) cells ("all" row: dataset, depth, seed cells) with $\geq 1$ pair, where the median paired difference favours DASH by more than $10^{-3}$, printed as $k/m$; median ratio = median of DASH distance / other distance (below 1 favours DASH); two-sided Wilcoxon signed-rank on the non-tied pairs. Pairs within a cell share one trained model, so the pooled p-value overstates the evidence; cells won counts model-level agreement.}
\label{tab:carla_l2_wtl}
\setbox0=\hbox{%
\begin{tabular}{lrrrrrrrr}
\toprule
dataset & vs & n paired & DASH wins & ties & DASH losses & cells won & median ratio & Wilcoxon p \\
\midrule
adult & DASH + data seed & 750 & 31 & 696 & 23 & 0/15 & 1.000 & 0.77 \\
adult & Wachter & 645 & 633 & 10 & 2 & 15/15 & 0.245 & 1.2e-105 \\
adult & DiCE & 703 & 703 & 0 & 0 & 15/15 & 0.798 & 9.3e-117 \\
adult & Greedy* & 685 & 685 & 0 & 0 & 15/15 & 0.754 & 8e-114 \\
adult & GS* & 261 & 259 & 2 & 0 & 15/15 & 0.761 & 3.1e-44 \\
adult & PGD-bisect* & 717 & 716 & 0 & 1 & 15/15 & 0.229 & 4.9e-119 \\
adult & NICE & 750 & 750 & 0 & 0 & 15/15 & 0.312 & 2e-124 \\
adult & MIP-EXP & 286 & 83 & 154 & 49 & 3/8 & 1.000 & 1.2e-11 \\
adult & MIP-OBJ & 299 & 149 & 118 & 32 & 3/6 & 1.000 & 9.3e-28 \\
compas & DASH + data seed & 740 & 16 & 718 & 6 & 0/15 & 1.000 & 0.26 \\
compas & Wachter & 647 & 614 & 30 & 3 & 15/15 & 0.102 & 1.5e-102 \\
compas & DiCE & 675 & 639 & 22 & 14 & 15/15 & 0.865 & 1.2e-104 \\
compas & Greedy* & 653 & 648 & 4 & 1 & 15/15 & 0.711 & 2.1e-107 \\
compas & GS* & 358 & 345 & 11 & 2 & 15/15 & 0.857 & 1.4e-58 \\
compas & PGD-bisect* & 676 & 673 & 3 & 0 & 15/15 & 0.084 & 7.3e-112 \\
compas & FMN* & 104 & 103 & 0 & 1 & 9/9 & 0.223 & 9.3e-19 \\
compas & NICE & 728 & 640 & 58 & 30 & 15/15 & 0.943 & 4.2e-93 \\
compas & FAB & 2 & 0 & 2 & 0 & 0/2 & 1.030 & NA \\
compas & MIP-EXP & 362 & 43 & 265 & 54 & 0/9 & 1.000 & 0.018 \\
compas & MIP-OBJ & 277 & 106 & 142 & 29 & 3/6 & 1.000 & 1e-19 \\
gmc & DASH + data seed & 742 & 8 & 724 & 10 & 0/15 & 1.000 & 0.73 \\
gmc & Wachter & 736 & 22 & 625 & 89 & 0/15 & 1.001 & 8.3e-05 \\
gmc & DiCE & 722 & 719 & 2 & 1 & 15/15 & 0.545 & 4.9e-119 \\
gmc & Greedy* & 720 & 717 & 2 & 1 & 15/15 & 0.288 & 4.3e-119 \\
gmc & GS* & 742 & 605 & 117 & 20 & 15/15 & 0.912 & 1.5e-89 \\
gmc & PGD-bisect* & 742 & 233 & 471 & 38 & 4/15 & 0.985 & 1.4e-24 \\
gmc & FMN* & 1 & 0 & 0 & 1 & 0/1 & 1.123 & NA \\
gmc & NICE & 735 & 565 & 166 & 4 & 15/15 & 0.893 & 1.2e-92 \\
gmc & FAB & 13 & 0 & 13 & 0 & 0/9 & 3.666 & NA \\
gmc & DeepFool & 3 & 0 & 3 & 0 & 0/3 & 0.474 & NA \\
gmc & MIP-EXP & 435 & 25 & 371 & 39 & 0/9 & 0.999 & 0.17 \\
gmc & MIP-OBJ & 296 & 50 & 236 & 10 & 1/8 & 1.000 & 1.4e-09 \\
heloc & DASH + data seed & 750 & 90 & 600 & 60 & 0/15 & 1.000 & 0.03 \\
heloc & Wachter & 750 & 264 & 311 & 175 & 0/15 & 0.999 & 4.5e-08 \\
heloc & DiCE & 749 & 749 & 0 & 0 & 15/15 & 0.775 & 2.9e-124 \\
heloc & Greedy* & 749 & 749 & 0 & 0 & 15/15 & 0.493 & 2.9e-124 \\
heloc & GS* & 750 & 750 & 0 & 0 & 15/15 & 0.627 & 2e-124 \\
heloc & PGD-bisect* & 750 & 28 & 390 & 332 & 0/15 & 1.005 & 9.5e-43 \\
heloc & FMN* & 368 & 18 & 138 & 212 & 0/15 & 1.007 & 1.1e-25 \\
heloc & NICE & 750 & 750 & 0 & 0 & 15/15 & 0.596 & 2e-124 \\
heloc & FAB & 715 & 51 & 394 & 270 & 0/15 & 1.003 & 4e-25 \\
heloc & DeepFool & 13 & 8 & 5 & 0 & 6/9 & 0.920 & 0.0078 \\
heloc & MIP-EXP & 183 & 93 & 78 & 12 & 4/9 & 0.990 & 3.8e-16 \\
heloc & MIP-OBJ & 297 & 146 & 108 & 43 & 3/6 & 1.000 & 3.6e-26 \\
all & DASH + data seed & 2982 & 145 & 2738 & 99 & 0/60 & 1.000 & 0.022 \\
all & Wachter & 2778 & 1533 & 976 & 269 & 30/60 & 0.979 & 1.2e-245 \\
all & DiCE & 2849 & 2810 & 24 & 15 & 60/60 & 0.772 & $<10^{-300}$ \\
all & Greedy* & 2807 & 2799 & 6 & 2 & 60/60 & 0.572 & $<10^{-300}$ \\
all & GS* & 2111 & 1959 & 130 & 22 & 60/60 & 0.756 & $<10^{-300}$ \\
all & PGD-bisect* & 2885 & 1650 & 864 & 371 & 34/60 & 0.863 & 1.8e-261 \\
all & FMN* & 473 & 121 & 138 & 214 & 9/25 & 1.003 & 0.058 \\
all & NICE & 2963 & 2705 & 224 & 34 & 60/60 & 0.709 & $<10^{-300}$ \\
all & FAB & 730 & 51 & 409 & 270 & 0/26 & 1.003 & 4e-25 \\
all & DeepFool & 16 & 8 & 8 & 0 & 6/12 & 0.913 & 0.0078 \\
all & MIP-EXP & 1266 & 244 & 868 & 154 & 7/35 & 1.000 & 1.5e-25 \\
all & MIP-OBJ & 1169 & 451 & 604 & 114 & 10/26 & 1.000 & 1.6e-78 \\
\bottomrule
\end{tabular}}
\ifdim\wd0>\linewidth\resizebox{\linewidth}{!}{\box0}\else\box0\fi
\end{table}

\begin{table}[htbp]
\centering\scriptsize\setlength{\tabcolsep}{3pt}
\caption{Validity (\%) per arm and cell, $\ell_2$ objective.}
\label{tab:carla_l2_validity}
\setbox0=\hbox{%
\begin{tabular}{lrrrrrrrrrrrrrr}
\toprule
dataset & depth & DASH & DASH + data seed & Wachter & DiCE & Greedy* & GS* & PGD-bisect* & FMN* & NICE & FAB & DeepFool & MIP-EXP & MIP-OBJ \\
\midrule
adult & 2 & 100 & 100 & 96 & 99 & 100 & 41 & 100 & 0 & 100 & 0 & 0 & 100 & 99 \\
 & 4 & 100 & 100 & 88 & 97 & 100 & 57 & 100 & 0 & 100 & 0 & 0 & 88 & 100 \\
 & 8 & 100 & 100 & 77 & 92 & 82 & 17 & 89 & 0 & 100 & 0 & 0 & 3 & 0 \\
 & 16 & 100 & 100 & 87 & 91 & 91 & 48 & 97 & 0 & 100 & 0 & 0 & -- & -- \\
 & 32 & 100 & 100 & 82 & 89 & 83 & 12 & 92 & 0 & 100 & 0 & 0 & -- & -- \\
compas & 2 & 99 & 99 & 88 & 98 & 97 & 69 & 88 & 0 & 98 & 0 & 0 & 99 & 99 \\
 & 4 & 98 & 98 & 86 & 97 & 97 & 47 & 87 & 0 & 95 & 0 & 0 & 94 & 85 \\
 & 8 & 98 & 98 & 88 & 93 & 85 & 37 & 89 & 11 & 98 & 1 & 0 & 48 & 0 \\
 & 16 & 98 & 98 & 91 & 89 & 79 & 42 & 91 & 23 & 97 & 1 & 0 & -- & -- \\
 & 32 & 100 & 100 & 78 & 73 & 77 & 44 & 95 & 35 & 99 & 0 & 0 & -- & -- \\
gmc & 2 & 99 & 99 & 99 & 99 & 99 & 99 & 99 & 0 & 97 & 1 & 0 & 99 & 97 \\
 & 4 & 100 & 100 & 99 & 99 & 99 & 100 & 100 & 0 & 98 & 3 & 1 & 100 & 98 \\
 & 8 & 99 & 99 & 97 & 91 & 91 & 99 & 99 & 1 & 98 & 1 & 1 & 91 & 2 \\
 & 16 & 99 & 99 & 99 & 99 & 99 & 99 & 99 & 0 & 99 & 3 & 0 & -- & -- \\
 & 32 & 98 & 98 & 96 & 93 & 93 & 98 & 98 & 0 & 98 & 1 & 1 & -- & -- \\
heloc & 2 & 100 & 100 & 100 & 100 & 100 & 100 & 100 & 24 & 100 & 100 & 1 & 87 & 99 \\
 & 4 & 100 & 100 & 100 & 100 & 100 & 100 & 100 & 41 & 100 & 100 & 1 & 24 & 99 \\
 & 8 & 100 & 100 & 100 & 99 & 99 & 100 & 100 & 51 & 100 & 96 & 1 & 11 & 0 \\
 & 16 & 100 & 100 & 100 & 100 & 100 & 100 & 100 & 65 & 100 & 94 & 1 & -- & -- \\
 & 32 & 100 & 100 & 100 & 100 & 100 & 100 & 100 & 64 & 100 & 87 & 4 & -- & -- \\
\bottomrule
\end{tabular}}
\ifdim\wd0>\linewidth\resizebox{\linewidth}{!}{\box0}\else\box0\fi
\end{table}

\begin{table}[htbp]
\centering\scriptsize\setlength{\tabcolsep}{3pt}
\caption{MNIST (3 vs 8, $n=784$, depth 8, width 256, 150 factuals), $\ell_2$ objective. Greedy* did not run at this scale; under $\ell_1$ the norm-adapted GS* and PGD-bisect* runs cover 70 of the 150 factuals (validity over the arm's own rows). PBMC3k is reported in the main text, after the shared marker completion. DASH = the CPU implementation; NA = no valid counterfactual, or a metric the run did not record.}
\label{tab:carla_l2_large}
\setbox0=\hbox{%
\begin{tabular}{lrrrrrrrrr}
\toprule
dataset & method & success \% & valid \% & $\ell_2$ & l0 & redund. & yNN & yNN-dist & time (s) \\
\midrule
mnist & DASH & \textbf{100} & \textbf{100} & 1.196 & 473.0 & 71.5 & 0.000 & 5.139 & 74.1663 \\
mnist & DASH + data seed & \textbf{100} & \textbf{100} & 1.197 & 473.5 & 75.0 & 0.000 & 5.128 & 73.5537 \\
mnist & Wachter & \textbf{100} & \textbf{100} & 1.398 & 764.0 & 387.5 & 0.000 & 5.136 & 2.2161 \\
mnist & DiCE & \textbf{100} & \textbf{100} & 1.437 & 201.5 & \textbf{1.0} & 0.000 & 5.153 & \textbf{0.0455} \\
mnist & GS* & \textbf{100} & \textbf{100} & 3.375 & 37.0 & 24.0 & 0.000 & 6.109 & 2.2197 \\
mnist & PGD-bisect* & \textbf{100} & \textbf{100} & 1.102 & 478.0 & 62.0 & 0.000 & \textbf{5.079} & 5.2108 \\
mnist & FMN* & 55 & 55 & 1.277 & 473.0 & 34.0 & 0.000 & 5.163 & 0.8296 \\
mnist & NICE & \textbf{100} & \textbf{100} & 2.959 & \textbf{22.0} & 6.0 & \textbf{0.200} & 5.626 & 0.7318 \\
mnist & FAB & \textbf{100} & \textbf{100} & \textbf{1.096} & 476.0 & 32.0 & 0.000 & \textbf{5.079} & 0.1364 \\
mnist & DeepFool & 100 & 0 & NA & NA & NA & NA & NA & NA \\
\bottomrule
\end{tabular}}
\ifdim\wd0>\linewidth\resizebox{\linewidth}{!}{\box0}\else\box0\fi
\end{table}

\begin{figure}[htbp]
\centering
\includegraphics[width=\linewidth]{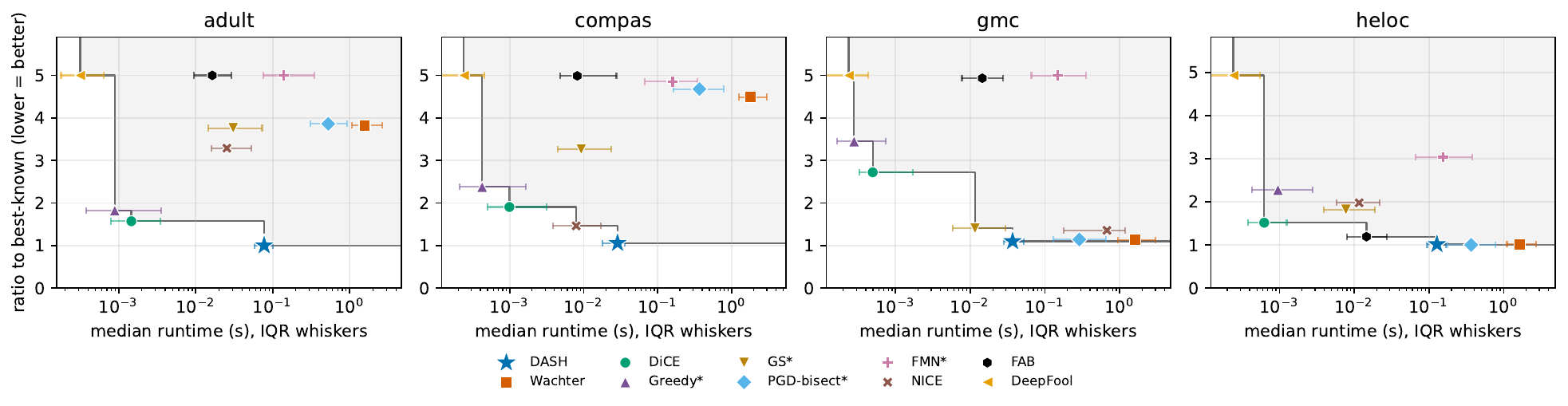}
\caption{Runtime-quality trade-off, $\ell_2$ objective (see Figure~\ref{fig:carla_pareto} for the definition).}
\label{fig:carla_l2_pareto}
\end{figure}
\begin{figure}[htbp]
\centering
\includegraphics[width=\linewidth]{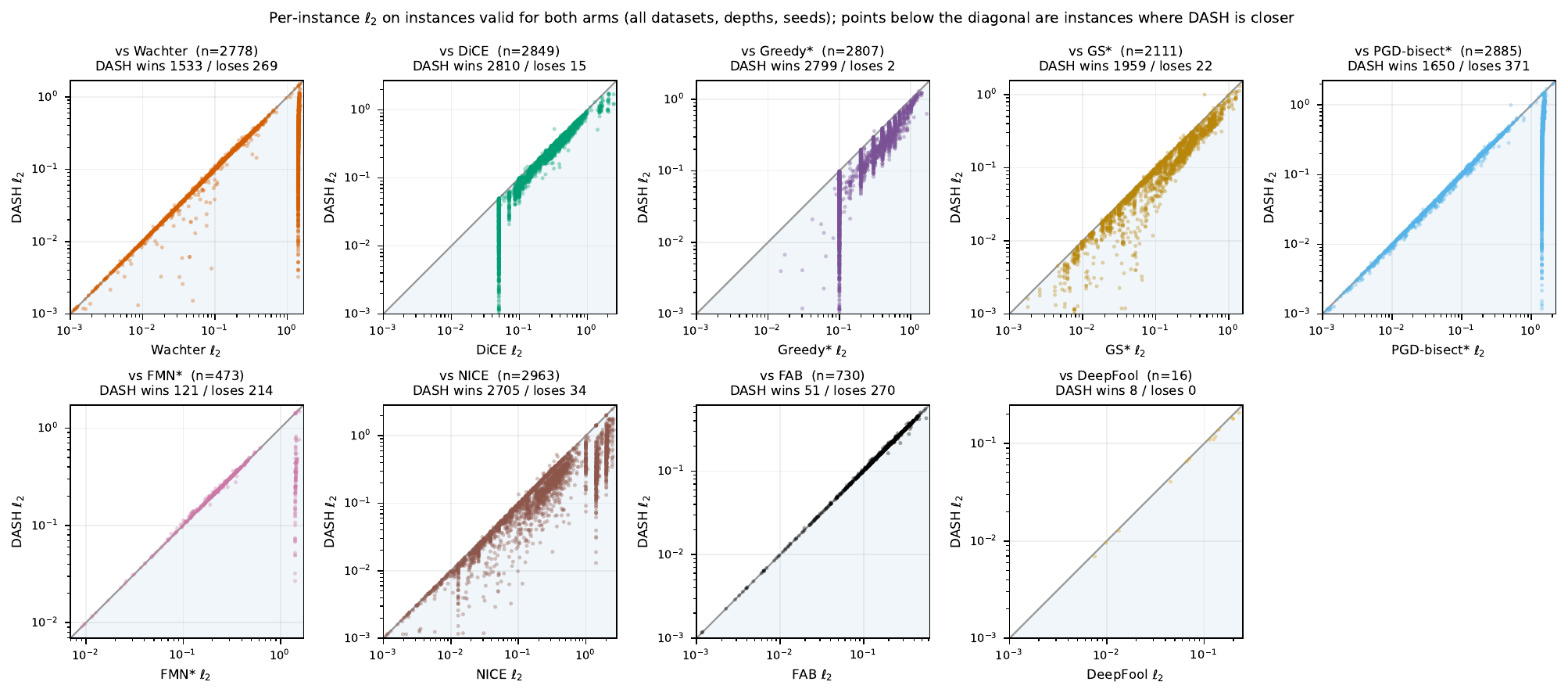}
\caption{Per-instance paired comparison, $\ell_2$ objective (see Figure~\ref{fig:carla_paired} for the definition).}
\label{fig:carla_l2_paired}
\end{figure}
\begin{figure}[htbp]
\centering
\includegraphics[width=\linewidth]{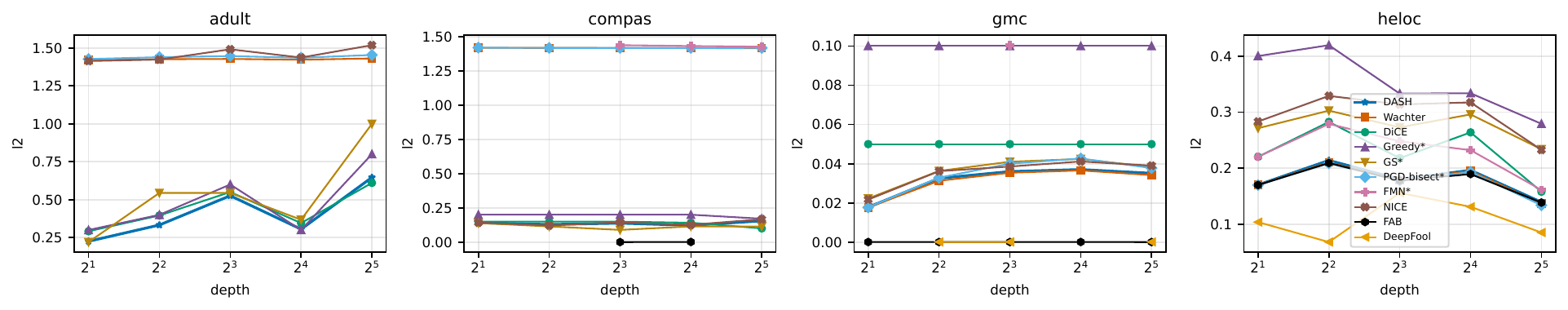}\\
\includegraphics[width=\linewidth]{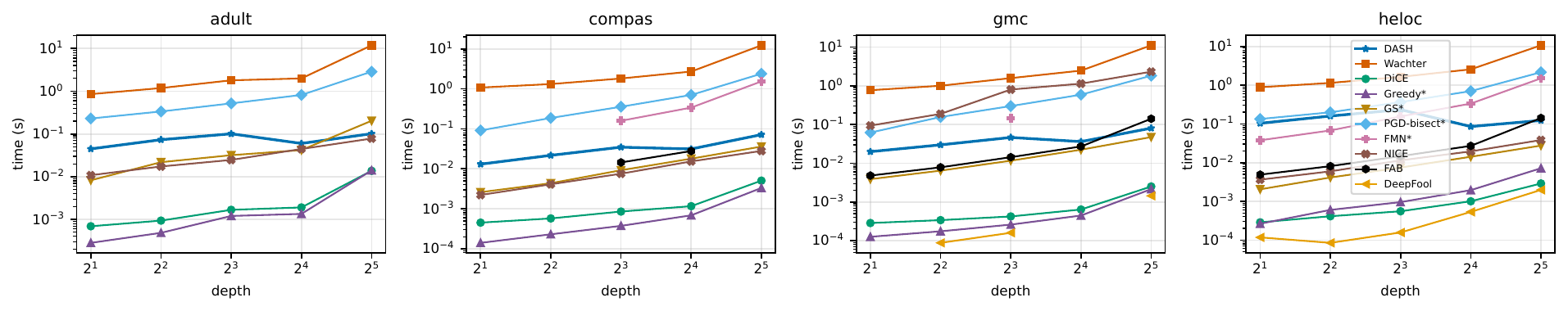}
\caption{Median $\ell_2$ and runtime versus depth, $\ell_2$ objective.}
\label{fig:carla_l2_depth}
\end{figure}
\clearpage
{\scriptsize\setlength{\tabcolsep}{3pt}
\begin{longtable}{lrrrrrrrrrrrr}
\caption{All tabular cells, $\ell_2$ objective (medians over valid rows; bold = best per column within a cell among arms with validity $\geq$ 80\%). Medians are over each arm's own valid rows; an arm with low validity is summarised on an easier subset, so compare arms with different validity through the paired table (Table~\ref{tab:carla_l2_wtl}) and the best-known table (Table~\ref{tab:carla_l2_bestknown}).} \label{tab:carla_l2_per_cell} \\
\toprule
dataset & depth & method & success \% & valid \% & $\ell_2$ & l0 & redund. & yNN & yNN-dist & time (s) & $\ell_1$ & $\ell_\infty$ \\
\midrule
\endfirsthead
\caption[]{All tabular cells, $\ell_2$ objective (medians over valid rows; bold = best per column within a cell among arms with validity $\geq$ 80\%). Medians are over each arm's own valid rows; an arm with low validity is summarised on an easier subset, so compare arms with different validity through the paired table (Table~\ref{tab:carla_l2_wtl}) and the best-known table (Table~\ref{tab:carla_l2_bestknown}).} \\
\toprule
dataset & depth & method & success \% & valid \% & $\ell_2$ & l0 & redund. & yNN & yNN-dist & time (s) & $\ell_1$ & $\ell_\infty$ \\
\midrule
\endhead
\midrule
\multicolumn{13}{r}{Continued on next page} \\
\midrule
\endfoot
\bottomrule
\endlastfoot
adult & 2 & DASH & \textbf{100} & \textbf{100} & \textbf{0.225} & 3.0 & \textbf{0.0} & \textbf{0.000} & \textbf{0.333} & 0.0448 & 0.287 & 0.216 \\
adult & 2 & DASH + data seed & \textbf{100} & \textbf{100} & 0.227 & 3.0 & \textbf{0.0} & \textbf{0.000} & \textbf{0.333} & 0.0502 & \textbf{0.282} & 0.218 \\
adult & 2 & Wachter & 97 & 96 & 1.420 & 5.0 & 3.0 & \textbf{0.000} & 1.425 & 0.8565 & 2.224 & 1.000 \\
adult & 2 & DiCE & \textbf{100} & 99 & 0.292 & 3.0 & \textbf{0.0} & \textbf{0.000} & 0.408 & 0.0007 & 0.500 & \textbf{0.190} \\
adult & 2 & Greedy* & \textbf{100} & \textbf{100} & 0.300 & \textbf{1.0} & \textbf{0.0} & \textbf{0.000} & 0.400 & \textbf{0.0003} & 0.300 & 0.300 \\
adult & 2 & GS* & 100 & 41 & 0.220 & 1.0 & 0.0 & 0.000 & 0.296 & 0.0082 & 0.220 & 0.220 \\
adult & 2 & PGD-bisect* & \textbf{100} & \textbf{100} & 1.427 & 4.0 & 2.0 & \textbf{0.000} & 1.429 & 0.2302 & 2.264 & 1.000 \\
adult & 2 & FMN* & 0 & 0 & NA & NA & NA & NA & NA & NA & NA & NA \\
\end{longtable}
}

\clearpage

\subsection{$\ell_\infty$ objective}
\begin{table}[htbp]
\centering\scriptsize\setlength{\tabcolsep}{3pt}
\caption{CARLA axes at depth 8 ($\ell_\infty$ objective, basic actionability; seeds 17/29/43).}
\label{tab:carla_linf_main_d8}
\setbox0=\hbox{%
\begin{tabular}{lrrrrrrrrr}
\toprule
dataset & method & success \% & valid \% & $\ell_\infty$ & l0 & redund. & yNN & yNN-dist & time (s) \\
\midrule
adult & DASH & \textbf{100} & \textbf{100} & 0.291 & 4.0 & \textbf{0.0} & 0.000 & 0.950 & 0.0997 \\
 & DASH + data seed & \textbf{100} & \textbf{100} & 0.291 & 4.0 & \textbf{0.0} & 0.000 & 0.950 & 0.1052 \\
 & Wachter & 78 & 77 & 1.000 & 5.0 & 3.0 & 0.000 & 1.430 & 1.2503 \\
 & DiCE & \textbf{100} & 83 & \textbf{0.276} & 4.0 & \textbf{0.0} & 0.000 & \textbf{0.765} & 0.0017 \\
 & Greedy* & \textbf{100} & 82 & 0.600 & \textbf{1.0} & \textbf{0.0} & 0.000 & 0.933 & \textbf{0.0012} \\
 & GS* & 100 & 25 & 0.321 & 2.0 & 0.0 & 0.000 & 0.484 & 0.0086 \\
 & PGD-bisect* & 73 & 73 & 1.000 & 5.0 & 4.0 & 0.000 & 1.429 & 0.3121 \\
\bottomrule
\end{tabular}}
\ifdim\wd0>\linewidth\resizebox{\linewidth}{!}{\box0}\else\box0\fi
\par\smallskip\footnotesize success = the returned point crosses the threshold (CARLA's success rate); valid = it also satisfies the actionability box and one-hot constraints. Bold = best per column within a dataset among arms with validity $\geq$ 80\%. Medians are over each arm's own valid rows; an arm with low validity is summarised on an easier subset, so compare arms with different validity through the paired table (Table~\ref{tab:carla_linf_wtl}) and the best-known table (Table~\ref{tab:carla_linf_bestknown}).
\end{table}

\begin{table}[htbp]
\centering\scriptsize\setlength{\tabcolsep}{3pt}
\caption{$\ell_\infty$ relative to the certified optimum (MIP-EXP or MIP-OBJ; Gurobi, 4 threads, 60\,s budget per factual, absolute optimality gap $10^{-4}$, depths 2/4/8, same 50 factuals per cell as the main run). certified n = instances on which either exact formulation proved an optimum in that dataset; arm valid = how many of those the arm returns a valid point on; at optimum = arm valid and distance $\leq$ ref $+10^{-3}$ (one-sided: below the solver's incumbent is inside its gap); within 5\% = arm valid and distance $\leq 1.05\times$ref (purely relative: where the optimum is below 0.02 the absolute tie band is the wider of the two, so the columns are not nested); both percentages over certified n; ratio columns over the arm-valid subset. A valid point can lie up to $6\times10^{-4}$ below the certified optimum (the solver's absolute gap and numerical tolerances), so ratios below 1 occur where the optimum itself is small: every ratio below 0.95 has an optimum under $2.5\times10^{-3}$.}
\label{tab:carla_linf_optimality}
\setbox0=\hbox{%
\begin{tabular}{lrrrrrrr}
\toprule
dataset & arm & certified n & arm valid & at optimum \% & within 5 \% & median ratio & max ratio \\
\midrule
adult & DASH & 279 & 279 & 93.2 & 99.3 & 1.000 & 1.106 \\
adult & DASH + data seed & 279 & 279 & 93.5 & 99.3 & 1.000 & 1.106 \\
adult & Wachter & 279 & 259 & 0.4 & 0.0 & 4.419 & 13.879 \\
adult & DiCE & 279 & 275 & 1.1 & 20.8 & 1.122 & 130.887 \\
adult & Greedy* & 279 & 279 & 0.0 & 0.0 & 1.957 & 261.774 \\
adult & GS* & 279 & 191 & 0.0 & 0.0 & 1.849 & 376.044 \\
adult & PGD-bisect* & 279 & 250 & 0.0 & 0.0 & 6.637 & 2617.742 \\
adult & FMN* & 279 & 227 & 0.0 & 0.0 & 6.427 & 2617.742 \\
adult & NICE & 279 & 279 & 0.0 & 0.0 & 4.688 & 1148.601 \\
adult & FAB & 279 & 0 & 0.0 & 0.0 & NA & NA \\
adult & DeepFool & 279 & 0 & 0.0 & 0.0 & NA & NA \\
compas & DASH & 326 & 326 & 85.0 & 95.1 & 1.001 & 1.217 \\
compas & DASH + data seed & 326 & 326 & 85.0 & 95.1 & 1.002 & 1.217 \\
compas & Wachter & 326 & 306 & 22.1 & 23.0 & 3.190 & 67.114 \\
compas & DiCE & 326 & 323 & 0.6 & 13.5 & 1.240 & 200.000 \\
compas & Greedy* & 326 & 318 & 0.0 & 4.0 & 1.667 & 400.000 \\
compas & GS* & 326 & 205 & 0.3 & 3.7 & 1.446 & 50.621 \\
compas & PGD-bisect* & 326 & 302 & 0.3 & 0.3 & 13.611 & 4000.000 \\
compas & FMN* & 326 & 275 & 0.0 & 0.0 & 14.255 & 4000.000 \\
compas & NICE & 326 & 324 & 3.1 & 35.3 & 1.088 & 11.172 \\
compas & FAB & 326 & 0 & 0.0 & 0.0 & NA & NA \\
compas & DeepFool & 326 & 0 & 0.0 & 0.0 & NA & NA \\
gmc & DASH & 419 & 419 & 82.8 & 82.3 & 1.001 & 1.615 \\
gmc & DASH + data seed & 419 & 419 & 83.5 & 82.3 & 1.001 & 1.615 \\
gmc & Wachter & 419 & 419 & 28.6 & 22.0 & 1.114 & 1.736 \\
gmc & DiCE & 419 & 414 & 0.0 & 1.7 & 2.681 & 583.733 \\
gmc & Greedy* & 419 & 412 & 0.0 & 0.7 & 5.340 & 1167.466 \\
gmc & GS* & 419 & 419 & 3.1 & 6.2 & 1.395 & 91.191 \\
gmc & PGD-bisect* & 419 & 419 & 99.0 & 98.3 & 1.001 & 1.358 \\
gmc & FMN* & 419 & 306 & 69.2 & 70.4 & 1.000 & 1.169 \\
gmc & NICE & 419 & 412 & 16.0 & 4.3 & 1.223 & 25.589 \\
gmc & FAB & 419 & 7 & 1.7 & 1.7 & 0.205 & 0.280 \\
gmc & DeepFool & 419 & 0 & 0.0 & 0.0 & NA & NA \\
heloc & DASH & 331 & 331 & 88.2 & 96.1 & 1.002 & 1.134 \\
heloc & DASH + data seed & 331 & 331 & 87.0 & 95.8 & 1.002 & 1.134 \\
heloc & Wachter & 331 & 331 & 0.6 & 0.3 & 1.266 & 6.272 \\
heloc & DiCE & 331 & 331 & 0.6 & 1.8 & 1.622 & 250.000 \\
heloc & Greedy* & 331 & 331 & 0.0 & 0.0 & 7.707 & 500.000 \\
heloc & GS* & 331 & 331 & 0.0 & 0.0 & 4.312 & 352.063 \\
heloc & PGD-bisect* & 331 & 331 & 100.0 & 99.7 & 1.001 & 1.221 \\
heloc & FMN* & 331 & 330 & 96.4 & 99.7 & 1.000 & 1.048 \\
heloc & NICE & 331 & 331 & 0.0 & 0.0 & 4.241 & 316.456 \\
heloc & FAB & 331 & 330 & 95.8 & 99.4 & 1.001 & 1.051 \\
heloc & DeepFool & 331 & 7 & 1.2 & 0.9 & 1.056 & 1.112 \\
all & DASH & 1355 & 1355 & 86.8 & 92.3 & 1.001 & 1.615 \\
all & DASH + data seed & 1355 & 1355 & 86.8 & 92.2 & 1.001 & 1.615 \\
all & Wachter & 1355 & 1315 & 14.4 & 12.4 & 1.281 & 67.114 \\
all & DiCE & 1355 & 1343 & 0.5 & 8.5 & 1.502 & 583.733 \\
all & Greedy* & 1355 & 1340 & 0.0 & 1.2 & 3.653 & 1167.466 \\
all & GS* & 1355 & 1146 & 1.0 & 2.8 & 2.027 & 376.044 \\
all & PGD-bisect* & 1355 & 1302 & 55.1 & 54.8 & 1.005 & 4000.000 \\
all & FMN* & 1355 & 1138 & 44.9 & 46.1 & 1.013 & 4000.000 \\
all & NICE & 1355 & 1346 & 5.7 & 9.8 & 1.645 & 1148.601 \\
all & FAB & 1355 & 337 & 23.9 & 24.8 & 1.001 & 1.051 \\
all & DeepFool & 1355 & 7 & 0.3 & 0.2 & 1.056 & 1.112 \\
\bottomrule
\end{tabular}}
\ifdim\wd0>\linewidth\resizebox{\linewidth}{!}{\box0}\else\box0\fi
\end{table}

\begin{table}[htbp]
\centering\scriptsize\setlength{\tabcolsep}{3pt}
\caption{MIP-EXP against DASH by depth, $\ell_\infty$ objective, same queries, pooled over datasets and seeds. Median times are over all of these queries, solved or not (MIP-EXP stops at its 60\,s budget).}
\label{tab:carla_linf_mip_depth}
\setbox0=\hbox{%
\begin{tabular}{lrrrrrr}
\toprule
depth & queries & MIP-EXP valid \% & MIP-EXP certified \% & MIP-EXP median time (s) & DASH valid \% & DASH median time (s) \\
\midrule
2 & 600 & 99.7 & 99.7 & 1.007 & 99.7 & 0.031 \\
4 & 600 & 96.3 & 66.5 & 9.036 & 99.5 & 0.059 \\
8 & 600 & 51.0 & 45.0 & 60.000 & 99.2 & 0.073 \\
\bottomrule
\end{tabular}}
\ifdim\wd0>\linewidth\resizebox{\linewidth}{!}{\box0}\else\box0\fi
\end{table}

\begin{table}[htbp]
\centering\scriptsize\setlength{\tabcolsep}{3pt}
\caption{Best-known counterfactual, $\ell_\infty$ objective: population is every tabular query of this norm (the queries column); best = smallest valid distance among DASH and the nine prior methods on that query (the exact solvers and the ablation are not in the reference); a query counts for an arm at $\tau$ iff the arm is valid and its distance $\leq(1+\tau)\times$best; column best uses the absolute tie band best $+10^{-3}$ instead, which is wider than 1\% where best is below 0.1; invalid rows never count, no penalty constant; 95\% CI = percentile cluster bootstrap over trained models (dataset $\times$ depth $\times$ seed), 2{,}000 resamples. Sorted by within 5\%; bold = best per column.}
\label{tab:carla_linf_bestknown}
\setbox0=\hbox{%
\begin{tabular}{lrrrrrrr}
\toprule
method & queries & valid \% & best & within 1 \% & within 5 \% & 95 \% CI (within 5 \%) & within 10 \% \\
\midrule
DASH & 3000 & \textbf{99.4} & \textbf{88.6} & \textbf{81.4} & \textbf{92.3} & 89.6--94.7 & \textbf{95.8} \\
PGD-bisect* & 3000 & 92.3 & 49.7 & 47.2 & 49.3 & 36.9--62.0 & 49.4 \\
FMN* & 3000 & 81.9 & 39.8 & 35.9 & 41.0 & 30.3--51.4 & 41.7 \\
FAB & 3000 & 24.4 & 21.5 & 19.4 & 23.9 & 13.6--34.1 & 24.3 \\
NICE & 3000 & 98.9 & 6.1 & 4.0 & 11.9 & 7.6--16.5 & 17.7 \\
DiCE & 3000 & 93.6 & 1.2 & 2.1 & 11.3 & 8.6--14.2 & 20.5 \\
Wachter & 3000 & 92.7 & 10.1 & 4.6 & 10.1 & 7.2--13.1 & 17.6 \\
GS* & 3000 & 72.2 & 0.9 & 0.5 & 2.8 & 1.7--4.0 & 6.9 \\
Greedy* & 3000 & 93.6 & 0.1 & 0.2 & 1.4 & 0.7--2.2 & 2.9 \\
DeepFool & 3000 & 0.4 & 0.2 & 0.0 & 0.1 & 0.0--0.3 & 0.3 \\
\bottomrule
\end{tabular}}
\ifdim\wd0>\linewidth\resizebox{\linewidth}{!}{\box0}\else\box0\fi
\end{table}

\begin{figure}[htbp]
\centering
\includegraphics[width=\linewidth]{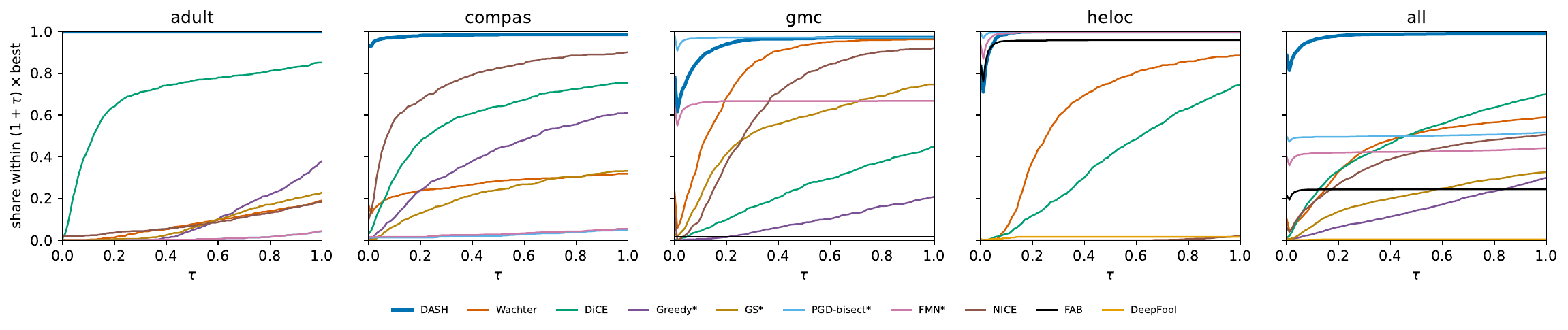}
\caption{Performance profile, $\ell_\infty$ objective (see Figure~\ref{fig:carla_profile} for the definition).}
\label{fig:carla_linf_profile}
\end{figure}
\begin{table}[htbp]
\centering\scriptsize\setlength{\tabcolsep}{3pt}
\caption{DASH against each arm on $\ell_\infty$, paired per factual over the instances where both are valid: DASH wins = DASH is closer by more than $10^{-3}$, losses = the other arm is closer; cells won = (depth, seed) cells ("all" row: dataset, depth, seed cells) with $\geq 1$ pair, where the median paired difference favours DASH by more than $10^{-3}$, printed as $k/m$; median ratio = median of DASH distance / other distance (below 1 favours DASH); two-sided Wilcoxon signed-rank on the non-tied pairs. Pairs within a cell share one trained model, so the pooled p-value overstates the evidence; cells won counts model-level agreement.}
\label{tab:carla_linf_wtl}
\setbox0=\hbox{%
\begin{tabular}{lrrrrrrrr}
\toprule
dataset & vs & n paired & DASH wins & ties & DASH losses & cells won & median ratio & Wilcoxon p \\
\midrule
adult & DASH + data seed & 750 & 4 & 741 & 5 & 0/15 & 1.000 & 0.73 \\
adult & Wachter & 646 & 643 & 3 & 0 & 15/15 & 0.256 & 5.7e-107 \\
adult & DiCE & 677 & 670 & 6 & 1 & 15/15 & 0.907 & 1.6e-111 \\
adult & Greedy* & 685 & 685 & 0 & 0 & 15/15 & 0.468 & 8e-114 \\
adult & GS* & 338 & 338 & 0 & 0 & 15/15 & 0.501 & 3.7e-57 \\
adult & PGD-bisect* & 612 & 612 & 0 & 0 & 15/15 & 0.185 & 6.6e-102 \\
adult & FMN* & 573 & 573 & 0 & 0 & 15/15 & 0.189 & 1.5e-95 \\
adult & NICE & 750 & 736 & 14 & 0 & 15/15 & 0.260 & 3.8e-122 \\
adult & MIP-EXP & 310 & 71 & 220 & 19 & 3/9 & 1.000 & 1.8e-11 \\
adult & MIP-OBJ & 298 & 11 & 269 & 18 & 0/6 & 1.000 & 0.34 \\
compas & DASH + data seed & 740 & 6 & 731 & 3 & 0/15 & 1.000 & 0.73 \\
compas & Wachter & 650 & 522 & 121 & 7 & 15/15 & 0.289 & 1.5e-87 \\
compas & DiCE & 661 & 631 & 19 & 11 & 15/15 & 0.855 & 1.5e-104 \\
compas & Greedy* & 653 & 648 & 4 & 1 & 15/15 & 0.672 & 2.5e-107 \\
compas & GS* & 336 & 334 & 2 & 0 & 15/15 & 0.704 & 1.7e-56 \\
compas & PGD-bisect* & 664 & 654 & 10 & 0 & 15/15 & 0.112 & 9.2e-109 \\
compas & FMN* & 618 & 609 & 8 & 1 & 15/15 & 0.110 & 1.4e-101 \\
compas & NICE & 728 & 651 & 46 & 31 & 15/15 & 0.936 & 8.5e-94 \\
compas & MIP-EXP & 359 & 41 & 272 & 46 & 0/9 & 1.000 & 0.0048 \\
compas & MIP-OBJ & 293 & 6 & 241 & 46 & 0/6 & 1.001 & 9e-05 \\
gmc & DASH + data seed & 742 & 7 & 722 & 13 & 0/15 & 1.000 & 0.09 \\
gmc & Wachter & 736 & 474 & 216 & 46 & 14/15 & 0.918 & 7.2e-52 \\
gmc & DiCE & 722 & 715 & 5 & 2 & 15/15 & 0.468 & 5.6e-119 \\
gmc & Greedy* & 720 & 717 & 2 & 1 & 15/15 & 0.246 & 3.5e-119 \\
gmc & GS* & 742 & 649 & 49 & 44 & 15/15 & 0.794 & 1e-84 \\
gmc & PGD-bisect* & 742 & 4 & 585 & 153 & 0/15 & 1.000 & 3.1e-25 \\
gmc & FMN* & 514 & 12 & 432 & 70 & 0/15 & 1.001 & 5.3e-10 \\
gmc & NICE & 735 & 605 & 117 & 13 & 15/15 & 0.819 & 2.4e-95 \\
gmc & FAB & 13 & 0 & 13 & 0 & 0/9 & 5.132 & NA \\
gmc & MIP-EXP & 434 & 10 & 349 & 75 & 0/9 & 1.000 & 2.1e-06 \\
gmc & MIP-OBJ & 293 & 3 & 243 & 47 & 0/8 & 1.002 & 1.3e-07 \\
heloc & DASH + data seed & 750 & 30 & 689 & 31 & 0/15 & 1.000 & 0.92 \\
heloc & Wachter & 750 & 747 & 2 & 1 & 15/15 & 0.808 & 4.3e-124 \\
heloc & DiCE & 749 & 747 & 2 & 0 & 15/15 & 0.625 & 6.1e-124 \\
heloc & Greedy* & 749 & 749 & 0 & 0 & 15/15 & 0.155 & 2.9e-124 \\
heloc & GS* & 750 & 750 & 0 & 0 & 15/15 & 0.258 & 2e-124 \\
heloc & PGD-bisect* & 750 & 3 & 625 & 122 & 0/15 & 1.002 & 3.5e-19 \\
heloc & FMN* & 749 & 22 & 629 & 98 & 0/15 & 1.001 & 1.9e-07 \\
heloc & NICE & 750 & 750 & 0 & 0 & 15/15 & 0.267 & 2e-124 \\
heloc & FAB & 720 & 53 & 590 & 77 & 0/15 & 1.000 & 0.19 \\
heloc & DeepFool & 12 & 7 & 5 & 0 & 6/9 & 0.924 & 0.016 \\
heloc & MIP-EXP & 379 & 37 & 300 & 42 & 0/9 & 1.000 & 0.0063 \\
heloc & MIP-OBJ & 297 & 71 & 193 & 33 & 2/6 & 1.001 & 4.4e-11 \\
all & DASH + data seed & 2982 & 47 & 2883 & 52 & 0/60 & 1.000 & 0.57 \\
all & Wachter & 2782 & 2386 & 342 & 54 & 59/60 & 0.745 & $<10^{-300}$ \\
all & DiCE & 2809 & 2763 & 32 & 14 & 60/60 & 0.724 & $<10^{-300}$ \\
all & Greedy* & 2807 & 2799 & 6 & 2 & 60/60 & 0.339 & $<10^{-300}$ \\
all & GS* & 2166 & 2071 & 51 & 44 & 60/60 & 0.437 & $<10^{-300}$ \\
all & PGD-bisect* & 2768 & 1273 & 1220 & 275 & 30/60 & 0.995 & 5.5e-223 \\
all & FMN* & 2454 & 1216 & 1069 & 169 & 30/60 & 0.982 & 4.5e-212 \\
all & NICE & 2963 & 2742 & 177 & 44 & 60/60 & 0.537 & $<10^{-300}$ \\
all & FAB & 733 & 53 & 603 & 77 & 0/24 & 1.000 & 0.19 \\
all & DeepFool & 12 & 7 & 5 & 0 & 6/9 & 0.924 & 0.016 \\
all & MIP-EXP & 1482 & 159 & 1141 & 182 & 3/36 & 1.000 & 1.4e-06 \\
all & MIP-OBJ & 1181 & 91 & 946 & 144 & 2/26 & 1.000 & 0.054 \\
\bottomrule
\end{tabular}}
\ifdim\wd0>\linewidth\resizebox{\linewidth}{!}{\box0}\else\box0\fi
\end{table}

\begin{table}[htbp]
\centering\scriptsize\setlength{\tabcolsep}{3pt}
\caption{Validity (\%) per arm and cell, $\ell_\infty$ objective.}
\label{tab:carla_linf_validity}
\setbox0=\hbox{%
\begin{tabular}{lrrrrrrrrrrrrrr}
\toprule
dataset & depth & DASH & DASH + data seed & Wachter & DiCE & Greedy* & GS* & PGD-bisect* & FMN* & NICE & FAB & DeepFool & MIP-EXP & MIP-OBJ \\
\midrule
adult & 2 & 100 & 100 & 95 & 99 & 100 & 71 & 92 & 84 & 100 & 0 & 0 & 100 & 99 \\
 & 4 & 100 & 100 & 88 & 97 & 100 & 61 & 83 & 76 & 100 & 0 & 0 & 100 & 100 \\
 & 8 & 100 & 100 & 77 & 83 & 82 & 25 & 73 & 70 & 100 & 0 & 0 & 7 & 0 \\
 & 16 & 100 & 100 & 89 & 91 & 91 & 46 & 85 & 83 & 100 & 0 & 0 & -- & -- \\
 & 32 & 100 & 100 & 82 & 81 & 83 & 23 & 74 & 69 & 100 & 0 & 0 & -- & -- \\
compas & 2 & 99 & 99 & 89 & 97 & 97 & 61 & 87 & 79 & 98 & 0 & 0 & 99 & 99 \\
 & 4 & 98 & 98 & 87 & 96 & 97 & 51 & 85 & 74 & 95 & 0 & 0 & 92 & 96 \\
 & 8 & 98 & 98 & 89 & 93 & 85 & 35 & 87 & 85 & 98 & 0 & 0 & 48 & 0 \\
 & 16 & 98 & 98 & 91 & 93 & 79 & 36 & 90 & 89 & 97 & 0 & 0 & -- & -- \\
 & 32 & 100 & 100 & 79 & 62 & 77 & 41 & 93 & 87 & 99 & 0 & 0 & -- & -- \\
gmc & 2 & 99 & 99 & 99 & 99 & 99 & 99 & 99 & 73 & 97 & 1 & 0 & 99 & 97 \\
 & 4 & 100 & 100 & 99 & 99 & 99 & 100 & 100 & 76 & 98 & 3 & 0 & 100 & 97 \\
 & 8 & 99 & 99 & 97 & 92 & 91 & 99 & 99 & 64 & 98 & 1 & 0 & 90 & 2 \\
 & 16 & 99 & 99 & 99 & 99 & 99 & 99 & 99 & 63 & 99 & 3 & 0 & -- & -- \\
 & 32 & 98 & 98 & 96 & 92 & 93 & 98 & 98 & 66 & 98 & 1 & 0 & -- & -- \\
heloc & 2 & 100 & 100 & 100 & 100 & 100 & 100 & 100 & 99 & 100 & 100 & 2 & 100 & 100 \\
 & 4 & 100 & 100 & 100 & 100 & 100 & 100 & 100 & 100 & 100 & 100 & 1 & 93 & 98 \\
 & 8 & 100 & 100 & 100 & 99 & 99 & 100 & 100 & 100 & 100 & 96 & 1 & 59 & 0 \\
 & 16 & 100 & 100 & 100 & 100 & 100 & 100 & 100 & 100 & 100 & 96 & 1 & -- & -- \\
 & 32 & 100 & 100 & 100 & 100 & 100 & 100 & 100 & 100 & 100 & 88 & 2 & -- & -- \\
\bottomrule
\end{tabular}}
\ifdim\wd0>\linewidth\resizebox{\linewidth}{!}{\box0}\else\box0\fi
\end{table}

\begin{table}[htbp]
\centering\scriptsize\setlength{\tabcolsep}{3pt}
\caption{MNIST (3 vs 8, $n=784$, depth 8, width 256, 150 factuals), $\ell_\infty$ objective. Greedy* did not run at this scale; under $\ell_1$ the norm-adapted GS* and PGD-bisect* runs cover 70 of the 150 factuals (validity over the arm's own rows). PBMC3k is reported in the main text, after the shared marker completion. DASH = the CPU implementation; NA = no valid counterfactual, or a metric the run did not record.}
\label{tab:carla_linf_large}
\setbox0=\hbox{%
\begin{tabular}{lrrrrrrrrr}
\toprule
dataset & method & success \% & valid \% & $\ell_\infty$ & l0 & redund. & yNN & yNN-dist & time (s) \\
\midrule
mnist & DASH & \textbf{100} & \textbf{100} & 0.075 & 477.0 & 59.5 & 0.000 & 5.194 & 74.2436 \\
mnist & DASH + data seed & \textbf{100} & \textbf{100} & 0.075 & 480.0 & 59.5 & 0.000 & 5.176 & 74.8974 \\
mnist & Wachter & \textbf{100} & \textbf{100} & 0.105 & 766.5 & 407.5 & 0.000 & NA & 2.0280 \\
mnist & DiCE & 77 & 77 & 0.100 & 161.5 & 0.0 & 0.000 & NA & 0.0372 \\
mnist & GS* & \textbf{100} & \textbf{100} & 0.838 & 33.0 & 23.5 & 0.000 & 6.251 & 0.2176 \\
mnist & PGD-bisect* & \textbf{100} & \textbf{100} & \textbf{0.069} & 477.0 & 27.0 & 0.000 & 5.127 & 3.0292 \\
mnist & FMN* & \textbf{100} & \textbf{100} & \textbf{0.069} & 476.0 & 70.0 & 0.000 & NA & 0.8799 \\
mnist & NICE & \textbf{100} & \textbf{100} & 0.992 & \textbf{22.0} & \textbf{6.0} & \textbf{0.200} & NA & 0.3853 \\
mnist & FAB & \textbf{100} & \textbf{100} & 0.070 & 479.0 & 35.0 & 0.000 & \textbf{5.117} & \textbf{0.1361} \\
mnist & DeepFool & 100 & 0 & NA & NA & NA & NA & NA & NA \\
\bottomrule
\end{tabular}}
\ifdim\wd0>\linewidth\resizebox{\linewidth}{!}{\box0}\else\box0\fi
\end{table}

\begin{figure}[htbp]
\centering
\includegraphics[width=\linewidth]{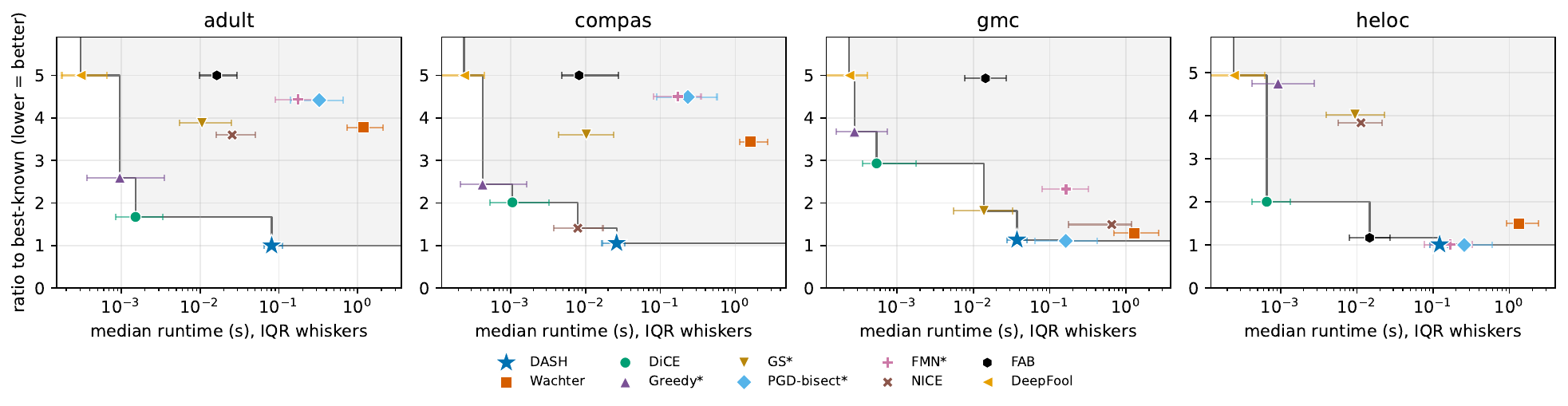}
\caption{Runtime-quality trade-off, $\ell_\infty$ objective (see Figure~\ref{fig:carla_pareto} for the definition).}
\label{fig:carla_linf_pareto}
\end{figure}
\begin{figure}[htbp]
\centering
\includegraphics[width=\linewidth]{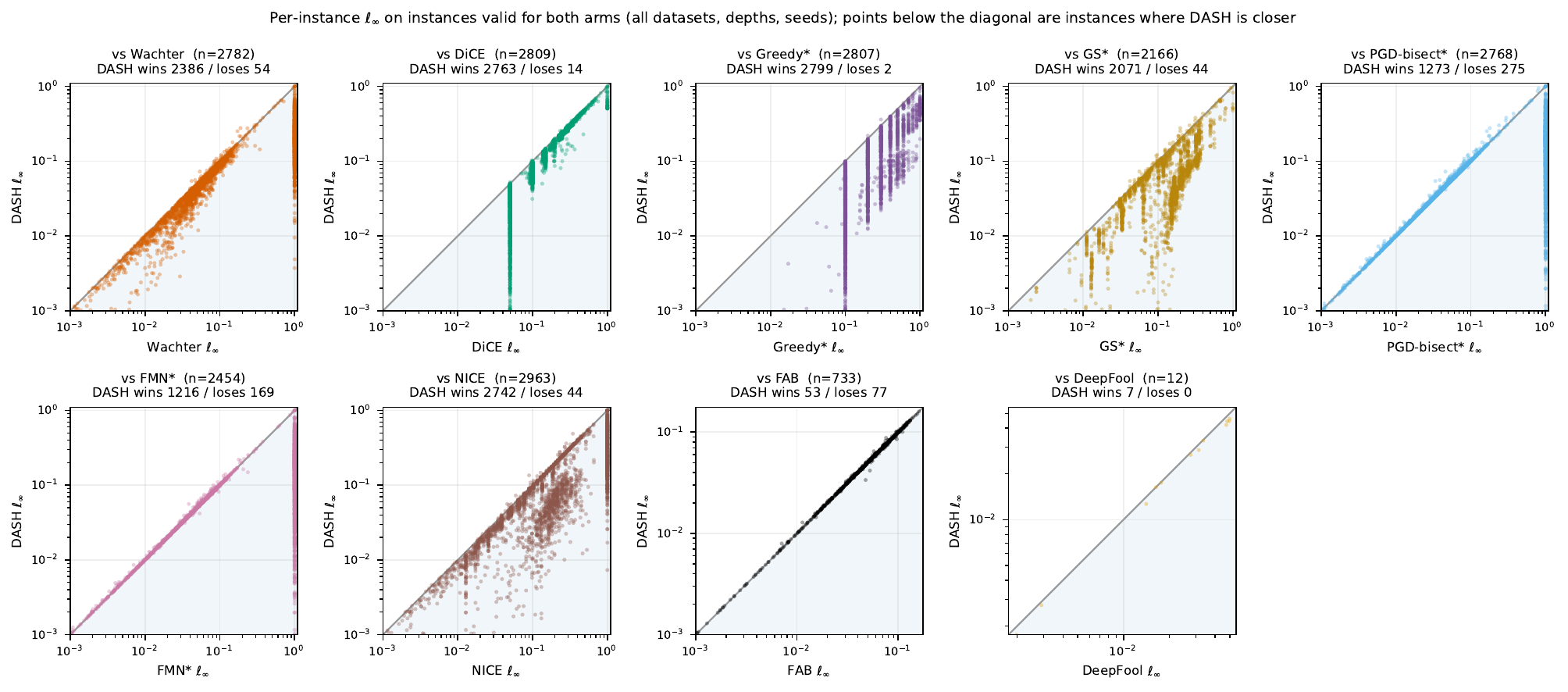}
\caption{Per-instance paired comparison, $\ell_\infty$ objective (see Figure~\ref{fig:carla_paired} for the definition).}
\label{fig:carla_linf_paired}
\end{figure}
\begin{figure}[htbp]
\centering
\includegraphics[width=\linewidth]{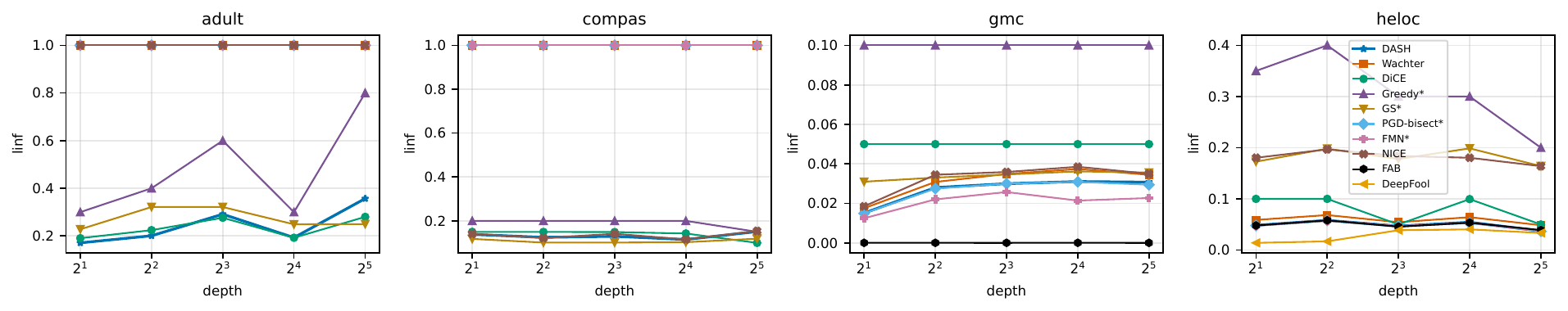}\\
\includegraphics[width=\linewidth]{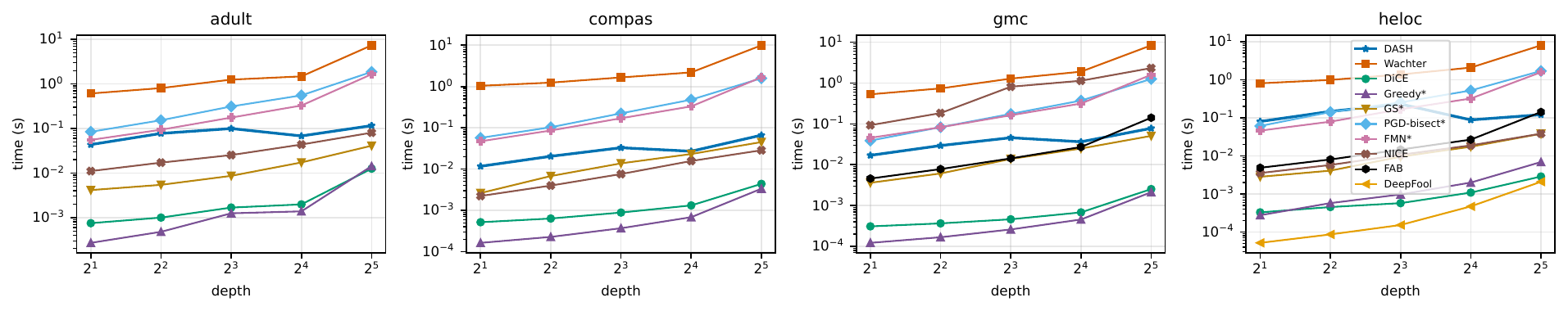}
\caption{Median $\ell_\infty$ and runtime versus depth, $\ell_\infty$ objective.}
\label{fig:carla_linf_depth}
\end{figure}
\clearpage
{\scriptsize\setlength{\tabcolsep}{3pt}
\begin{longtable}{lrrrrrrrrrrrr}
\caption{All tabular cells, $\ell_\infty$ objective (medians over valid rows; bold = best per column within a cell among arms with validity $\geq$ 80\%). Medians are over each arm's own valid rows; an arm with low validity is summarised on an easier subset, so compare arms with different validity through the paired table (Table~\ref{tab:carla_linf_wtl}) and the best-known table (Table~\ref{tab:carla_linf_bestknown}).} \label{tab:carla_linf_per_cell} \\
\toprule
dataset & depth & method & success \% & valid \% & $\ell_\infty$ & l0 & redund. & yNN & yNN-dist & time (s) & $\ell_1$ & $\ell_2$ \\
\midrule
\endfirsthead
\caption[]{All tabular cells, $\ell_\infty$ objective (medians over valid rows; bold = best per column within a cell among arms with validity $\geq$ 80\%). Medians are over each arm's own valid rows; an arm with low validity is summarised on an easier subset, so compare arms with different validity through the paired table (Table~\ref{tab:carla_linf_wtl}) and the best-known table (Table~\ref{tab:carla_linf_bestknown}).} \\
\toprule
dataset & depth & method & success \% & valid \% & $\ell_\infty$ & l0 & redund. & yNN & yNN-dist & time (s) & $\ell_1$ & $\ell_2$ \\
\midrule
\endhead
\midrule
\multicolumn{13}{r}{Continued on next page} \\
\midrule
\endfoot
\bottomrule
\endlastfoot
adult & 2 & DASH & \textbf{100} & \textbf{100} & 0.171 & 3.0 & \textbf{0.0} & 0.000 & 0.407 & 0.0435 & 0.553 & 0.300 \\
adult & 2 & DASH + data seed & \textbf{100} & \textbf{100} & 0.171 & 3.0 & \textbf{0.0} & 0.000 & 0.407 & 0.0486 & 0.552 & 0.300 \\
adult & 2 & Wachter & 96 & 95 & 1.000 & 5.0 & 3.0 & 0.000 & 1.425 & 0.6108 & 2.258 & 1.421 \\
adult & 2 & DiCE & \textbf{100} & 99 & 0.191 & 3.0 & \textbf{0.0} & 0.000 & 0.411 & 0.0007 & 0.451 & \textbf{0.294} \\
adult & 2 & Greedy* & \textbf{100} & \textbf{100} & 0.300 & \textbf{1.0} & \textbf{0.0} & 0.000 & 0.400 & \textbf{0.0003} & \textbf{0.300} & 0.300 \\
adult & 2 & GS* & 100 & 71 & 0.228 & 1.0 & 0.0 & 0.000 & 0.339 & 0.0041 & 0.251 & 0.251 \\
adult & 2 & PGD-bisect* & 92 & 92 & 1.000 & 5.0 & 3.0 & 0.000 & 1.425 & 0.0842 & 2.256 & 1.421 \\
adult & 2 & FMN* & 84 & 84 & 1.000 & 5.0 & 2.5 & 0.000 & 1.425 & 0.0554 & 2.260 & 1.421 \\
\end{longtable}
}

\clearpage

\end{document}